\documentclass[preprint,12pt,authoryear,times]{elsarticle}
\usepackage[margin=1.2in]{geometry}
\usepackage[dvipsnames]{xcolor}
\usepackage{inconsolata} %
\usepackage{enumitem}
\usepackage[normalem]{ulem}
\usepackage{booktabs}
\usepackage{longtable}
\usepackage{multirow}
\usepackage{float}
\usepackage{expex}
\usepackage[T1]{fontenc}
\usepackage[hidelinks]{hyperref}
\usepackage{caption}
\usepackage{spverbatim}
\usepackage{subcaption}
\usepackage{appendix}

\newcommand{\gap}[1][0.2in]{\underline{\hspace*{#1}}}

\definecolor{advantage}{HTML}{CC6677} 
\definecolor{estimation}{HTML}{6699CC}
\definecolor{price}{HTML}{117733}
\definecolor{duration}{HTML}{332288}
\definecolor{ooze}{HTML}{AA4499}
\definecolor{agt-pat}{HTML}{44AA99}
\definecolor{exp-th}{HTML}{E59C00}

\begin{document}
\begin{frontmatter}
\title{Manipulating language models’ training data to study syntactic constraint learning: the case of English passivization}
\author[1]{Cara Su-Yi Leong\corref{cor1}}\ead{caraleong@nyu.edu}
\author[2]{Tal Linzen}
\cortext[cor1]{Corresponding author}
\affiliation[1]{organization={Department of Linguistics, New York University},
           addressline={10 Washington Place}, 
           city={New York},
           postcode={10003}, 
           state={NY},
           country={USA}}
\affiliation[2]{organization={Department of Linguistics and Center for Data Science, New York University},
           addressline={60 5th Avenue}, 
           city={New York},
           postcode={10012}, 
           state={NY},
           country={USA}}
\begin{abstract}
    Grammatical rules in natural languages are often characterized by exceptions. How do language learners learn these exceptions to otherwise general patterns? Here, we study this question through the case study of English passivization. While passivization is in general quite productive, there are cases where it cannot apply (cf. the following sentence is ungrammatical: \textit{*One hour was lasted by the meeting}). Using neural network language models as theories of language acquisition, we explore the sources of \emph{indirect evidence} that a learner can leverage to learn whether a verb can be passivized. We first characterize English speakers' judgments of exceptions to the passive, and confirm that speakers find some verbs more passivizable than others. We then show that a neural network language model's verb passivizability judgments are largely similar to those displayed by humans, suggesting that evidence for these exceptions is available in the linguistic input. 
    Finally, we test two hypotheses as to the source of evidence that language models use to learn these restrictions: frequency (entrenchment) and semantics (affectedness). We do so by training models on versions of the corpus that have had sentences of the types implicated by each hypothesis removed, altered, or introduced. We find support for both hypotheses: entrenchment and affectedness make independent contributions to a verb's passivizability. From a methodological point of view, this study highlights the utility of altering a language model's training data for answering questions where complete control over a learner's input is vital. 
\end{abstract}
\end{frontmatter}

\section{Introduction}
Grammatical generalizations in natural languages are often characterized by systematic exceptions: classes of cases where the generalizations do not apply. What sources of evidence do learns use to acquire these exceptions? In the current study, we address this question through a computational study of neural network language models' learning of the constraints on English passivization. Passivization is in general quite productive in English: Speakers freely use most transitive verbs in both active and passive voice, and English-speaking children who learn novel transitive verbs in the active voice can use those verbs  in the passive voice \citep{brookstomasello1999,pinkeretal1987}. But for a small set of verbs this generalization does not hold; the verb \textit{last}, for example, is acceptable in the active but not in the passive:
\pex\label{last-example}
\a\label{last-active} The meeting lasted one hour.
\a\ljudge{*}\label{last-passive} One hour was lasted by the meeting.
\xe

\noindent Why do speakers judge (\lastx b) as unacceptable? One tempting explanation may be that they do so because they simply have never heard a sentence such as (\lastx b), where \emph{last} is passivized. But that explanation is most likely insufficient. Consider the verb \textit{defenestrate}:

\pex
\label{defenestrate-example}
\a The writer defenestrated the editor.
\a The editor was defenestrated by the writer.
\xe 
Because \textit{defenestrate} is a rare lemma, and passives are rare in everyday speech (on average, only one out of ten utterances uses the passive voice; \citealt{rolandetal2007}), the odds that a particular speaker has heard a sentence like (\lastx b), where \textit{defenestrate} is passivized, are very low, quite possibly as low as for (\ref{last-passive}). Yet English speakers judge (\lastx b), but not (\ref{last-passive}), as acceptable. How do learners of English consistently arrive at a grammar under which \textit{last} cannot be passivized but \textit{defenestrate} can? This learnability challenge---separating forms that do not occur because they are unacceptable from forms that are not observed simply due to chance---is sometimes referred to in linguistics as ``Baker's Paradox'' \citep{baker1979}.

Could innate constraints account for exceptions to passivization? The answer to this question is unlikely to be entirely positive: because exceptions to passivization vary across languages, they need to be learned. For example, stative verbs like \textit{cost} and \textit{have} cannot be passivized in English, but they can in Kinyarwanda \citep{keenandryer2007}:
\ex\label{unpassivizable}\ljudge{*} A new car is had by John.
\xe
\ex
\begingl
\gla Ibifuungo bibiri bi-fit-w-e n-\^{\i}sha\^{a}ti//
\glb
buttons two they-have-\textsc{pass}-\textsc{asp} by-shirt//
\glft `Two buttons are had by the shirt.'//
\trailingcitation{\citep[332]{keenandryer2007}}
\endgl
\xe
While there are certain regularities across languages in terms of the set of verbs that can and cannot passivize \citep{ambridgeetal2023}, the particular verbs that are subject to restrictions on passivization are language-specific. Since this information is likely not explicitly taught to children by caregivers, it must be acquired by learners of the language through exposure to \emph{indirect evidence}.

In this paper, we explore two hypotheses concerning the kinds of indirect evidence for these exceptions that learners might rely on. According to the \emph{entrenchment hypothesis} \citep{brainebrooks1995,goldberg2006,theakston2004}, learners use the statistical distribution of verbs in different constructions to determine where a verb can appear and infer where it cannot occur. Under this hypothesis, learners who never encounter a particular verb in the passive but see it consistently in other contexts will conclude that the verb cannot appear in the passive.

A second potential source of indirect evidence for passive exceptions is the \emph{lexical semantics} of the verb \citep{ambridgeetal2016,pinker1989}. Under this hypothesis, a verb is passivizable when it denotes an action whose theme participant is \emph{affected} \citep{pinker1989}, that is, the theme undergoes a change in state, location, or existence caused by the action's agent participant \citep{beavers2011}.  Verbs inconsistent with these semantics are unacceptable in the passive; (\ref{last-passive}) \textit{*One hour was lasted by the meeting}, for example, is ungrammatical because \textit{one hour} is not affected by \textit{the meeting} lasting an hour.

Both the affectedness of a verb and its degree of entrenchment in the active, measured by how frequently the verb occurs in the active compared to the passive, \emph{correlate} with English speakers' judgments of its passivizability \citep{ambridgeetal2016}. Yet it is difficult to study if these factors \emph{causally} affect the learning of restrictions on passivization. Since verbs with low-affectedness semantics are used in the passive only very infrequently, affectedness and entrenchment are highly correlated, and it is hard to know whether speakers make passivizability judgments using one of these factors to the exclusion of the other, or possibly using an entirely different source of evidence. In a counterfactual world, we could experimentally disentangle these factors by manipulating the language that a human language learner is exposed to, for example by artificially increasing the frequency of passive forms of low-affectedness verbs. This is, of course, impossible; instead, we investigate the relationship between properties of the input and the outcome of learning using neural networks as models of language acquisition \citep{warstadtbowman2022,baroni2022}. We discuss these computational models and their role in cognitive modeling in the following section. 

\subsection{Testing Learning Hypotheses by Manipulating Language Models' Training Corpora}
Neural network language models are systems that learn probability distributions over sequences of words based on a text corpus (for a review, see \citealt{linzen2021syntactic}). While they are not trained or designed to provide acceptability judgments, such judgments can be derived from them through \textit{targeted syntactic evaluation} \citep{lauetal2017,linzenetal2016,marvinlinzen2018,warstadtetal2020}: given a minimal pair of sentences such as~(\ref{last-example}), one of which is grammatical and one is not, we use the language model to assign a probability score to each sentence. If the model assigns a higher probability to the grammatical sentence (\ref{last-active}) than the ungrammatical (\ref{last-passive}), we conclude that it shows sensitivity to the underlying syntactic differences between the two sentences. Targeted syntactic evaluation has shown that neural network language models are sensitive to a variety of syntactic and semantic constraints, including subject-verb agreement, constraints on negative polarity items, and island constraints on filler-gap dependencies \citep{linzen2021syntactic}. 

By using language models as theories of acquisition, we can address the limitation that acquisition studies with humans are by necessity observational only, as these models allowing for complete control of the input provided to the learner: we can train multiple models on corpora that differ in controlled and targeted ways, and compare how learners with the same initial state and learning objective but different input diverge in their behavior at the end of learning (\citealt{jumeletetal2021,misramahowald2024,weietal2021}).

Here, we apply this method to study the emergence of exceptions to English passivization. We train neural network language models on approximately the same amount of linguistic data that humans are exposed to, and use these models to answer two questions: first, is this amount of linguistic input sufficient for the models to learn to make judgments that are similar to human judgments? Second, what kinds of information are available in the linguistic input for the learner to come to make those judgments?

In Experiment~1, we answer the first question in the affirmative. We show that neural network language models' verb passivizability judgments are highly correlated with those of English speakers ($r = 0.9$), a significantly higher correlation that we observed for simpler frequency-based models. Such behavior suggests that language models can use evidence from the linguistic input to learn exceptions.

To answer the second question, we generate counterfactual training corpora that manipulate different sources of evidence for a verb's passivizability. In Experiments 2A and 2B, we withhold the evidence considered to be critical under the affectedness hypothesis and entrenchment hypothesis. We then train language models on both types of modified corpora, and compare the acceptability judgments provided by these to those of models trained on the original corpus. Finally, in Experiment 3, we introduce a novel verb into the corpus and manipulate how often and in what semantic environments it occurs. This makes it possible to measure the interaction between the affectedness and entrenchment hypotheses in a controlled way.

To foreshadow our results, we do not find clear support for one hypothesis to the exclusion of the other. Rather, entrenchment and affectedness each contribute to a verb's passivizability, and we do not find evidence for an interaction between them. We also find that neither of the hypotheses alone can account for the full difference between passivizable and unpassivizable verbs, which suggests that the input contains sources of evidence for a verb's passivizability other than its frequency and affectedness. More broadly, this empirical study illustrates a method by which researchers can examine how controlled changes to the learner's input affect the outcome of learning.

\subsection{Overview of experiments}
In Experiment~1, we compare human passivizability judgments with those of a model we train. This experiment has two parts. In Experiment~1A, we collect acceptability judgments from English speakers on a set of active and passive sentences containing verbs that are reported in the literature as unacceptable in the passive \citep{bach1980,levin1993,postal2004,zwicky1987}. In Experiment~1B, we compare our previously-collected human judgments against to judgments derived from a neural network language model which we train on 100 million words of English.

In Experiment~2, we evaluate the causal factors that our model used to learn to approximate human judgments. In Experiment~2A, we evaluate the entrenchment hypothesis by testing if models' acceptability judgments for passive sentences containing a highly passivizable verb change if the models are trained on a corpus in which that verb occurs much less frequently in the passive than in the original corpus. In Experiment~2B, we evaluate the affectedness hypothesis, which predicts a link between the affectedness of the theme argument and the verb's passivizability. We do so by training models on a corpus where an unpassivizable verb co-occurs with arguments that are associated with a canonically passivizable verb. This alters the proportion of the verb's arguments in the corpus that are affected, signaling to the learner that it is higher in affectedness. 

Finally, in Experiment~3, we test if a verb's frequency of occurrence and its semantic context interact in determining its passivizability. To do so, we introduce a novel verb which only occurs in active-voice sentences, and vary the semantic contexts and frequency (and therefore active-to-passive ratio) with which the novel verb occurs.

\section{Materials}
English passives are subject to some restrictions that are uncontroversial and do not appear challenging to learn; in particular, the entire class of intransitive verbs cannot occur in the passive \citep{comrieetal1977}. In this work, we focus on more subtle restrictions on the passivization of verbs that appear to be transitive, in that they are typically followed by a noun phrase; the evidence for the exceptionality of these verbs is less clear. To compare human and language model judgments for such verbs, we created a dataset of 140 pairs of active and passive sentences---280 sentences in total---using 28 different verbs, 10 of which were ordinary passivizable verbs (\textit{control verbs}), and 18 verbs characterized as unpassivizable in the linguistics literature (\textit{critical verbs}; \citealt{bach1980,levin1993,postal2004,zwicky1987}). Each sentence pair consisted of an active transitive sentence (e.g. \textit{A boy dropped the cup}) and a corresponding passive sentence, where the passive is introduced by a form of the verb \textit{be} (e.g. \textit{The cup was dropped by a boy}). Passive sentences always included an explicit \textit{by}-phrase that matched the subject of the active sentence (i.e. \textit{The cup was dropped by a boy}, but not \textit{The cup was dropped}).

The critical verbs belonged to five verb classes. The meanings of the verbs in each class were sufficiently similar that they could all be substituted into the same frame. We constructed five frames for each class; for examples of the frames, see Table~\ref{tab:sentence_frames}, and for the full list of critical and control sentences see \ref{sec:test-sentences}. The verbs in each verb class were:

\begin{itemize}[noitemsep]
    \item \textcolor{advantage}{Advantage} verbs: \textit{benefit}, \textit{help}, \textit{profit}, \textit{strengthen}
    \item \textcolor{price}{Price} verbs: \textit{cost}, \textit{earn}, \textit{fetch}
    \item \textcolor{ooze}{Ooze} verbs: \textit{discharge}, \textit{emanate}, \textit{emit}, \textit{radiate}
    \item \textcolor{duration}{Duration} verbs: \textit{last},  \textit{require}, \textit{take}
    \item \textcolor{estimation}{Estimation} verbs: \textit{approximate}, \textit{match}, \textit{mirror}, \textit{resemble}
\end{itemize}

\noindent Some of the test verbs have multiple senses, of which only one is exceptional. Our sentences used the sense of the verb reported as unpassivizable in the literature. For \textit{take}, for instance, sentences did not use the sense in (\nextx a), only the one illustrated in (\nextx b):

\pex
\a The photo was \uline{taken} by the boy.
\a\ljudge{*} Two days was \uline{taken} by the meeting. 
\xe

\begin{table*}[t]
    \centering
    \begin{tabular}{lll}
        \toprule
        Verb class & Active sentence frame & Passive sentence frame\\
        \midrule
         \textcolor{advantage}{Advantage} & The gift \gap{} my organization. & My organization was \gap{} by the gift. \\
         \textcolor{price}{Price} & Your book \gap{} thirty dollars. & Thirty dollars was \gap{} by your book.\\
         \textcolor{ooze}{Ooze} & My machine \gap{} a sound. & A sound was \gap{} by my machine. \\
         \textcolor{duration}{Duration} & Her speech \gap{} seventeen minutes & Seventeen minutes was \gap{} by her speech.\\
         \textcolor{estimation}{Estimation} & Your friend \gap{} my brother. & My sketch was \gap{} by your friend.\\
         \bottomrule
    \end{tabular}
    \caption{\textit{Example sentence frames}. Each verb in the verb class was substituted into frames specific to the class (the table shows one of the five frames used for each class).}
    \label{tab:sentence_frames}
\end{table*}

All verbs in the class were inserted into each sentence frame, resulting in 90 total test sentence pairs: 20 pairs each from the \textcolor{advantage}{advantage}, \textcolor{ooze}{ooze} and \textcolor{estimation}{estimation} classes, which have four test verbs each, and 15 pairs from the \textcolor{price}{price} and \textcolor{duration}{duration} classes, which have three test verbs each. Example (\nextx) demonstrates a sentence pair generated from the sentence frame in Table~\ref{tab:sentence_frames} using the verb \textit{benefit}:

\pex
\a The gift \uline{benefited} my organization.
\a My organization was \uline{benefited} by the gift.
\xe

In addition to the five critical verb classes, which contained verbs expected to be unacceptable in the passive, we created stimuli for two control verb classes which we expected to be acceptable in both the active and the passive voice:
\begin{itemize}[leftmargin=*,noitemsep]
    \item \textcolor{agt-pat}{Agent-patient}: \textit{hit}, \textit{push}, \textit{wash}, \textit{drop}, \textit{carry}
    \item \textcolor{exp-th}{Experiencer-theme}: \textit{see}, \textit{hear}, \textit{know}, \textit{like}, \textit{remember}
\end{itemize}
\noindent Given the diverse semantics of the verbs in these groups, we used unique sentence frames for each verb. This resulted in 50 control test sentence pairs, and a total of 140 sentence pairs across both critical and control verbs.

The experiment also included filler sentences with similar lengths to the critical items (for a list, see \ref{sec:filler-sentences}). Since the passives of control sentences were expected to be acceptable, we included a larger number of ungrammatical than grammatical fillers (52 vs. 26) such that the experiment as a whole contained the same number of grammatical and ungrammatical sentences.

\section{Experiment~1A: English speakers find some verbs more passivizable than others}
We first conducted a human acceptability judgment study. This study had two goals: first, to verify the linguists' judgments reported in the syntax literature; and second, to measure any gradient differences in the degree to which different verb classes and individual verbs can be passivized, providing a fine-grained benchmark against which language model judgments can be compared.\footnote{This work was originally reported in \cite{leonglinzen2023}.}

\subsection{Procedure}
We collected acceptability judgments from English speakers for the active and passive sentences described in the Materials section. Each participant rated either the active or the passive version of any given sentence pair. Specifically, we divided the 140 sentence pairs into two groups of 70 sentence pairs (i.e. 140 sentences) such that each group contained either two or three sentence frames per verb. We then split each group into two sets of 70 sentences such that the active and passive versions of each item were in different sets.

We further counterbalanced the presentation order by creating four ordered lists for each group, as follows. We organized each group into two lists such that an item that appeared in the first half of one list appeared in the second half of the other list. We pseudorandomized the order of items within those lists to avoid more than two consecutive active sentences, more than two consecutive passive sentences, and more than two consecutive sentences from the same verb class. Each experimental sentence (critical or control) was followed by at least one filler sentence. We then created reversed versions of these two lists, resulting in four sentence lists per group (eight sentence lists in total).

\begin{figure}[t]
    \centering
    \includegraphics[width=.8\textwidth]{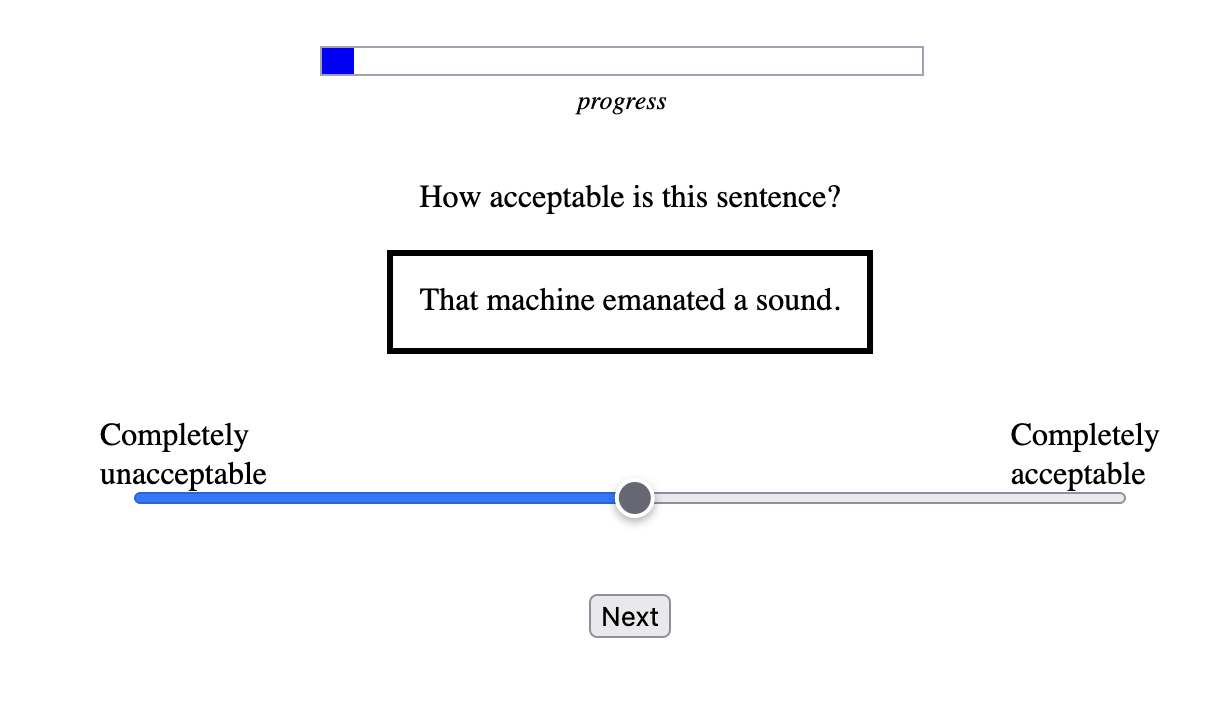}
    \caption{An example of a trial in the human acceptability judgment experiment.}
    \label{fig:survey_question}
\end{figure}

Participants were instructed to rate the acceptability of each sentence based on their ``gut reaction'', and were told that there were no right or wrong answers (for the full instructions, see \ref{sec:instructions}). They rated sentences by moving a slider from ``completely unacceptable'' to ``completely acceptable''; the location of the slider corresponded to an integer score between 0 and 100. This score was not made visible to participants. Participants could not rate a sentence with a score of 50 (the initial location of the slides): they had to move the slider at least slightly to the right or left on each trial such that the score was either lower or higher than 50. Before the experiment began, participants were familiarized with the experimental setup by rating two practice sentences, one of which was expected to be fully acceptable (\textit{The mirrors reflected light}) and one expected to be unacceptable (\textit{The teacher was spoke}). In one of the practice trials, participants were told that many people find the sentence acceptable, and if they agree they should move the slider to the right edge of the scale. The other practice trial was similar except the sentence was unacceptable. Figure~\ref{fig:survey_question} shows an example of the interface that participants used to rate sentences.

\subsection{Participants}
We used the Prolific crowdsourcing platform to recruit 84 participants whose IP addresses were located in the US and who self-reported as native English speakers. Each participant rated 140 sentences (70 test sentences and 70 filler sentences) and was paid US\$3.50. The experiment took an average of 12 minutes to complete.

\subsection{Results}
\begin{figure}[t]
    \centering
    \includegraphics[width=\linewidth]{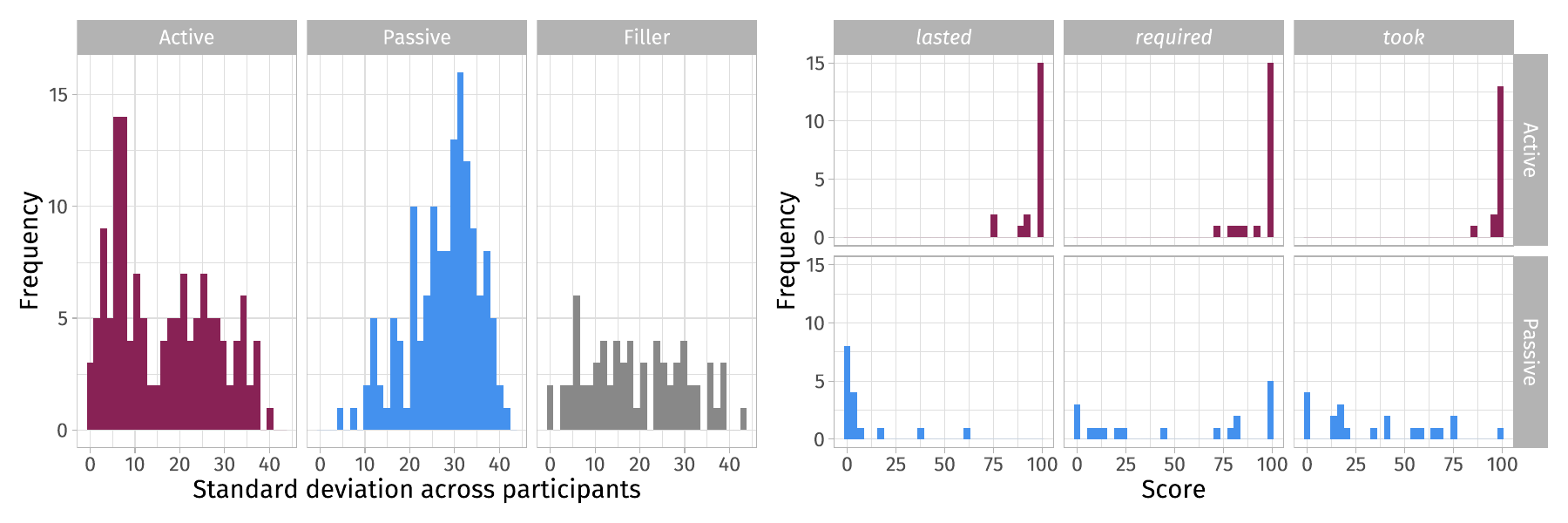}
    \caption{\textit{Distribution of human acceptability judgments}. Left: Histogram of standard deviations of acceptability ratings across participants. Active sentences received more consistent ratings than passive sentences. Right: Histogram of the ratings of the sentence pairs with the active frame \textit{Her speech \gap{} seventeen minutes} and passive frame \textit{Seventeen minutes was \gap{} by her speech}, for the three verbs \textit{lasted}, \textit{required} and \textit{took}, illustrating differences in the spread of ratings across sentences.}
    \label{fig:rating-distribution}
\end{figure}

\begin{figure}[t]
    \centering
    \includegraphics[width=\linewidth]{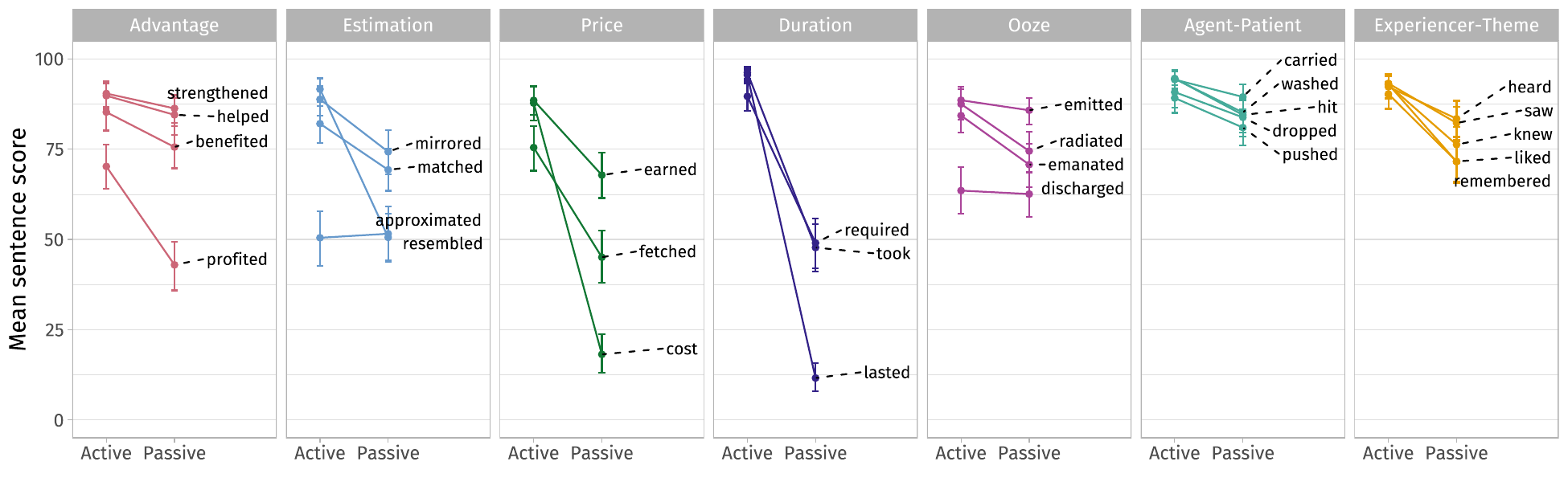}
    \caption{\textit{Passive drop in human acceptability judgments of active and passive sentences by verb}. The steeper the downward gradient between active and passive conditions, the larger the passive drop. Error bars indicate bootstrapped 95\% confidence intervals.}
    \label{fig:human-duckbill}
\end{figure}
Participants were excluded from analysis if they rated more than 15 filler sentences differently than expected, either by giving ungrammatical sentences scores above 50 or giving grammatical sentences scores below 50. This resulted in the exclusion of 24 participants. In the analysis that follows, each sentence was rated by at least 13 participants.

Across all verb classes, participants gave higher scores on average to active sentences (mean: 88.5 points) than passive sentences (mean: 66.4 points). Participants were also in greater agreement with each other in their judgments for active than passive sentences (Figure~\ref{fig:rating-distribution}, left). The right panel of Figure~\ref{fig:rating-distribution} illustrates this pattern with one particular pair of sentence frames: \textit{Her speech \gap{} seventeen minutes} and \textit{Seventeen minutes was \gap{} by her speech}. Active sentences with this frame received ratings that were close to maximal, regardless of the verb, while passive sentences showed more variance: judgments for \textit{lasted} in this frame were unimodal and close to 0, judgments for \textit{took} were spread relatively uniformly across the entire scale, and judgments for verb \textit{required} were bimodal. We leave an investigation of the variability across participants to future work.

We next report differences in judgments between the active and passive sentences. In what follows, we focus on comparisons between the active and passive sentences within a sentence pair; because these sentences contained the same lexical items except for the auxiliary \textit{was}/\textit{were} and \textit{by}, which are common across all sentences, comparing the sentences within a pair isolates the effect of passivization from any frequency effects that might increase the acceptability of sentences with more common verbs such as \textit{helped} compared to low-frequency ones such as \textit{profited}. We define the \textbf{passive drop} of a sentence pair as the difference in mean acceptability rating between its active and passive version. The results are reported in Figure~\ref{fig:human-duckbill}; a steeper downward slope corresponds with a larger average passive drop for sentences with the verb in question.

Although the average passive drop was positive for all verbs, its magnitude differed considerably across verb classes. The \textcolor{duration}{duration} class showed the largest mean passive drop (61.9 points). The \textcolor{ooze}{ooze} class showed the lowest mean passive drop (8.4 points); in fact, although verbs from this class are considered in the linguistics literature to be unpassivizable, their mean passive drop was comparable to that observed for the \textcolor{agt-pat}{agent-patient} class of control verbs, which are considered to be passivizable (8.9 points). In summary, this experiment confirmed linguists' judgments for a majority of verbs, but not all of them.

To determine whether the difference in passive drop between verb classes was significant, we fit a linear mixed-effects model to predict \textsc{sentence score} from the following predictors: as fixed effects, \textsc{sentence type} and \textsc{verb class} as well as their interaction; \textsc{frame} and \textsc{verb} as random intercepts; and by-participant random slopes and intercepts for \textsc{sentence type}. We used the canonically passivizable \textcolor{agt-pat}{agent-patient} verb class as the reference level.
We found significant interactions between \textsc{sentence type} and \textsc{verb class} in four cases: \textcolor{estimation}{estimation} verbs, \textcolor{price}{price} verbs, \textcolor{duration}{duration} verbs and \textcolor{exp-th}{experiencer-theme} verbs (all $p<0.001$). This indicates that the passive drop for these verb classes was significantly different from the passive drop for the canonically passivizable \textcolor{agt-pat}{agent-patient} verbs. On the other hand, there was no significant interaction between \textsc{sentence type} and \textsc{verb class} in the sentence scores obtained from \textcolor{agt-pat}{agent-patient} verbs and \textcolor{ooze}{ooze} verbs ($p=0.75$) or \textcolor{advantage}{advantage} verbs ($p=0.1$).

Within the verb classes that were significantly less passivizable than \textcolor{agt-pat}{agent-patient} verbs, some verbs were more passivizable than others. To examine this variation, we fit separate linear mixed-effects models within each verb class, predicting \textsc{sentence score} from \textsc{verb} and \textsc{sentence type} as fixed effects, with random intercepts for \textsc{frame} and by-participant random slopes for \textsc{sentence type}. We used the verb with the lowest passive drop in the class as the reference level. We found a significant interaction between \textsc{sentence type} and \textsc{verb} in some but not all cases. Within the \textcolor{duration}{duration} class, for example, \textcolor{duration}{\textit{last}} was significantly less passivizable than \textcolor{duration}{\textit{required}} ($p<0.001$), but \textcolor{duration}{\textit{took}} was not ($p=0.36$). Likewise, within the \textcolor{price}{price} class, \textcolor{price}{\textit{cost}} was less passivizable than \textcolor{price}{\textit{earned}} ($p<0.001$) but \textcolor{price}{\textit{fetched}} was not ($p= 0.26$). 
These results point to the fact that, even among verbs that can occur in the same frames, some verbs are more passivizable than others. This suggests either that verb-specific learning is necessarily \citep{zwicky1987}, or that speakers are sensitive to fine-grained semantic differences across verbs in the same class.

\subsection{Estimating judgment reliability across participants}
How much of the variance in acceptability judgments across verbs can we hope to explain using a computational model, and how much of it is due to inherent variability in our measurements? To answer this question, we conducted a split-half reliability analysis, following \citet{huang2024large}; this analysis is predicated on the premise that a computational model cannot be expected to show a higher correlation with the empirical data than two randomly sampled halves of the data are expected to show with each other. 

We repeated the following analysis ten times, and averaged the ten sets of results. In each instance of the analysis, participants were randomly split into two groups. We computed the mean acceptability judgment score of each item within each half. We then calculated the Pearson correlation coefficient between the sets of estimates derived from each half of the results. The mean of these ten correlation coefficients was then entered into the Spearman-Brown prophecy formula \citep{spearman1910} to calculate the reliability coefficient. We found that the reliability of the collected acceptability judgment scores was high across all test items (0.93 on average). While reliability was somewhat lower for some verb classes (the lowest was 0.71, for \textcolor{agt-pat}{agent-patient}; for the full results by verb class, see Table~\ref{tab:split-half}), overall we conclude that participants' judgments were largely consistent for most items. 

\begin{table}[h]
    \centering
    \begin{tabular}{ll}
        \toprule
        \textbf{Verb class} &  \textbf{Split-half reliability}\\
         \midrule
        All items except fillers & 0.93 \\
        \midrule
        Advantage & 0.88 \\
        Estimation & 0.91 \\
        Price & 0.95 \\
        Duration & 0.97 \\
        Ooze & 0.84 \\
        Agent-Patient & 0.71\\
        Experiencer-Theme & 0.83\\
        \midrule
        Fillers & 0.99
        \\
        \bottomrule
    \end{tabular}
    \caption{\textit{Spearman-Brown-corrected split-half reliability for each verb class}. Acceptability judgments showed high reliability on all test items as well as within verb classes.}
    \label{tab:split-half}
\end{table}

\subsection{Discussion}

Overall, we found that for the vast majority of verbs,  including canonically passivizable \textcolor{agt-pat}{agent-patient} ones, participants rated active sentences more highly than passive ones (the only exception was \textcolor{estimation}{\textit{approximated}}). This difference may reflect pragmatic factors: each sentence in the acceptability judgment task was presented to participants without any surrounding context. Because the passive construction is more pragmatically marked than the active, and requires more contextual support \citep{comrie1988}, this setting might have caused participants to rate passive sentences as less acceptable than their active counterparts even in the control verb classes.

Notably, although \textcolor{exp-th}{experiencer-theme} verbs are often thought of as passivizable, they showed a significant difference in passive drop from the other class of canonically passivizable verbs, the \textcolor{agt-pat}{agent-patient} verbs. Conversely, despite being reported as unpassivizable in the literature, the \textcolor{advantage}{advantage} and \textcolor{ooze}{ooze} verb classes did not differ in their passive drop from the canonically passivizable \textcolor{agt-pat}{agent-patient} class. As we showed in the previous section, these patterns are robust across participants. Overall, the pattern of gradient judgments that emerges from this experiment is considerably richer than that captured by the binary judgments from the linguistics articles that identified the verb classes in question, pointing to the value of formal acceptability judgment experiments for complex phenomena \citep{sprousealmeida2017}.

In summary, Experiment~1A demonstrated that some verbs in the verb classes being tested are degraded in the passive voice, and that the degree of unacceptability was graded across verbs. For a model to adequately approximate the human pattern of behaviour, then, it must capture the following patterns:

\begin{itemize}
    \item \textbf{Class-level exceptionality}: Some verbs classes (e.g. \textcolor{duration}{duration} verbs) exhibit passive drops that are significantly higher than the baseline passive drop expected of the canonically passivizable \textcolor{agt-pat}{agent-patient} verbs. 
    \item \textbf{Verb-level exceptionality}: Some verbs within a verb class (e.g. \textcolor{duration}{last} and \textcolor{price}{cost}) display passive drops that are significantly different from the passive drops of other verbs in the same class.
    \item \textbf{Gradience}: Acceptability is gradient rather than categorical. Verbs' passive drops span a broad range of values and do not obviously cluster into two groups (passivizable and unpassivizable).
\end{itemize}

\section{Experiment~1B: Comparing language model and human judgments}
In the previous section, we established that English speakers judge unpassivizability on a cline, rating some verbs such as \textcolor{price}{\textit{cost}} as highly unpassivizable, and other verbs such as \textcolor{agt-pat}{\textit{pushed}} as highly passivizable. In this section, we use a computational model to test if there is indirect evidence for the unpassivizability of these verbs. We do so by comparing the human judgments to those derived from a neural network language model. To adequately capture the indirect evidence available to humans, we train the language model on a corpus of 100 million words, comparable in size to the linguistic input available to English speakers by adolescence \citep{linzen2020,warstadtetal2023,wilcox2025}. 

What can we expect the results of this experiment to be? Some positive indications that exceptions can be learned from indirect evidence come from Bayesian modeling of the dative alternation \citep{perforsetal2010}. Exceptions to the dative alternation in English follow a similar pattern to exceptions to passivization. Most ditransitive verbs can occur in both the double object construction (e.g. \textit{Lucy gave Divya a bag}) and the prepositional dative construction (e.g. \textit{Lucy gave a bag to Divya}), but not all verbs can can appear in both constructions: \emph{donate}, for example, can only occur in the prepositional object construction (e.g. \textit{Lucy donated the car to Divya}, but \textit{*Lucy donated Divya the car}). \citet{perforsetal2010} find that a hierarchical Bayesian model can learn to identify verbs that participate in this alternation after exposure to a subset of the CHILDES child-directed speech corpus \citep{brown1973,macwhinney2000}. 

These results indicate that exceptions to an otherwise productive constraint can be learned from indirect evidence with a sufficiently powerful learning algorithm. However, it is unclear if such findings would extend to the language models we use in this work, which, like the majority of contemporary language models, are based on the transformer architecture \citep{vaswanietal2017}: these models may not have strong enough inductive biases to implement the explicit inference procedure implemented by a Bayesian model, and as such might not be sufficiently sensitive to indirect evidence. Indeed, there is evidence that transformers sometimes \textit{over-generalize}, for instance by translating English idioms like ``kick the bucket'' compositionally instead of treating such multi-word expressions as exceptions that should resist the compositional interpretation rule \citep{dankersetal2022}. Even if neural network language models are sensitive to indirect evidence, their weaker biases may lead them to require much more data than humans to learn effectively; quite generally, neural network models are less data-efficient learners than humans \citep{warstadtbowman2022}, and in practice are usually trained using large corpora that are vastly larger than the input available to humans \citep{linzen2020,frank2023bridging}. It is thus unclear how similar to humans a neural network will be if it is trained on a corpus that approximates the amount of access to the passive that a human might have.

\subsection{Model architecture}

The models we trained were based on the transformer architecture as implemented in \mbox{GPT-2} \citep{radfordetal2019}, specifically \mbox{GPT-2} small, which has 117M parameters. We used the Adam optimizer \citep{kingmaba2015} with an initial learning rate of 0.0006. The maximum input length was 512 tokens and the batch size was 16. Models were trained for up to 50 epochs, with early stopping if validation loss did not decrease for three consecutive evaluation steps.  The initial weights of the neural network and other aspects of training vary based on random seed, which could lead to different judgments on our test sentences. Such variability is particularly common for tests of linguistic generalization (e.g. \citealt{mccoy-etal-2020-berts}). We thus trained five different models with a different random seed each. 

We adopted the transformer architecture not only because it is the dominant language modeling architecture at the time of writing (e.g., \citealt{achiam2023gpt,touvron2023llama}), but also because there is reason to believe, based on prior work, that transformers may be able to learn exceptions to passivization: the original \mbox{GPT-2} models, albeit trained on a much larger corpus than our models, produced judgments that correlated well with human acceptability judgments of passive exceptions \citep{leonglinzen2023}. \mbox{GPT-2} is also sensitive to more general restrictions on English passives; \citet{warstadtetal2020} showed that \mbox{GPT-2} assigns lower scores to passive sentences containing intransitive verbs (e.g. \textit{Jeffrey’s sons are smiled by Tina’s supervisor}) than sentences containing transitive verbs (e.g. \textit{Jeffrey’s sons are insulted by Tina’s supervisor}), demonstrating that the model is sensitive to the fact that intransitive verbs cannot be passivized in English. Finally, \mbox{GPT-2} demonstrates sensitivity to exceptions to verb argument structure rules; for example, it can differentiate between verbs which do and do not participate in the dative alternation \citep{hawkinsetal2020}. That being said, since we train our models on substantially fewer words than were used to train OpenAI's \mbox{GPT-2}, it is an empirical question whether our models will behave similarly.

\subsection{Training corpus} \mbox{GPT-2} was trained on OpenAI's proprietary WebText corpus, which contains 40GB of data from the web---approximately 8B words, assuming an encoding such as UTF-8 with one byte per English character and an average of 5 character per word. By contrast, English-speaking children are exposed to 2--7M words per year \citep{gilkersonetal2017}, or 26M--91M words by the age of 13. As our goal is to determine what can be learned from the data available to humans, we trained our models using a significantly smaller training corpus than \citet{radfordetal2019}.
Rounding to the nearest order of magnitude, we trained our models on a corpus of 100M words. This corpus was a subset of OpenWebText \citep{gokaslancohen2019}, an open-source reproduction of the OpenAI Web Text corpus, which contains websites linked from Reddit with at least three upvotes. This selection method aims to choose a wide range of web text curated by humans. We recognize, of course, that this corpus is not entirely plausible as a representation of the input available to an English learner; future experiments using our methodology could use more cognitively plausible corpora as they become available \citep{wilcox2025}.

\subsection{Trigram model}

To what extent can human passivizability judgments be reduced to a consequence of the frequency of particular sequences of words (for example, ``was lasted by'' is likely to be a very low-frequency sequence)? To address this question, we also trained a trigram word-level language model on the same 100M word corpus we used for our transformers. A trigram model predicts the upcoming word from the most recent two words, based on simple computations related to the frequency of sequences of three or fewer words in the corpus; unlike transformers, it does not construct a semantic representation of words. We used Kneser-Ney smoothing as implemented in KenLM \citep{heafield2011}.

\subsection{Procedure}
We used a modified version of the targeted syntactic evaluation paradigm \citep{linzenetal2016,lauetal2017,marvinlinzen2018} to derive acceptability judgments from our language models. Many studies in this paradigm derive binary judgments by comparing the probabilities assigned by the model to a minimal pair of sentences, and taking the sentence with the higher probability to be more acceptable. Here, to capture gradient judgments, we obtained the log-probability score for each sentence, defined as the sum of the log-transformed probabilities of all of the tokens in the sentence, and computed the passive drop of each sentence pair by subtracting the score of the active sentence from the score of the passive one. As in the human case, this procedure controls for the lexical effects of the open-class words in each sentence pair.

In addition to the models' gradient passivization judgments, we also characterize the models' broader syntactic competence by deriving binary acceptability judgments for a range of grammatical phenomena included in the Benchmark of Linguistic Minimal Pairs for English (BLiMP; \citealt{warstadtetal2020}); these include, among others, subject-verb agreement, restrictions on the distribution of quantifiers like \textit{at least}, and, most pertinently, argument structure restrictions such as the unpassivizability of intransitive verbs.

\subsection{Materials}

We compare our models' passivizability judgments to human judgments for two sets of experimental materials. The first set of materials is the one we used in Experiment~1A, our human experiment above. The second is drawn from \citet{ambridgeetal2016}. Ambridge and colleagues collected judgments from 20 human participants for active and passive sentences containing 475 verbs. Their study differed from ours in a number of ways. First, Ambridge and colleagues covered a wide range of verbs with a smaller number of participants per item. They also included a large number of verbs that they expected to be passivizable, as well as some phrasal verbs (e.g. \textit{succumb to, look like}). This dataset thus provides a broader but potentially noisier sense of our models' passivizability judgments.

To quantify the variability in the \citeauthor{ambridgeetal2016} dataset that we can hope to account for with a computational model, we conducted a split-half reliability analysis using the same method we used for Experiment~1A. The results are reported in Table~\ref{tab:split-half-ambridge}.
\begin{table}[h]
    \centering
    \begin{tabular}{ll}
        \toprule
         \textbf{Verb class} & \textbf{Split-half reliability} \\
         \midrule
         All items & 0.73 \\
         \midrule
         Agent-Patient & 0.61 \\
         Theme-Experiencer & 0.41 \\
         Experiencer-Theme & 0.70 \\
         Other Passivizable & 0.69 \\
         Non-Passivizable &  0.92 \\
         \bottomrule
    \end{tabular}
    \caption{\textit{Spearman-Brown-corrected split-half reliability for each verb class in the \protect\citet{ambridgeetal2016} human experiment}. Human judgments in their experiment showed moderate reliability on all test items and moderate reliability within verb classes.}
    \label{tab:split-half-ambridge}
\end{table}
The reliability of human judgments in \citeauthor{ambridgeetal2016} is considerably lower than in our Experiment~1A (see Table \ref{tab:split-half}); consequently, there is a lower ceiling for the amount of variance that a computational model can account for.

\subsection{Results}

\begin{figure}[t]
    \centering
    \includegraphics[width=\textwidth]{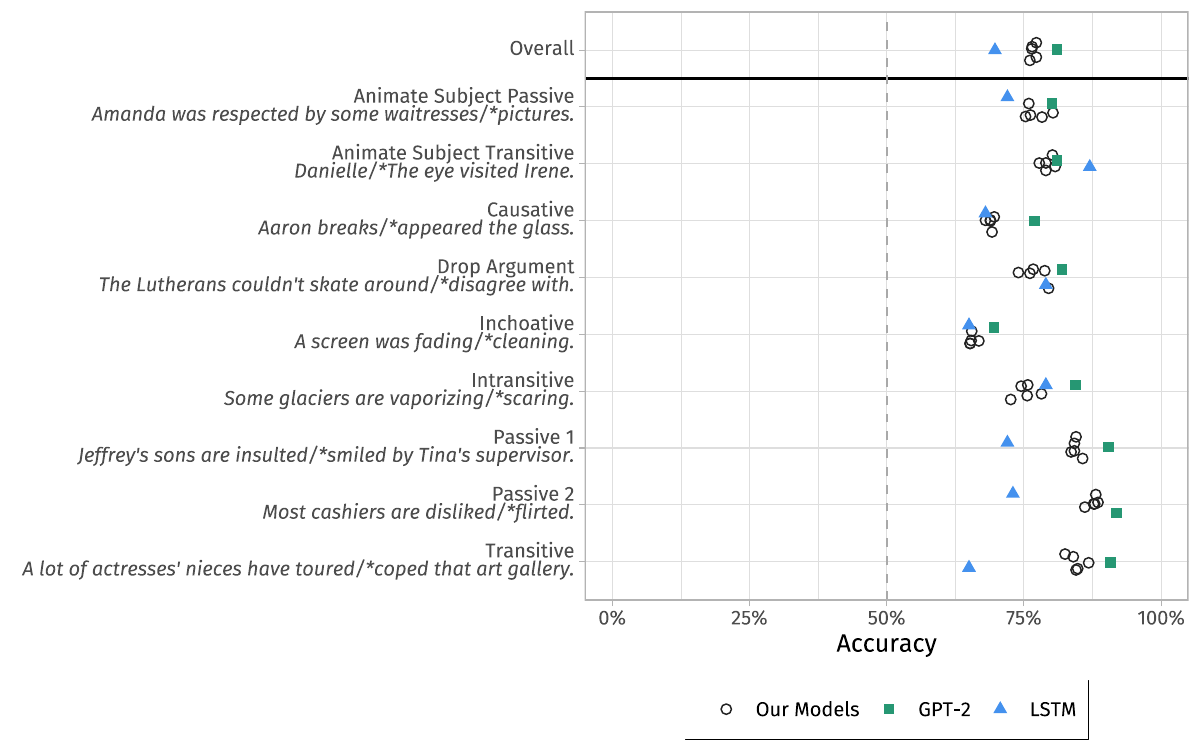}
    \caption{\textit{Accuracy of our models' acceptability judgments on the Benchmark of Linguistic Minimal Pairs (BLiMP, \protect\citealt{warstadtetal2020})}. We report overall accuracy over the entire benchmark, as well as detailed accuracy on the subset of constructions in BLiMP that are related to argument structure. Our models perform better than the \citet{gulordavaetal2018} LSTM model and marginally worse than OpenAI's \mbox{GPT-2}, which was trained on considerably more data. The dashed line indicates chance-level accuracy.}
    \label{fig:argument-structure-blimp}
\end{figure}

\begin{figure}[h]
    \centering
    \includegraphics[width=\linewidth]{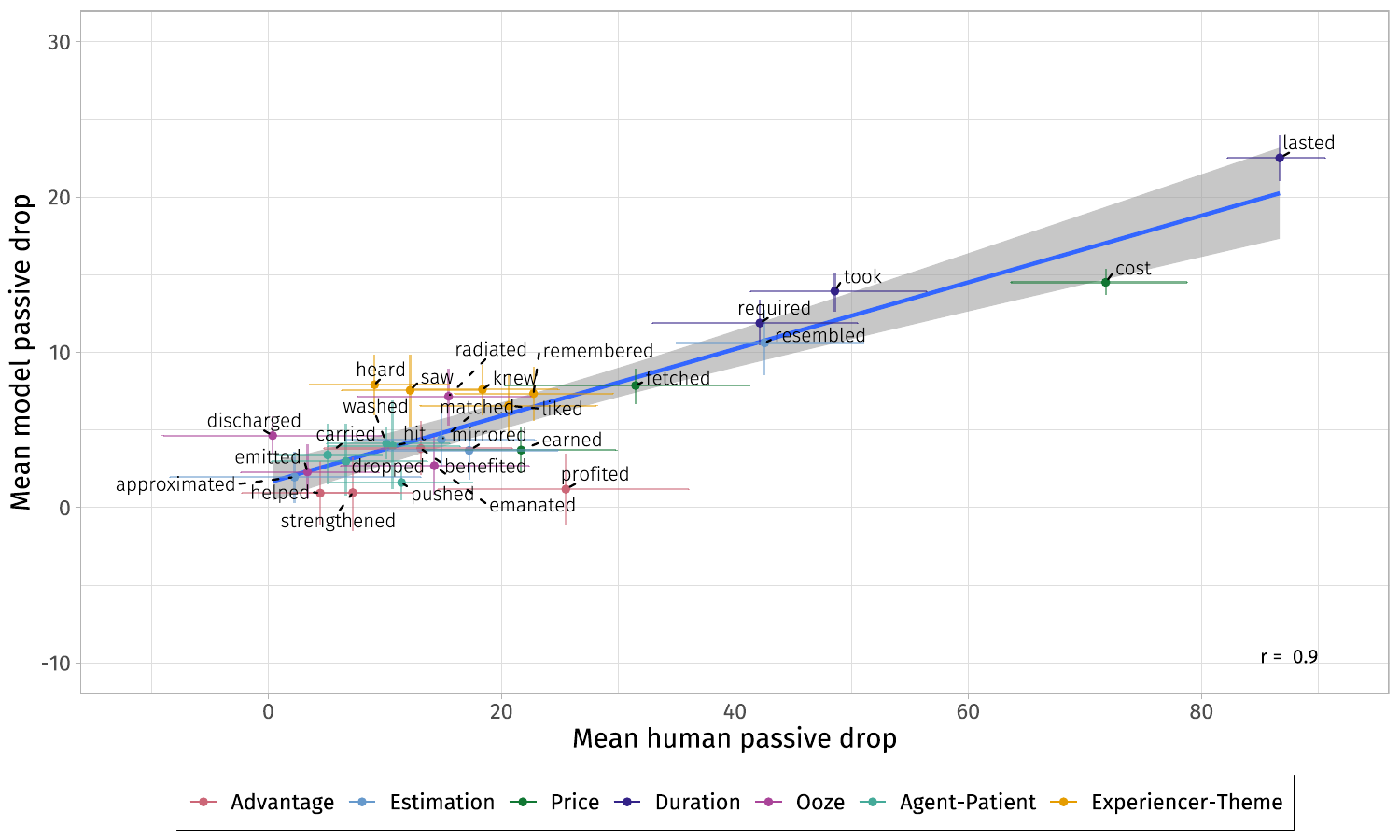}
    \caption{\textit{Passive drop in humans (Experiment~1A) vs. neural network language models}. At the verb level, the correlation between human and model judgments is $r = 0.9$. Each point represents the average passive drop of a verb in five sentence frames scored by five models. Horizontal  error bars indicate bootstrapped 95\% confidence intervals over human judgments for sentence frames; vertical error bars indicate bootstrapped 95\% confidence intervals over model judgments for sentence frames. All error bars are corrected for between-participant and between-model variance \citep{bakemanmcarthur1996}.}
    \label{fig:summary_scatter}
\end{figure}

\subsubsection{Overall syntactic competence} We first assess our models' broad syntactic competence. Figure~\ref{fig:argument-structure-blimp} shows our five models' performance on BLiMP, compared with OpenAI's \mbox{GPT-2} trained on the 40 GB WebText corpus, and the LSTM model trained by \citet{gulordavaetal2018} on 83 million words from the English Wikipedia (\mbox{GPT-2} and LSTM results were obtained from \protect\citealt{warstadtetal2020}). Across a wide variety of syntactic and semantic phenomena, our models preferred grammatical sentences to ungrammatical ones an average of 76.7\% of the time, suggesting that the sentence scores produced by our models are sensitive to syntactic and semantic constraints. The variability across the five models is fairly limited. Our models performed better than the \citet{gulordavaetal2018} LSTM model, although both types of model were trained on approximately the same amount of data. Finally, our models were only marginally worse than OpenAI's \mbox{GPT-2} despite being trained on 80 times less data.

Focusing on the subsets of BLiMP that test for models' sensitivity to restrictions on passivization, we find that all five models were able to determine, with an accuracy substantially higher than chance, that intransitive verbs are less acceptable than transitive verbs in the passive voice (see the \textsc{passive 1} and \textsc{passive 2} rows in Figure~\ref{fig:argument-structure-blimp}). They also showed a preference for animate over inanimate subjects in passive sentences (\textsc{animate subject passive} test in Figure~\ref{fig:argument-structure-blimp}). Overall, we conclude that our models gleaned from their training data considerable information about English grammar broadly and, more specifically, about the broad generalizations about the environments in which the passive construction is acceptable.

We do not report tests of the trigram model's syntactic competence: Many of the phenomena in the BLiMP dataset span more than three words and as such this model, which only has access to a window of three words, has no hope of capturing them; in fact, \citet{warstadtetal2020} show that even a 5-gram model is unable to capture most of the phenomena in BLiMP.

\subsubsection{Comparison to Experiment 1A}
We next compare the models' judgments for our test sentences to those of humans. Figure~\ref{fig:summary_scatter} graphs the models' average passive drop for each verb against the passive drop observed in our human experiment. We found a high Pearson's correlation coefficient of $r = 0.91$ between the human and model passive drop at the verb level. At the individual item level, the correlation was $r = 0.65$; while this correlation is still fairly high, it remains below the upper bound of $0.93$ derived from the split-half reliability analysis (Table~\ref{tab:split-half}). The models matched humans' judgments of \textbf{exceptionality} within verb classes: among verbs with similar meanings, the same verbs displayed high passive drops for across humans and models. For instance, for both humans and models, the passive drops of \textcolor{price}{\textit{earned}} and \textcolor{ooze}{\textit{discharged}} were low compared to other verbs in their respective classes. Likewise, like humans, our models predicted high passive drops for \textcolor{duration}{\textit{lasted}}, \textcolor{estimation}{\textit{resembled}} and \textcolor{price}{\textit{cost}}, compared to other verbs in the respective classes.  Finally, our models also largely matched human judgments of \textbf{gradience}: the sentence scores obtained from our models predict not only low and high passive drops, but also intermediate levels of passive drop in verbs such as \textcolor{duration}{\textit{took}} and \textcolor{duration}{\textit{required}}.

\begin{figure}[h]
    \centering 
    \includegraphics[width=\linewidth]{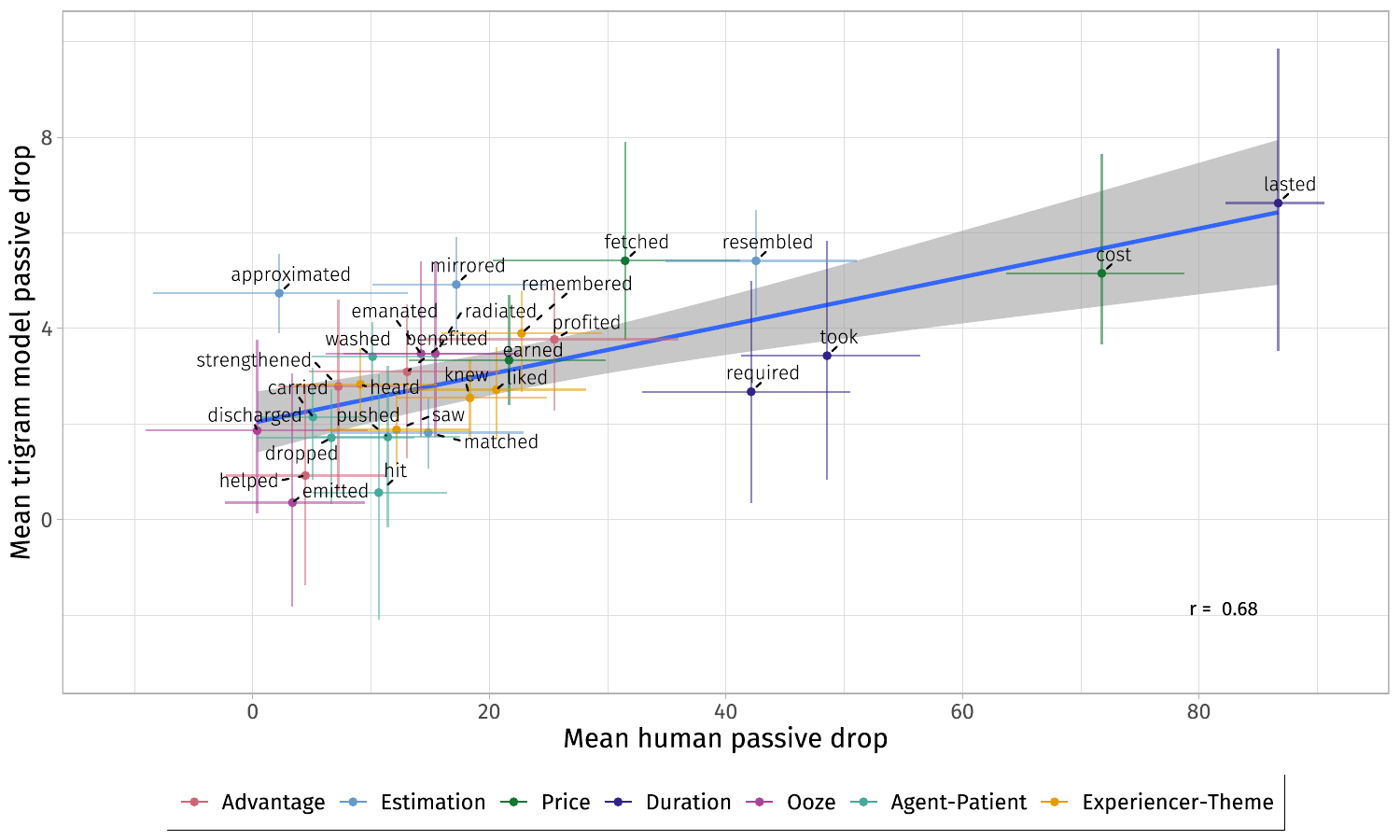}
    \caption{\textit{Passive drop in humans (Experiment~1A) vs. passive drop predicted by the trigram model estimated from the training corpus}.}
    \label{fig:trigram-results}
\end{figure}

Moving to the trigram model, we found a moderate-to-strong linear correlation with human passive drop at the verb level ($r = 0.68$;  Figure~\ref{fig:trigram-results}) and a moderate correlation at the item level ($r=0.34$). These results suggest that although some information about the passivizability of the verbs in our materials is recoverable from the frequency of short sequences of words, transformer models are able to capture more information about passivizability than can be attributed to trigram frequency alone. We attribute the moderately high performance of the trigram model to the relatively simple structure of the materials from the human experiments, which included low-probability trigrams such as \textit{was lasted by}; if the materials were modified to introduce additional words around the verb, for example \textit{was previously lasted in total by}, judgments derived from the trigram model would likely no longer be able to capture the unpassivizability of the verb.

A data contamination analysis found that one of our active test sentences, \textit{The journey lasted three days}, occurred verbatim in the training dataset, which would have inflated the probability assigned to the sequence and thereby the gap between the active and passive versions of this particular sentence. Because no other items were present in the training data, the effect of data contamination on our results is minimal.

\subsubsection{Exceptions to passivization: comparison to \citet{ambridgeetal2016}}
\begin{figure}[h]
    \centering
    \includegraphics[width=\linewidth]{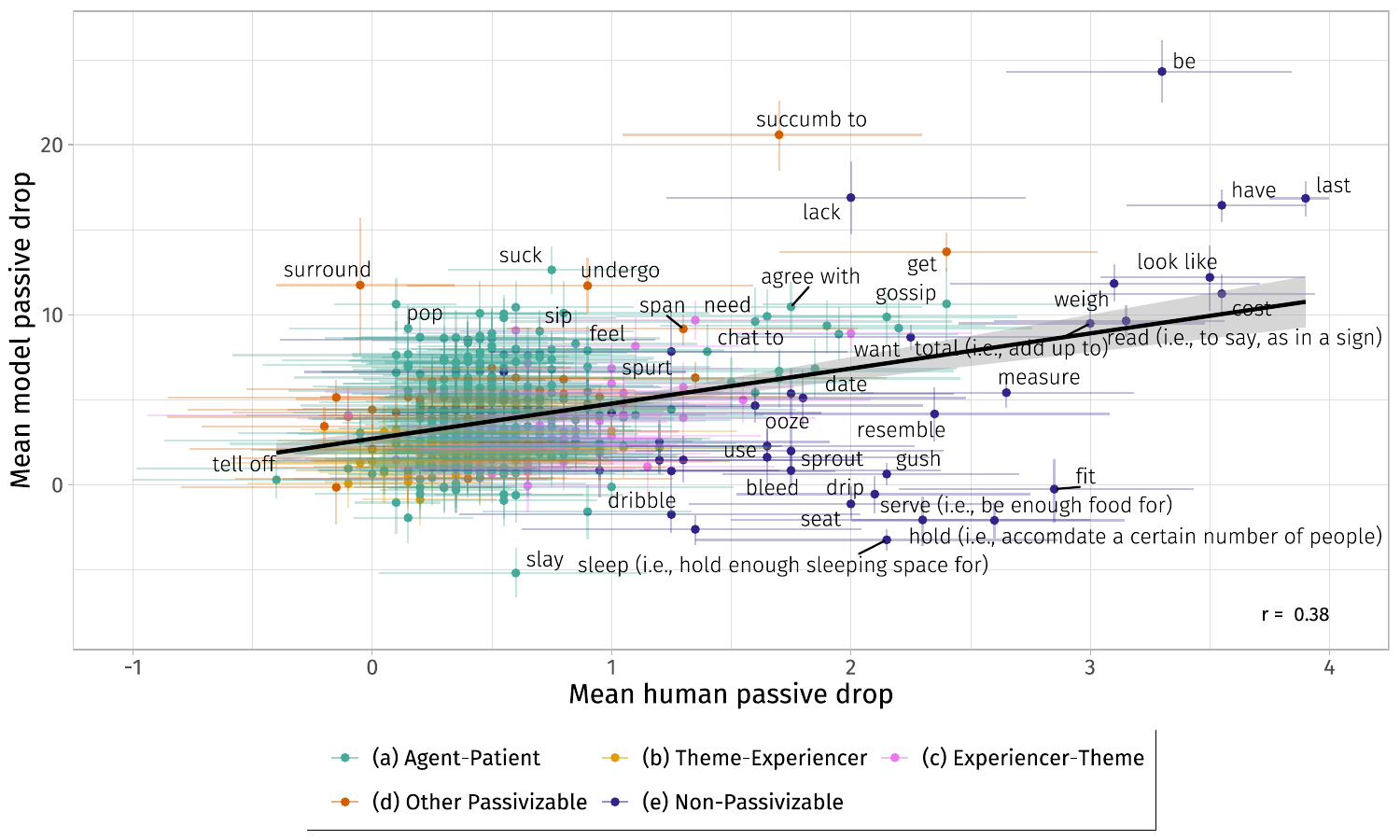}
    \caption{\textit{Passive drop in humans from \citet{ambridgeetal2016} vs. neural network language models.}}
    \label{fig:ambridge_scores}
\end{figure}

Figure~\ref{fig:ambridge_scores} shows the transformers' average passive drop for each verb against the passive drop calculated based on human scores reported by \citet{ambridgeetal2016}. While there was a positive linear correlation between human and model passive drop ($r = 0.42$ at the verb level, $r= 0.19$ at the item level), this correlation is substantially weaker than the correlation we observed for our stimuli from Experiment~1A. Qualitatively, the models predicted poorly the extent to which different verbs resist passivization: they overpredicted the unpassivizability of some verbs, such as \textit{serve} and \textit{fit}, and underpredicted the unpassivizability of others (e.g. \textit{last}, \textit{lack}).

We should emphasize that the correlation coefficients cannot be directly compared across Experiment~1A and \citet{ambridgeetal2016}, because, as we described above, the human judgments from  \citet{ambridgeetal2016} are much more variable, which leads to a lower upper bound on any model's potential correlations (this variability is reflected in the width of the horizontal error bars in Figure~\ref{fig:ambridge_scores}). That being said, the gap between the empirical correlation at the item level and the ceiling reflected by the split-half reliability  is larger for the \citet{ambridgeetal2016} materials ($0.20$ vs. $0.75$) than for Experiment~1A ($0.64$ vs. $0.93$).

The trigram model performed much worse than the transformers on the data provided by \citet{ambridgeetal2016}, with a \textit{negative} linear correlation between human and model passive drop at the verb level ($r=-0.27$). We hypothesize that this behavior arose from the use of proper nouns (e.g. \textit{Marge}, \textit{Wendy}) in the test sentences: approximately 32\% of the tokens in this dataset were out of vocabulary for the trigram model. Such data sparsity-related issues, in combination with the short length of the test sentences (e.g. \textit{Bob eluded Wendy}), likely affected the trigram model's accuracy.

\subsection{Discussion}
Broadly, transformer language models captured human patterns of verb passivizability well when evaluated on the test items from Experiment~1A. Qualitatively, our models showed gradient judgments that varied across individual verbs in each verb class as well as across verb classes. Quantitatively, the transformers' judgments were strongly correlated with human judgments ($r=0.91$). We note, however, that the highly passivizable verbs in our test set were all relatively frequent, and likely occurred in the passive a substantial number of times in the training corpus. As such, none of them posed the problem we referred to in the introduction as Baker's paradox, which would be posed by an infrequent transitive verb such as \textit{defenestrate}.

In comparison with the transformers, judgments based on a simpler frequency-based trigram language were only moderately correlated with human judgments. In other words, the trigram model was unable to predict the human results of Experiment~1A as effectively as the transformers, even though our test sentences were structurally simple, and did not have adverbs that intervened between the auxiliary and the verb, such that `was/were [VERBed] by' was a contiguous sequence.

Evaluating our models against the materials of \citet{ambridgeetal2016}, in contrast, showed a discrepancy between our language models' and humans' passivizability judgments that was not present in our data. Although positive, the correlation between model and human judgments of passivizability was weak, and our models predicted the passive drop of non-passivizable verbs more poorly than verbs in other classes. Some of this difference in performance can be attributed to the large variance in human judgments. However, a further study with higher power comparing human and model judgments on more verbs might point to divergences between model and human judgments.

\section{Experiment~2: What indirect evidence do models use to learn restrictions on passivization?}
In the previous sections, we showed that 100 million words of English text provide sufficient evidence for a transformer language model to produce judgments of passive exceptions that align to a significant extent, though not fully, with those of humans. In the following sections, we turn to our second research question: which aspects of the linguistic input serve as evidence for models to learn these patterns? To answer this question, we take inspiration from work that has used controlled interventions on a model's training corpus to draw causal links between aspects of the training data and the model's behavior \citep{jumeletetal2021,misramahowald2024,patiletal2024,weietal2021}. This approach manipulates particular sources of evidence in the training corpus and compares models trained on the original dataset with models trained on the modified dataset; for example, to assess the causal effect of verb frequency on a language model's subject-verb agreement prediction accuracy, \citet{weietal2021} removed an increasing number of occurrences of the verb from the corpus, retrained the model, and measured the resulting change in its accuracy on this task.

We apply this method to test two hypotheses proposed in the literature as to the evidence that supports humans' acquisition of passive exceptions: the \emph{entrenchment hypothesis} \citep{brainebrooks1995,demuth2011,theakston2004} and the \emph{affectedness hypothesis} \citep{ambridgeetal2016,darmasetiyawanetal2022,messengeretal2012,pinker1989}. We tested each hypothesis by first altering or removing elements of the corpus that are crucial for learning under each hypothesis and retraining models on these modified corpora. If models trained on a modified corpus consistently differ from models trained on the original corpus in their acceptability judgments on passive exceptions, then we can attribute the change in behavior to the particular intervention that we made on the training data.

\section{Experiment~2A: Testing the entrenchment hypothesis}

The first hypothesis we tested is the \emph{entrenchment hypothesis} \citep{ambridgeetal2015,regiergahl2004,theakston2004}. According to this hypothesis, learners track the frequency of verbs in particular constructions \citep{gordonchafetz1990}, and if that verb never appears in a particular context but does appear with substantial frequency in other contexts, they conclude the verb is ungrammatical in that context. In the passivization case, the relevant factor is the relative frequency of the active compared to the passive: as learners are exposed to more and more occurrences of a verb in the active but not in the passive, they gradually conclude that the passive form is not just rare but ungrammatical. While the classic entrenchment hypothesis assumes the unacceptable form has to be completely absent from the corpus, here we test a softer, gradient version of the hypothesis, where different degrees of frequency asymmetries between the active and the passive forms of a verb result in gradient levels of acceptability for the passive form.

To test this hypothesis, we identified verbs which appeared relatively frequently in the active and highly infrequently in the passive. We chose the verbs \textcolor{duration}{\textit{last}}, \textcolor{price}{\textit{cost}} and \textcolor{estimation}{\textit{resemble}} as our \textsc{target verbs}, that is, the verbs whose relative frequencies we would try to emulate. All three verbs were highly unpassivizable, and had large \textit{active-passive ratios} (henceforth A/P ratio); they occurred much more frequently in the active compared to the passive. We then chose \textsc{mutating verbs}, whose frequency we would modify in the training corpus. These verbs were drawn from the highly passivizable \textcolor{agt-pat}{agent-patient} class. We computed the A/P ratio of each target verb, then matched the mutating verb's A/P ratio to the target verb's A/P ratio by removing as many passive sentences with that verb as necessary (and in some cases removing a small number of active sentences to match the ratio more closely).

The main motivation for this design is as follows: if A/P ratio serves as a cue for passivizability, we expect this intervention to cause the originally passivizable \textsc{mutating verb} to decrease in passivizability. In fact, if the \emph{only} cue to a verb's passivizability is its relative frequency of occurrence in the active voice compared to the passive, then we expect the \textsc{mutating verb} would become just as unpassivizable as the \textsc{target verb}.

\subsection{Estimating corpus frequencies}
\label{sec:exp2a-procedure}
\begin{figure}[h]
    \centering
    \includegraphics[width=.8\textwidth]{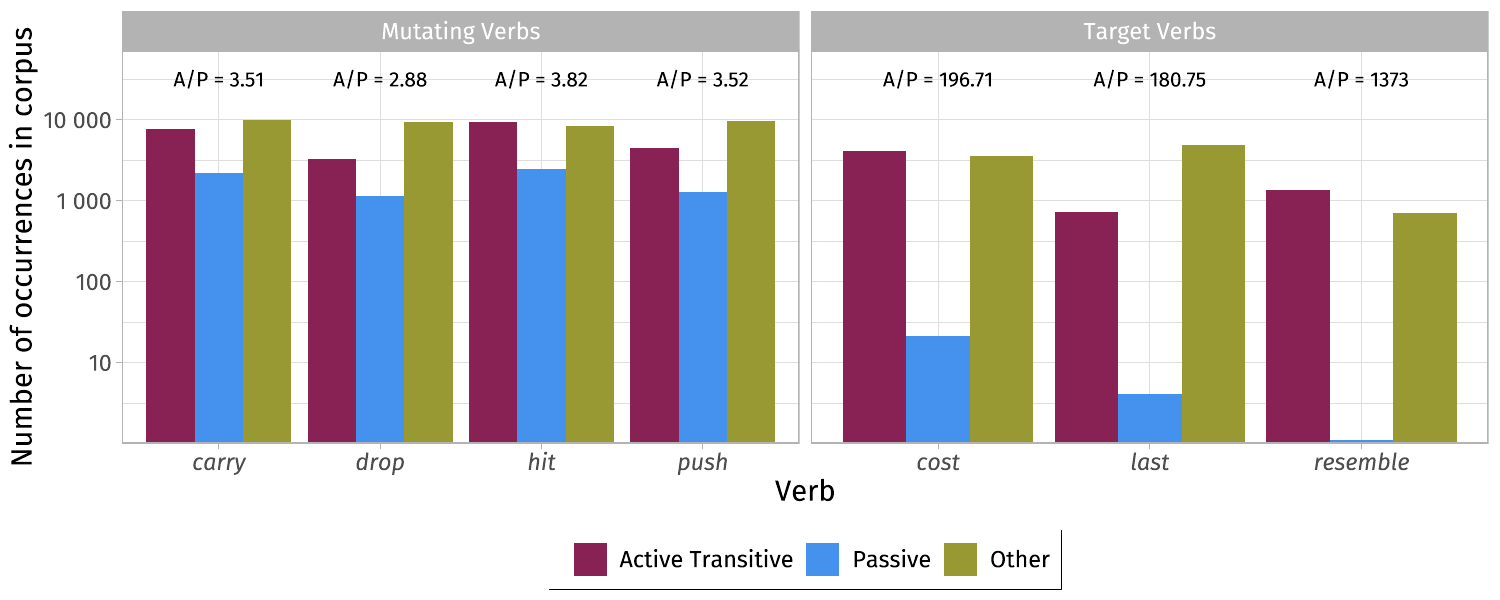}
    \caption{\textit{Frequency of occurrence of mutating and target verbs in the original corpus}. We include sentences that were parsed as transitive and active in our count of active sentences, and sentences parsed as passive in our count of passive sentences. All other sentences were labeled \textsc{other}.}
    \label{fig:verb-frequency}
\end{figure}

We used \texttt{spaCy}'s \texttt{en\_core\_web\_trf} pipeline \citep{honnibaletal2020} to obtain dependency parses for each sentence in the training corpus. We then counted the number of times our mutating and target verbs were used in transitive active sentences and passive sentences. Specifically, sentences where the verb had a dependency edge to a passive auxiliary (\texttt{auxpass}), a passive nominal subject (\texttt{nsubjpass}) or a passive clausal subject (\texttt{csubjpass}) were classified as \textsc{passive} sentences, while sentences with a direct object (\texttt{dobj}) or a clausal complement (\texttt{ccomp}) dependency edge from the verb were classified as \textsc{active} sentences. All other sentences were classified as \textsc{other}. 
The results of this analysis are shown in Figure~\ref{fig:verb-frequency}.

The \textsc{active} category only included sentences where the verb appeared in an unambiguously transitive frame, since such sentences expressed events that could also be expressed in the passive voice. Sentences with other syntactic structures, or that the filter failed to identify as active transitive, such as those in (\nextx) for \textcolor{agt-pat}{\textit{drop}}, were classified as \textsc{other} and excluded from analysis:
\pex
\a Realtors believe home resales, which dropped in September, peaked in July and August.
\a I apologize for that pun, but it’s definitely not worse than the ones Arnold Schwarzenegger drops.
\xe

\noindent This filtering mechanism prioritized precision over recall; that is, we prioritized the accuracy of the classification of a sentence as \textsc{active} or \textsc{passive}, potentially at the expense of classifying a larger number sentences as \textsc{other}. To assess the success of this strategy for the \textsc{passive} class, we used a regular expression to extract from the corpus 500 strings with auxiliaries such as \textit{had been} and \textit{was} followed by a past participle within the same paragraph. We found that only 24 (or 4.9\%) of these strings were not classified as passives by our filter (binomial 95\% confidence interval: 3.1--6.8\%), indicating that recall was only minimally affected. We additionally verified the precision of the filtering mechanism by hand-checking 500 sentences sentences that were labelled as passive sentences. We found that only 10 out of the 500 sentences (2.0\%) were misidentified as passive (binomial 95\% confidence interval: 0.9--3.3\%), indicating high precision. Examples of misparsed passive sentences for the verb \textit{last} are given in \ref{sec:errors}.
\begin{figure}[h]
    \centering
    \includegraphics[width=0.9\linewidth]{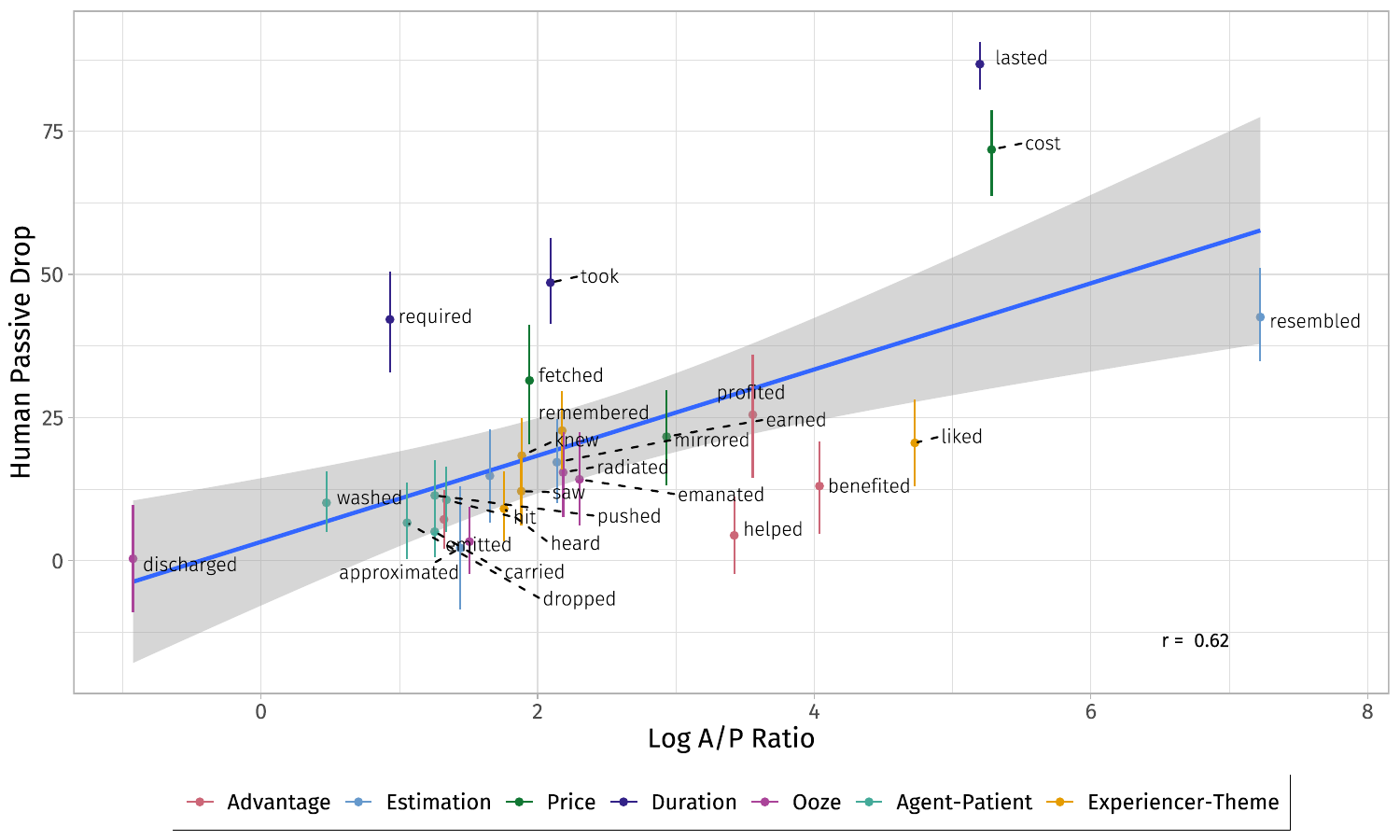}
    \caption{\textit{Correlation between a verb's relative corpus frequency in the active vs. passive and human passive drop.}}
    \label{fig:exp2a-freq-human}
\end{figure}

\subsubsection{Does relative corpus frequency predict human judgments?}
\label{sec:corpus-frequencies}
Before we discuss the corpus intervention experiment, we test if a verb's A/P ratio on its own can predict human passivizability judgments, as predicted by the entrenchment hypothesis. We find only partial support for this hypothesis (Figure~\ref{fig:verb-frequency}). Consistent with the hypothesis, all four mutating verbs, which had relatively low A/P ratios (2.88--3.82), were judged as highly passivizable, and all three target verbs, which had high A/P ratios, also displayed high passive drops in the human experiments. At the same time, the target verb with the highest A/P ratio, \textcolor{estimation}{\textit{resemble}}---which we expect to have the largest passive drop if corpus frequency alone drives passivizability---was instead judged by both our models and human participants as more passivizable than the other two target verbs with lower A/P ratios.

Moving to the full set of verbs, we observe a moderate positive correlation ($r=0.62$) between log-transformed A/P ratio and human passive drop (Figure~\ref{fig:exp2a-freq-human}). This correlation is similar to the one we observed between the human judgments and those derived from the trigram model, which similarly reflects count-based information. The substantial gap between this correlation coefficient and the one obtained by the transformers ($r=0.90$) suggests that frequency asymmetries alone do not fully account  for transformers' ability to predict human judgments; in particular, frequency does not explain patterns at the verb class level---it underpredicts the passive drop associated with all three \textcolor{duration}{duration} verbs while overestimating the passive drop associated with \textcolor{advantage}{advantage} verbs.

\subsection{Corpus intervention procedure}

\begin{figure}[h]
    \centering
    \includegraphics[width=\textwidth]{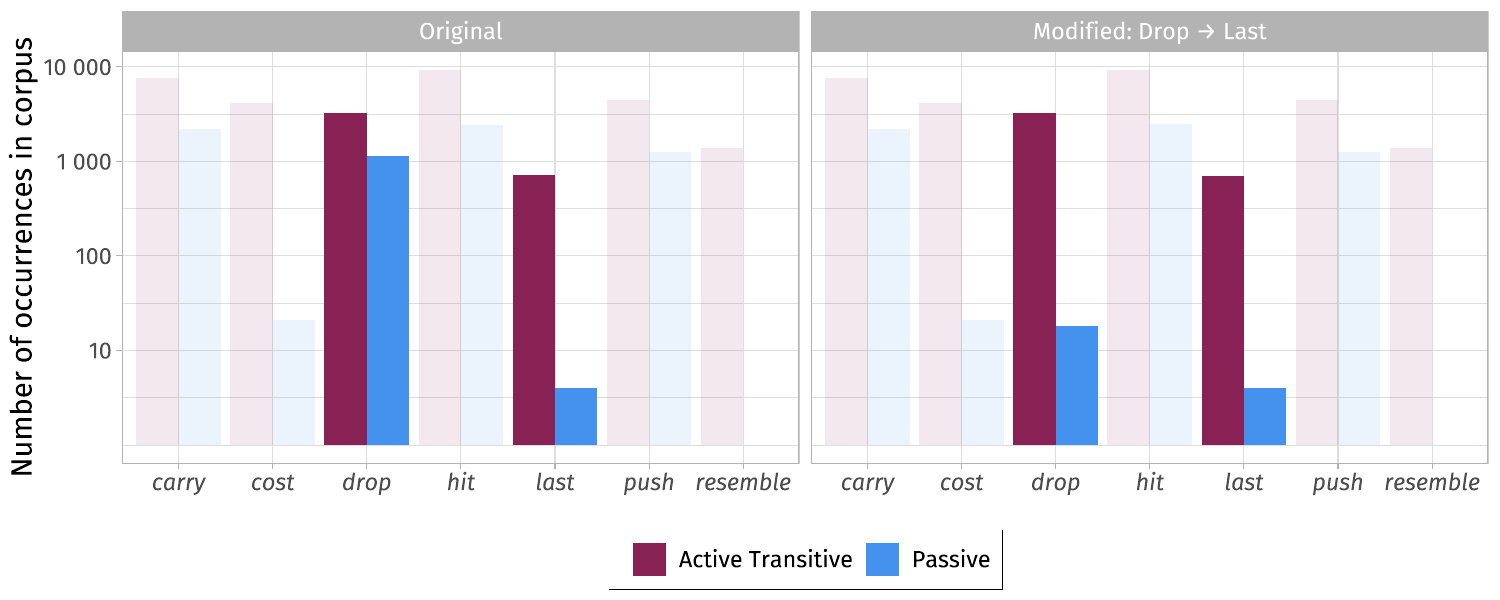}
    \caption{\textit{Corpus statistics before and after the intervention used for Experiment 2A. In this example, the mutating verb is \textcolor{agt-pat}{\textit{drop}} and the target verb is \textcolor{duration}{\textit{last}}}. The frequency of the mutating verb \textcolor{agt-pat}{\textit{drop}} is decreased in both the active and passive to match the relative frequency of the target verb \textcolor{duration}{\textit{last}}. All other verbs do not undergo any change.}
    \label{fig:corpus-comparison}
\end{figure}

We created modified corpora for all combinations of the four mutating and three target verbs (a total of 12 verb pairs), as follows. In each corpus, we used the A/P ratio of the target verb $A/P_{target}$ to obtain the corresponding number of occurrences of the mutating verb in the active and passive that should be kept in the modified corpus so that, keeping as many active sentences as possible, we had $A/P_{mutating} \approx A/P_{target}$. For example, in the original corpus the mutating verb \textcolor{agt-pat}{\textit{drop}} occurred 3279 times in transitive active sentences and 1146 times in the passive ($A/P_{drop} = 2.88$), and the target verb \textcolor{duration}{\textit{last}} occurred 723 times in transitive active sentences and four times in the passive ($A/P_{last}= 180.75$). To match the A/P ratios of the two verbs, we randomly chose 3253 active occurrences and 18 passive occurrences of \textcolor{agt-pat}{\textit{drop}} in the training corpus and discarded all other active and passive occurrences of \textcolor{agt-pat}{\textit{drop}}, such that $A/P_{drop}\approx180.75$ in the modified corpus. Figure~\ref{fig:corpus-comparison} illustrates the distribution of verbs in the the training corpora before and after we performed the intervention for this particular verb pair (\textcolor{duration}{\textit{last}} and \textcolor{agt-pat}{\textit{drop}}). Raw counts for each of the 12 verb pairs are available in \ref{sec:raw_counts}.

We trained five models, each with a different random seed, on each of the 12 modified corpora, using the same training procedure outlined in Experiment~1B. We then obtained acceptability judgments from these models and compared them to the judgments of the models trained on the original corpus in Experiment~1B.

If a verb's A/P ratio significantly affects its passivizability, then we expect the mutating verb to be judged as less passivizable (i.e. be given a higher passive drop) by models trained on the modified corpora compared to models trained on the original corpus. We expect the passive drops of all verbs other than the mutating verb to remain the same.
\subsection{Results}

\begin{figure}[h]
    \centering
\includegraphics[width=\textwidth]{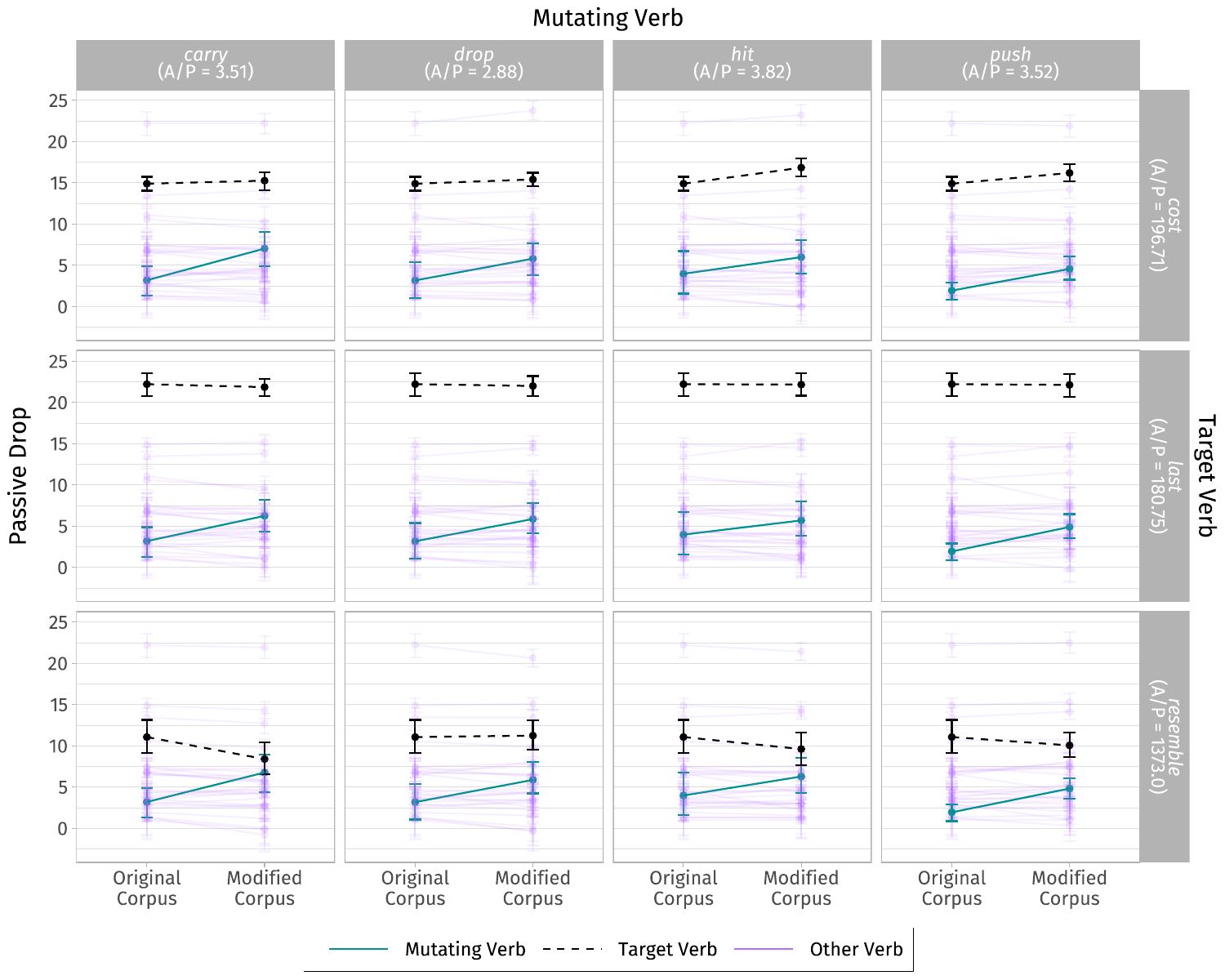}
    \caption{\textit{Change in passive drop as a result of training on a modified corpus with a higher active-to-passive frequency ratio for the mutating verb}. Passive drop of mutating verbs (in green) increases when their distribution is modified, but only reaches the same level as the target verb when that verb is \textit{resemble}.}
    \label{fig:freq-change-all}
\end{figure}

The results of Experiment~2A are shown in Figure~\ref{fig:freq-change-all}. Across all target and mutating verb pairs, the mutating verb showed an increase in passive drop as a result of the intervention. To test whether the intervention resulted in a larger increase in the passive drop of the mutating verb compared to any overall difference in passive drop across all verbs, we fit two linear mixed-effects models: one predicting \textsc{passive drop} as a function of \textsc{training corpus}, with by-\textsc{model}, by-\textsc{verb}, by-\textsc{verb class} and by-\textsc{frame} random intercepts; and another model that included all of these predictors as well as an additional fixed effect indicating whether the verb is \textsc{mutating} in the corpus. A likelihood-ratio test indicated that mutating verbs showed increases in passive drop not shown by other verbs ($\chi^2(1) =141.2 , p< 0.001 $).

The intervention did not significantly affect the passive drop for verbs that did not undergo mutation (black and purple lines in Figure~\ref{fig:freq-change-all}). To determine that, we fit two linear mixed-effects models for the non-mutating verbs: one predicting \textsc{passive drop} as a function of \textsc{verb} and \textsc{verb class}, with random intercepts for \textsc{frame} and \textsc{seed}; and another model that further included a fixed effect indicating whether the corpus was modified. A likelihood-ratio test indicated that there was no significant effect of modifying the corpus ($\chi^2(1) =0.52, p = 0.478$).

Although each mutating verb's passive drop increased when it was mutated, in the majority of cases the intervention did not cause the mutating verb to become as unpassivizable as the target verb. Across all verb pairs, the mutating verbs in all twelve conditions showed an increase in passive drop of 2 to 4 points after intervention, regardless of the passive drop of the target verb. In the one case where the two verbs' passive drops converged after the intervention (for the mutating verb \textit{hit} and the target verb \textit{resemble}), this was because the target verb's passive drop unexpectedly decreased. Overall, while the intervention reliably increased the mutating verb's passive drop, in general it did not fully close the gap between the mutating verb and the target verb. 

\subsection{Discussion}

In this experiment, we found that increasing the A/P ratio of a verb consistently led language models to judge the verb as less passivizable. This suggests that the A/P ratio is one potential source of evidence by which models learn whether a verb is or is not passivizable. This is most likely not models' \textit{only} source of evidence for passivizability, however: if it were, we would expect each mutating verb to behave exactly like the target verb after our intervention. That was not the case: although the mutating verbs occurred in the same A/P ratio as the target verbs, the passive drop of these mutating verbs did not increase to the level of unpassivizability of their respective target verbs. This source of evidence is also most likely not the only one that humans use to make passivizability judgments: if it were, we would expect corpus frequencies to be just as predictive of human passive drop as were the transformers' predictions, which was not the case (Section~\ref{sec:corpus-frequencies}).

We note that although we reduced the number of passive sentences that our mutating verbs appeared in, we did not eliminate all instances of the passive---each verb appeared at least once in the passive in each training corpus (see \ref{sec:raw_counts} for details). This meant that we tested a weaker version of the entrenchment hypothesis: while the model was exposed to a smaller number of passives of the mutating verb than in the original corpus, this number was not zero. The reason we matched the A/P ratio of an existing verb was to test the hypothesis that that would lead to convergence between the passive drop of the mutating verb and the target verb. We also reasoned that this weaker version of the entrenchment hypothesis mirrors natural language acquisition more faithfully: speech errors and misparses may well lead a learner to conclude that a particular verb was observed in the passive even when the sentence was not intended by the speaker to be a passive one (indeed, some of the sentences that our parser classified as passives may have been misparses). In the novel verb experiment below (Experiment 3) we test a stronger version of the entrenchment hypothesis, in that the modified corpora contain no passive occurrences at all.

By removing many or most of the occurrences of the mutating verb in the passive forms, we naturally also reduced the total number of passives in the corpus, as well as the total number of words in the corpus. While in principle either of these factors could confound our results, we suspect that the effect is negligible: we eliminated a few thousand sentences out of many millions of sentences in the original corpus, of which hundreds of thousands were passive sentences. Indeed, empirically we do not see changes in the passive drops of verbs other than the mutating verb, suggesting that the manipulation did not materially affect the models' ability to learn the passive. That being said, we note this as a potential limitation of the filtering methodology we use; in cases where filtering is more aggressive, it may be appropriate to replace the sentences that were filtered with new sentences from the corpus that are not relevant to the hypothesis, so that the total number of sentences is matched between the original and modified corpus \citep{misramahowald2024}.

\section{Experiment~2B: Testing the affectedness hypothesis}

We next consider the hypothesis, proposed by Pinker (\citeyear{pinkeretal1987,pinker1989}), that the passivizability of a verb can be predicted from its lexical semantics, specifically the \emph{affectedness} (or lack thereof) of the theme argument of the verb. In the sentence \textit{The apple was eaten by the boy}, for example, the by-object \textit{the boy} can be said to \textit{affect} the subject of the passive (e.g. \textit{the apple}), in that it causes a change of state, location, or existence to the subject. Support for this hypothesis comes, for instance, from \citet{ambridgeetal2016}, where participants rated verbs on a series of proxies for affectedness in sentences where the arguments were not explicitly specified; for the sentence \textit{A likes B}, for example, participants might be asked to rate the statements \textit{A is responsible}, or \textit{A is doing something to B}. By aggregating these proxies, \citeauthor{ambridgeetal2016} computed a measure of the verb's prototypical affectedness (that is, its affectedness with the prototypical arguments inferred by participants when arguments are not explicitly mentioned). They found a positive correlation between a verb's prototypical affectedness and participants' acceptability judgments for sentences where it was used in the passive. Similar effects of affectedness on passivizability have been documented across languages \citep{ambridgeetal2023,aryawibawaambridge2018,bidgoodetal2020,darmasetiyawanetal2022,liuambridge2021}.

Experiment~2B tests the hypothesis that affectedness causally explains language models' passivizability judgments. We modify the training corpus by intervening on the same pairs of verbs as in Experiment~2A, but in contrast with Experiment~2A, here we use the highly-unpassivizable verb as the \textsc{mutating verb}, and the highly-passivizable verb as the \textsc{target verb}.

In transformer language models, as in most neural networks, lexical semantics is encoded by word vector representations (embeddings) that reflect the contexts in which the word appeared in the training corpus. Because the individual dimensions of this vector are not interpretable, it is not straightforward to modulate the degree of affectedness of a verb by intervening on its embedding. Instead, we aim to shift the semantic representation of the mutating verb in a more or less affected direction by changing the contexts in which this verb appeared. We do so by placing the mutating verb in active sentences that originally contained the target verb; this allows the mutating verb, which is in general highly unpassivizable, to co-occur in the active with the agent-like subjects and patient-like objects that are normally associated with the target verb (see examples in the next section). Crucially, we only intervened on active sentences: we did not manipulate the passive sentences containing the mutating verb, or add new passive sentences to the corpus.

\subsection{Procedure}

For each pair of mutating and target verbs, we randomly selected a portion of active transitive sentences (identified using the procedure outlined in Section~\ref{sec:exp2a-procedure}) in the training corpus containing the target verb and  replaced the target verb with the mutating verb (e.g. \textit{dropping} $\rightarrow$ \textit{lasting}; \textit{dropped} $\rightarrow$ \textit{lasted}), thus making the mutating verb appear in contexts that previously contained the target verb while making minimal changes to the syntactic environments in which the verb occurs. Examples of sentences that underwent intervention for the experiment where the mutating verb was \textit{last} and the target verb was \textit{drop} are given in (\nextx):

\pex 
\a 	You know, people are always \sout{dropping} \uline{lasting} off samples of gluten-free products at our office.

\a The pilot, worried the bomb might break loose from the damaged plane, \sout{dropped} \uline{lasted} it into the water outside of Savannah, Ga. near Wassaw Sound.

\xe
\noindent Placing the mutating verb in environments that previously contained the target verb caused the mutating verb to co-occur with some of the subjects and objects that previously co-occurred with the target verb, which in turn caused the distribution of its arguments to more closely resemble that of the target verb. Since we only accounted for syntax, phrasal verbs and idiomatic expressions were also affected by this process, as illustrated in (\lastx a).

Note that unlike in Experiment~2A, in which our intervention changed the distribution of the mutating verb but not the target verb in the training corpus, the intervention in Experiment~2B affected the both target and mutating verbs. The mutating verb appeared in the sentences it originally occurred in as well as the newly mutated sentences, while the target verb was seen less frequently in the corpus, since some of its occurrences were mutated. For this reason, the target verb's passive drop after the intervention may not be representative of its passive drop before the intervention.

The entrenchment hypothesis, for which we found support in Experiment~2A, predicts that increasing the A/P ratio of a verb would make the mutating verb less passivizable. To counteract this, we removed the same number of active transitive sentences that originally contained the mutating verb. Thus, in both corpora, the mutating verb appears the same number of times in active transitive sentences, but in different contexts.

We created two such modified corpora for each pair of mutating and target verbs (12 pairs in total), varying the proportion of a verb's occurrences that were altered. In one corpus, we replaced a moderate amount (30\%) of the active transitive sentences that originally contained the mutating verb. In the other, we replaced a large proportion (70\%) of the active transitive sentences containing the mutating verb. Our particular choice of how many sentences to replace (either 30\% or 70\%) was arbitrary, and future work could consider modulating or interpolating this proportion, and thereby modulating the similarity between the distribution of the mutating verb and target verb. We trained five models on each modified corpus, and obtained acceptability judgments from those models following the procedure outlined in Experiment~1B.

\subsection{Results}
\begin{figure}[t]
    \centering    
    \includegraphics[width=\linewidth]{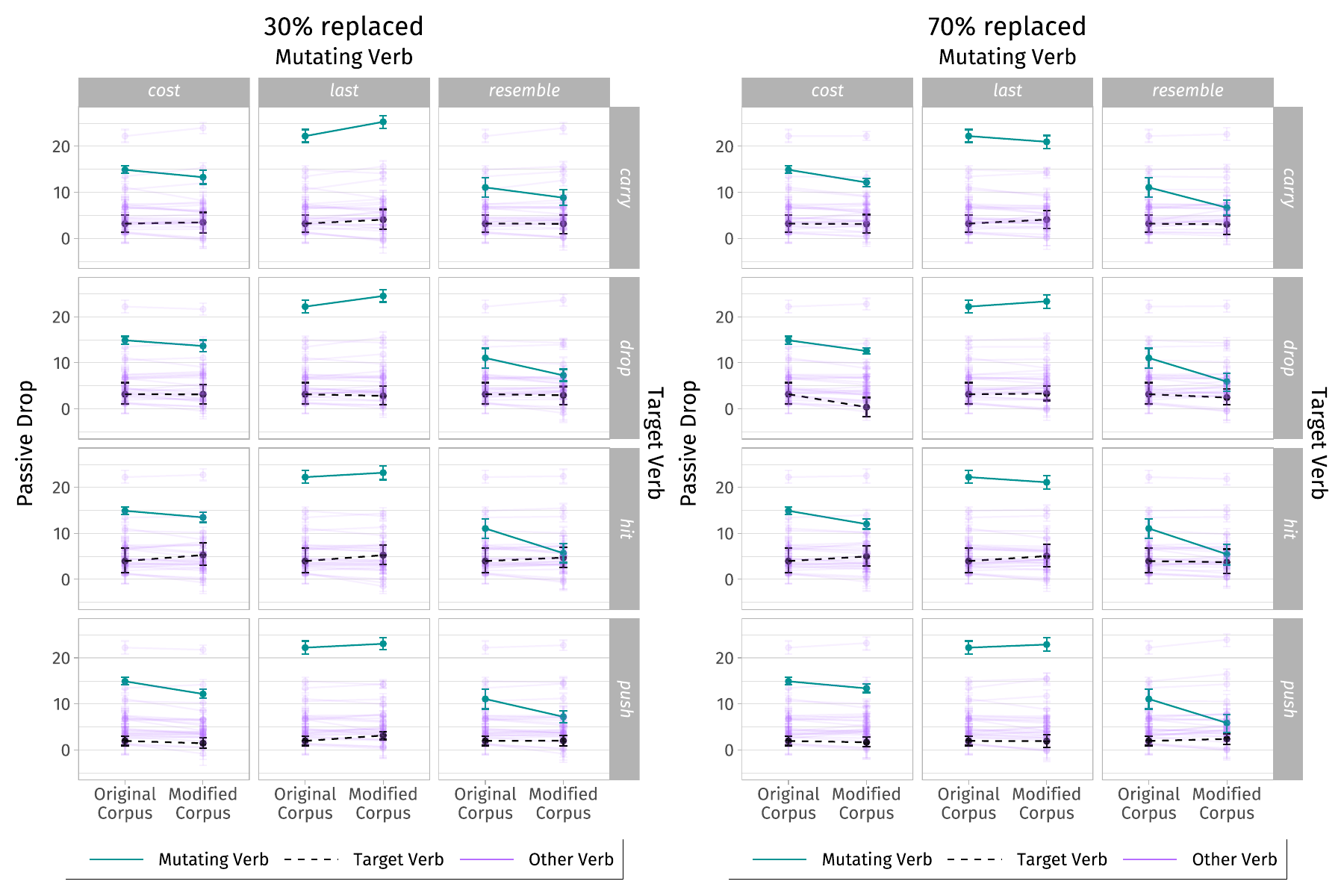}
    \caption{\textit{Change in passive drop as a result of training on data with target verb-like arguments}. Mutating verbs which were altered to appear with the arguments of target verbs in the training data did not show a consistent decrease in passive drop.}
    \label{fig:swap-change-all}
\end{figure}

The results of Experiment~2B are shown in Figure~\ref{fig:swap-change-all}. When 30\% of the mutating verb was replaced, the passive drop of verbs with no change across training corpora (``other" verbs, purple lines in Figure~\ref{fig:swap-change-all}) decreased by a mean of 0.19 points, while the passive drop of mutating verbs decreased by a mean of 1.25 points and the passive drop of target verbs increased by 0.38 points. When 70\% of the mutating verb was replaced, the passive drop of `other' verbs decreased by a mean of 0.16 points, while the passive drop of mutating verbs decreased by a mean of 2.53 points, and the passive drop of target verbs decreased by a mean of 0.06 points.

Non-mutating verbs did not show a significant change in passive drop  across training corpora. To verify this, we compared two linear mixed-effects models predicting the passive drop of non-mutating verbs. The full model used \textsc{corpus type}, \textsc{verb}, and \textsc{verb class} as main effects and random intercepts for \textsc{frame} and \textsc{model seed}, while the reduced model did not include \textsc{corpus type} as a fixed effect. A likelihood ratio test showed that including corpus type did not significantly improve the fit, $\chi^2(1)=2.50, p = 0.11$ in the 30\% replacement scenario, and $\chi^2(1)=2.15, p = 0.14$ in the 70\% replacement scenario.

We next turn to the significance of our intervention on the passive drop of mutating verbs. We used as a reduced model a linear-mixed effects model with \textsc{corpus type} as a fixed effect. We compared the reduced model to a full model that also used whether the verb was a \textsc{mutating} verb as a fixed effect. Both models included \textsc{verb}, \textsc{verb class}, \textsc{frame}, and \textsc{model seed} as random effects. In the 30\% replacement condition, a likelihood ratio test showed that the full model was a significantly better fit, $\chi^2(1)=7.14, p=0.008$, suggesting that argument affectedness does affect model judgments of passivizability. We found similar results in the 70\% condition, $\chi^2(1)=70.0, p<0.001$.

Unlike in Experiment~2A, where a proportionate change in the frequency of occurrence affected every mutating verb's passive drop similarly, in Experiment~2B we found that altering a verb's distribution did not always result in a decrease in passive drop: intervening on the verbs \textit{resemble} and \textit{cost} consistently reduced passive drop, while intervening on \textit{last} did not. To test whether verb identity significantly moderated the effect of our intervention, we compared two mixed-effects models. Both models included main effects of \textsc{corpus type}, \textsc{verb}, whether the verb was \textsc{mutated}, and manipulation \textsc{percentage}, as well as random intercepts for each \textsc{frame} and \textsc{model seed}. The full model added an interaction term between mutation status and verb identity (\textsc{mutating $\times$ verb}), allowing different verbs to show different responses when their distributions were altered. A likelihood ratio test revealed that the full model provided a significantly better fit to the data than the reduced model, $\chi^2(2) = 47.00, p < 0.001$. This indicates that verb identity significantly moderated the effect of the manipulation.

Was there a difference between intervening on a moderate or large proportion of a verb's occurrences? We investigated this question by comparing the passive drop assigned to verbs at 30\% and 70\% replacement rates. We compared a reduced linear mixed-effects model with whether the verb was \textsc{mutating} as a fixed effect against a full model that additionally included the \textsc{percentage} of occurrences altered as a discrete variable. Both models included \textsc{verb}, \textsc{verb class}, \textsc{frame}, and \textsc{model seed} as random effects. A likelihood ratio test showed that the full model did not provide a significantly better fit to the data, $\chi^2(1)=0.52, p=0.47$, suggesting that passive drop was not significantly affected by increasing how extensive our intervention was.

\subsection{Discussion}

The goal of Experiment~2B was to test if the semantics of a verb has a causal effect on language models' passivizability judgments for the verb. We manipulated distributional cues to the semantics of the verb: we placed the mutating verb in sentences with arguments that occur with \textcolor{agt-pat}{agent-patient} verbs, and are consequently more likely to be affected. We found a significant main effect of our intervention on our mutating verbs' passivizability, suggesting that affectedness impacts our model's judgments of verb passivizability. We additionally found that changing how many of a verb's active transitive occurrences we intervened on did not significantly affect the verb's change in passivizability.

While we found a significant main effect of affectedness, we found that the magnitude of the effect interacted with the mutating verb's identity, with \textit{cost} and \textit{resemble} demonstrating consistent decreases in passive drop, while \textit{last} did not. One possible account for why \textit{last} behaved differently from \textit{cost} and \textit{resemble} arises from the original distributional statistics of the verbs. Whereas the majority of the total verbal occurrences of \textit{cost} (53.3\%) and \textit{resemble} (66.3\%) were classified as active transitive in the training corpus, the same is not true for \textit{last}; only 13.0\% of the sentences that \textit{last} appeared in were classified as active transitive, while 86.9\% of the occurrences of \textit{last} were classified as \textsc{other}. Some instances of \textit{last} classified as \textsc{other} are given in (\nextx):
\pex
\a Don't put something in just one form and expect it to last.
\a There's a very large body of research that says that more generous benefits and benefits that last longer ...
\xe
Thus, although we manipulated an equal proportion of the active transitive sentences that each verb occurred in, our intervention changed 16.0\% of the \textit{total} occurrences of \textit{cost} and 19.9\% of the total occurrences of \textit{resemble}, but only 3.9\% of the total occurrences of \textit{last}.
As the neural networks we trained used a single vector embedding to represent all uses of a word, as is standard, this finding could indicate that in neural networks passivizability can ``spill over'' from uses of a verb in contexts not immediately relevant to passivization such as intransitive sentences.
This raises the question of whether similar spillover effects might occur in the other direction: would changing the affectedness of intransitive uses of \textit{last} affect the passivizability of transitive uses of the verb?

Our choice to intervene only on active transitive sentences may also explain why we found no significant difference between a moderate (30\%) and a more extensive (70\%) manipulation of the verb's active transitive occurrences. Although the extensive manipulation intervened on 70\% of the active transitive uses of the mutating verbs, occurrences of the mutating verbs in other contexts were not replaced, resulting in more than half of the total occurrences of each mutating verb being left in their original contexts. Specifically, in the condition labeled as 70\%, we intervened on 37.3\% (vs. 16.0\% in the `moderate' manipulation) of the total occurrences of \textit{cost}, 46.4\% (vs. 19.9\%) of the total occurrences of \textit{resemble}, and just 9.1\% (vs. 3.9\%) of the total occurrences of \textit{last}. While we more than doubled the number of sentences we intervened on, then, it is possible that the more extensive manipulation did not shift the overall distribution of each verb's contexts enough to produce a detectable change in affectedness. Future work can consider manipulations or interpolations over \textit{all} occurrences of a verb without limiting the syntactic context of intervention. Doing so would allow for more extensive changes to the distributions of verbs, although this procedure would require strict control over which target verb sentences are used to replace mutating verb sentences, in order to avoid changing the syntactic distribution of the verb (e.g. by replacing a transitive occurrence of a verb with an intransitive occurrence).

Overall, results from this experiment suggest that changing the distribution of a verb's arguments affects its passivizability, although the extent to which this occurs is clouded by limitations in the filtering process. This finding is consistent with the body of experimental and computational work showing that the affectedness of a verb is a significant predictor of human acceptability ratings across languages (e.g. \citealt{ambridgeetal2016,ambridgeetal2023,liuambridge2021}), and suggests that model judgments of acceptability may be driven by similar factors as human judgments.

\section{Experiment~3: Testing the interaction between entrenchment and affectedness}

Experiments~2A and~2B were designed to test, by manipulating the distribution of the active and passive forms of verbs in the training corpus, how passivizability judgments are affected by the frequency of active and passive forms of the verb and by the lexical semantics of the verb. Experiment~2A found evidence that a verb's active-to-passive frequency ratio, its A/P ratio, was a significant predictor of passivizability. Experiment~2B found that the affectedness of a verb's semantics affects its passivizability, although this effect may have been appeared verb-dependent. These studies tested each hypothesis independently, and were operationalized by filtering an existing corpus.

Experiment~3 presents a complementary method to test both the entrenchment and the affectedness hypothesis that further allows us to explore the relationship between the entrenchment and affectedness hypotheses. In this experiment, we ask whether one factor is more primary than the other in determining a verb's passivizability, and whether entrenchment and affectedness interact.

Instead of ablating or altering existing parts of our training corpus, we introduced a \textbf{novel} verb into the corpus that \textit{only} occurred in active sentences. We then tested if passivizability judgments from models trained on this corpus were consistent with the predictions of the entrenchment and affectedness hypotheses. We repeated this experiment a number of times, manipulating two factors: first, whether the verb occurred in sentences that had high or low affectedness; and second, the number of times the verb appeared in the training corpus (again, only in the active voice). This design allowed us to test whether the lexical semantics of the novel verb's context changed its passivizability in the absence of confounds caused by verbs' naturally differing syntactic distributions as well as limitations in our filtering procedure. It also allowed us to test for an interaction between the entrenchment and affectedness hypotheses.

The two hypotheses make the following predictions for this design. The entrenchment hypothesis, which attributes unpassivizability to a high relative frequency of the active compared to the passive, predicts that, since the novel verb never appears in the passive voice, increasing the number of times it appears in the active voice should cause its passive drop to increase, regardless of the semantic contexts in which the verb appears. On the other hand, the affectedness hypothesis predicts that a novel verb will be more passivizable if it occurs only in high-affectedness contexts than if it occurs only in low-affectedness contexts.

What interaction patterns can we expect between entrenchment and affectedness? One possible outcome could be entrenchment effects that arise only when the verb occurs in high-affectedness contexts. If a verb occurs in high-affectedness contexts, learners may expect the novel verb to be potentially passivizable, and will then rely on the active-to-passive ratio to determine if that is the case. By contrast, if a verb occurs in low-affectedness contexts, learners may have weaker expectations about its likelihood of appearing in the passive, and as such may be less sensitive to the relative frequency of the active compared to the passive.

\subsection{Procedure}

We generated sets of sentences with high and low affectedness using GPT-4o \citep{openaietal2024}. We first prompted the model to generate carrier sentences based on a set of seed verbs used by \citet{ambridgeetal2016}. We next prompted GPT-4o to rate the affectedness of each carrier sentence based on questions adapted from \citet{ambridgeetal2016} and \citet{reisingeretal2015}, who used the questions to collect sentence affectedness ratings from human raters on a scale of one to five. We only kept low-affectedness sentences, defined as sentences whose average affectedness rating was below two, and high-affectedness sentences, whose average affectedness rating was above four. Table~\ref{tab:generated_data} provides examples of high-affectedness and low-affectedness carrier sentences, and \ref{sec:exp3_verbs} gives a list of the verbs used in the generated sentences. The prompts we used are given in \ref{sec:exp3_prompts}. All stimuli were generated in October 2024. 

\begin{table}[h]
    \centering
    \begin{tabular}{ll}
    \toprule
    High affectedness & The mob boss \textbf{murdered} the rival gang leader. \\
     & The performer \textbf{frightened} the audience with a sudden scream.
 \\
    & The thief \textbf{robbed} the jewelry store.\\
    \midrule
     Low affectedness & The property \textbf{bordered} the national park.\\
     & The dog \textbf{feared} the sound of thunder.
\\
    & The musician \textbf{heard} the applause from the audience.\\
     \bottomrule
    \end{tabular}
    \caption{Examples of high and low affectedness carrier sentences generated by GPT-4o.}
    \label{tab:generated_data}
\end{table}

Using these carrier sentences, we formed a training set containing 2000 high-affectedness carrier sentences and 2000 low-affectedness carrier sentences. Each carrier sentence was surrounded by two sentences of GPT-4o generated context before and after the carrier sentence, forming a five sentence-long paragraph. Next, we substituted a novel verb into some of the carrier sentences. We varied whether the novel verb occurred in high-affectedness or low-affectedness contexts, as well as the number of sentences $n$ the novel verb occurred in, $n \in \{0,1,2,4,8,10,16,32,64,100,128, 256, 500,$ $512, 1000,1024,2000\}$. Carrier sentences where the verb was not replaced with the novel verb retained their original verb, such that the total size of the corpus was matched across conditions (all of the conditions other than $n = 2000$ included such sentences). We then shuffled this training set into the same 100 million word training corpus we have used in all of our experiments so far, and trained five randomly initialized models with different seeds on each combined corpus using the same setup as in Experiments~2A and~2B.

We used a new set of carrier sentences, distinct from those in the training set, to form test sets of 100 sentences from which to obtain passivizability judgments. In order to isolate the changing effect of the novel verb's passivizability on passive drop from variability in passive drop attributable to other parts of the carrier sentences, we created test sets that substituted the \textit{original} verb or the \textit{novel} verb into the carrier sentence. Low-affectedness test sets contained 100 carrier sentences where the verb was seen in low-affectedness contexts, while high-affectedness test sets that contained 100 carrier sentences with high affectedness. We tested the novel verb's passivizability only on the test sets that corresponded to the training condition: models trained on a dataset where the novel verb was seen in high-affectedness contexts during training were tested on the 100 sentences where the novel verb was also seen in high-affectedness contexts, and likewise for low-affectedness contexts. We made this choice to avoid an adversarial testing setting where the semantic material surrounding the verb sometimes differed sharply between the training and test sets.

\subsection{Results}
\begin{figure}[t]
    \centering
    \includegraphics[width=.9\textwidth]{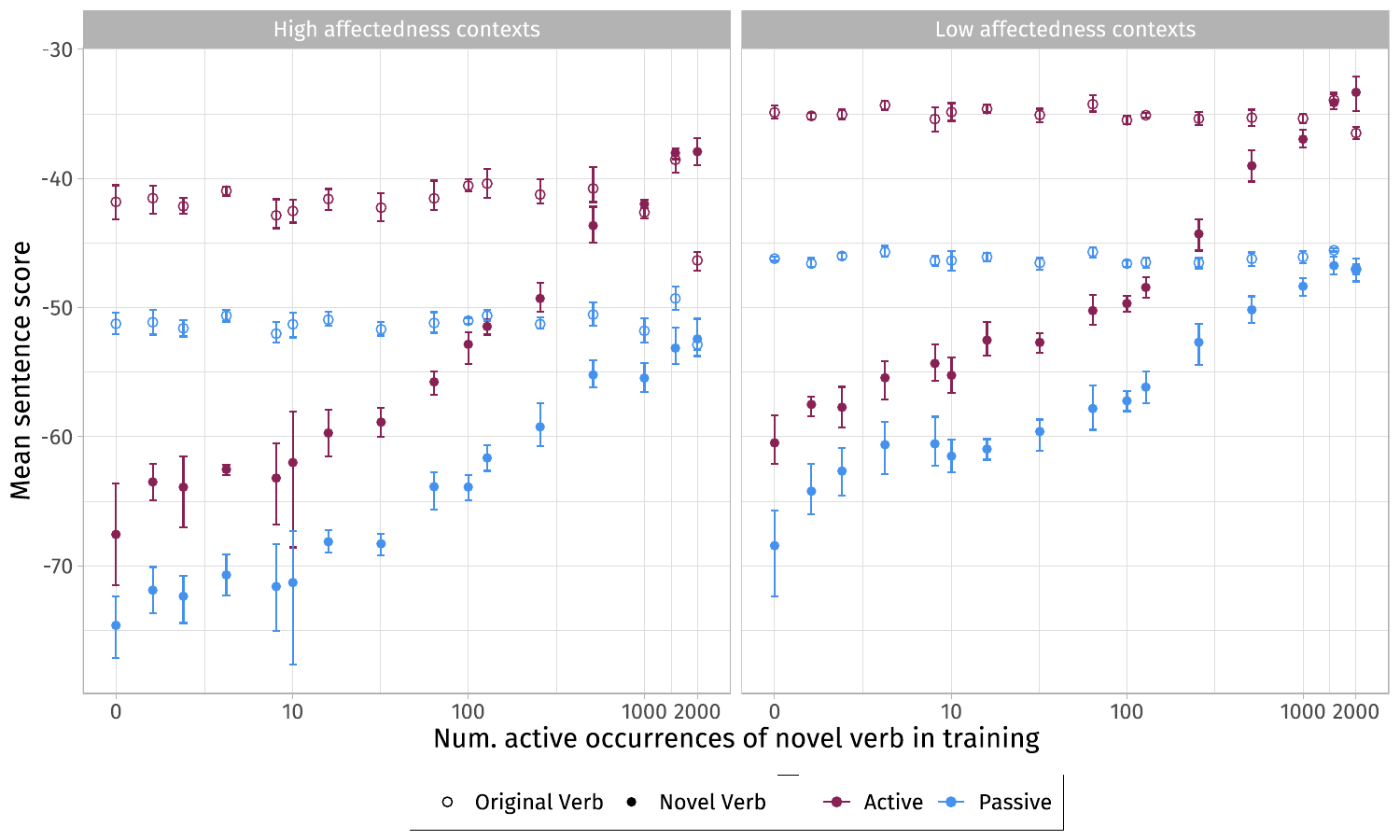}
    \caption{\textit{Mean sentence scores assigned to test sentences by models with differing levels of exposure to a novel verb in the active (the novel verb never occurred in the passive)}. Sentence scores assigned to sentences including the novel verb increased as the novel verb was used more frequently in the corpus. Scores assigned to original versions of the same test sentences (hollow dots) are shown as a baseline.}
    \label{fig:exp3_raw_scores}
\end{figure}

\begin{figure}
    \centering
    \includegraphics[width=0.9\linewidth]{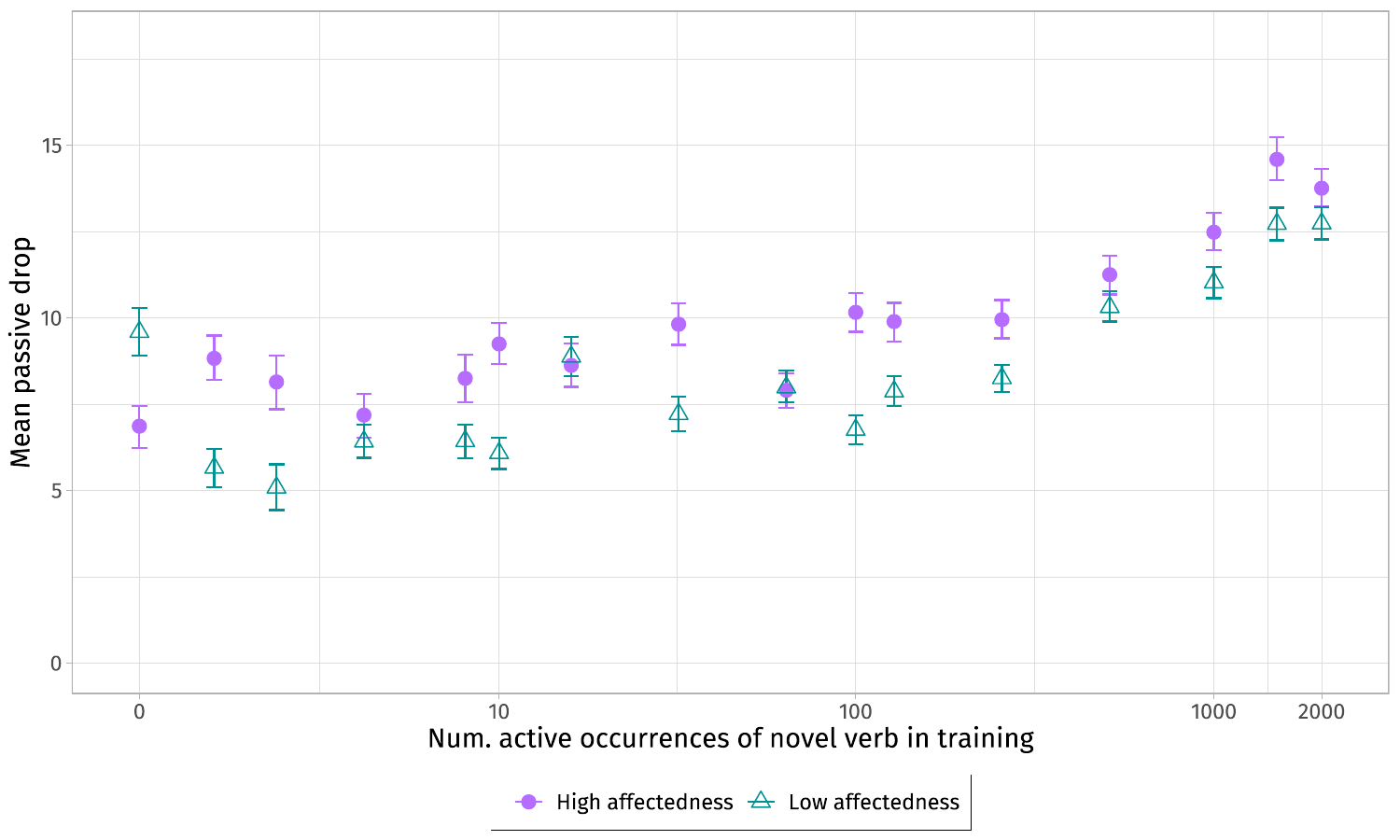}
    \caption{\textit{Change in passive drop as a verb is seen more often in the active and in more semantically affected contexts}. Passive drop increased as models were exposed more frequently to the novel verb in active sentences while not seeing the verb in passive sentences. The novel verb also had a higher passive drop when trained in high-affectedness contexts.}
    \label{fig:exp3_pass_drop}
\end{figure}

We first examined the mean sentence scores assigned to active and passive test sentences containing the novel verb (filled points in Figure~\ref{fig:exp3_raw_scores}) compared to the original verb (hollow points). Across all conditions, active sentences were given higher sentence scores than passive sentences. When the novel verb was not seen in the training data at all---i.e. none of the verbs in the corpus generated by GPT-4o were replaced with the novel verb---test sentences containing the novel verb were rated worse than the original versions of the test sentences (hollow points). In both high and low-affectedness conditions, the sentence scores assigned to sentences containing the novel verb increased as the novel verb was seen more frequently in the training corpus and, at around 1000 occurrences converged with the scores assigned to the original test sentences.

Sentences with high-affectedness contexts had lower sentence scores than sentences with low-affectedness contexts, regardless of whether they included the original verb or the novel verb. We suspect this is due to the fact that test sentences in the high affectedness condition were on average two words longer than test sentences in the low affectedness condition. Language model acceptability judgments are known to be sensitive to sentence length, though the relationship may be complex \citep{tjuatjaetal2025}. That being said, because passive drop is computed within-item, comparing the active and passive versions from the same minimal pair, any difference in length across sentence pairs should cancel out.

Next, we consider how the passive drop associated with the novel verb varied as models were exposed to more occurrences of the novel verb in the active voice, and no occurrences of this verb in the passive (Figure~\ref{fig:exp3_pass_drop}). Regardless of affectedness, passive drop increased as the number of occurrences increased. To evaluate the statistical significance of this pattern, we fit a linear mixed-effects model predicting passive drop from the \textsc{affectedness} of the context as a fixed effect and random intercepts for \textsc{model seed} and \textsc{test item}. We compared this model to a model that additionally included \textsc{number of occurrences} of the novel verb as a fixed effect. A likelihood ratio test showed that number of occurrences was a highly significant predictor of passive drop ($\chi^2(1)=74.25, p<0.001$).

The novel verb's passivizability was also impacted by affectedness: for a given number of occurrences of the novel verb in the training corpus, sentences containing the verb were assigned a higher passive drop in high-affectedness contexts than in low-affectedness ones. To verify the significance of this effect, we fit a linear mixed-effects model predicting passive drop from \textsc{number of occurrences} of the novel verb as a fixed effect and random intercepts for \textsc{model seed} and \textsc{test item}, and compared this model to a full model that additionally included \textsc{affectedness} as a fixed effect. A likelihood ratio test showed that high affectedness significantly increased passive drop, though this effect was not as dramatic as that of frequency ($\chi^2(1)=3.88, p = 0.049$).

Although there were significant main effects of both affectedness and frequency on passive drop, we did not find a significant interaction between the two. We compared a linear mixed-effects model predicting passive drop using \textsc{affectedness} and \textsc{frequency} as fixed effects and \textsc{model seed} and \textsc{test item} as random effects against a full model that additionally included the interaction between affectedness and frequency as a fixed effect. A likelihood-ratio test showed that there was no significant difference between the full model and the reduced model ($\chi^2(1) = 0.22, p=0.645$).

\subsection{Discussion}

Experiment~3 found support for both the entrenchment and affectedness hypothesis. As we entrenched a novel verb in the active voice, by increasing the number of active occurrences of the verb without including any passive ones in the corpus, the novel verb became increasingly unpassivizable. At the same time, the novel verb was also assigned higher passive drop if it was seen in high-affectedness contexts than if it was seen in low-affectedness contexts. The affectedness effect, while consistent, was smaller than the effect of frequency.

These results are consistent with our findings in Experiment~2A, There, we showed that entrenchment effects significantly affect passive drop---the higher the active-to-passive ratio, the less passivizable a verb was. The current experiment, where there were no passive occurrences of the verb at all, found a similar relationship between entrenchment and passive drop. 

The results of this experiment are also consistent with the results of Experiment~2B, which found a main effect of affectedness on passivizability. Experiment~3, where each verb \textit{only} occurred in active transitive sentences, found that passive drop was consistently higher when the novel verb was placed in high-affectedness contexts. Experiment~3 further found that frequency and affectedness independently affect a verb's passivizability, without a clear interaction between them.

An extension of the method we introduce here could explore other facets of input context. For example, by selectively introducing a small number of passive sentences, one could test how entrenchment effects diminish as there is increased competition between the two structures. Another potential extension would study graded affectedness effects: for Experiment~3 we selected semantic contexts at the two extremes of the affectedness spectrum, to maximize the likelihood of detecting an effect. But the novel verb's lexical semantics could also be varied in a more fine-grained way, for instance by allowing the novel verb to appear in both high- and low-affectedness contexts to differing proportions, or by generating sentences with gradient affectedness.

\color{black}
\section{General Discussion}
Even highly productive grammatical rules often have environments in which they cannot apply. Since learners typically do not receive direct evidence that a particular sentence is \textit{not} grammatical, they must infer such exceptions to generalizations from indirect evidence. In this study, we explored the sources of indirect evidence that a learner might use to acquire the restrictions on English passivization. We focused on two hypotheses: the entrenchment hypothesis, which attribute the unpassivizability of particular verbs to frequency asymmetries between the active and passive forms of those verbs; and the affectedness hypothesis, which attribute unpassivizability to an incompatibility between the semantics of the verb and the passive construction, a construction where the subject of the sentence is expected to be affected by the event described in the sentence.

To test these hypotheses, we used transformer language models trained on a corpus that approximates the amount of exposure to language that human learners receive. This aspect of our design is critical as mainstream language models trained on much larger amounts of data have access to considerably more indirect evidence than humans \citep{wilcox2025}.
We found that language models' judgments of verb passivizability correlate highly with human judgments ($r=0.9$). Then, by performing targeted interventions on the model's training data, we showed that the relative frequency with which a verb appears in the active and passive constructions in the training data provided indirect evidence for the model to learn exceptionality, though this factor did not fully explain the magnitude of the difference between highly passivizable verbs and highly unpassivizable ones. We also found a smaller effect of affectedness, where verbs observed during training in high-affectedness contexts were rated as more passivizable.

\subsection{Human judgments of exceptions to passivization} 
We used the English passive as a case study of the acquisition of exceptions to syntactic generalizations. As much of the existing literature on these exceptions relies on linguists' acceptability judgments, in Experiment~1A we sought to first verify these judgments with naive participants. A further goal of the experiment was to identify any gradience in these judgments that could serve as a benchmark for quantitative computational modeling. In general, this experiment reproduced the linguists' judgments for a number of individual verbs and classes of verbs, but there were discrepancies in some cases---specifically, verbs in the \textit{ooze} and \textit{advantage} classes did not significantly differ in their passivizability from canonically passivizable verbs. We further showed that English speakers' judgments of passive exceptions are more nuanced than binary acceptability judgments might suggest. Not all verbs reported to be unpassivizable were equally unacceptable: for instance, although \textit{last} and \textit{resemble} are both reported as unacceptable in the literature, we found a much larger average passive drop for \textit{last} than \textit{resemble}.

\subsection{Using neural networks as models of human learners}

What benefits do neural network language models bring to the study of human language acquisition? Existing studies of children's acquisition of the restrictions on passivization (e.g. \citealt{foxgrodzinsky1998,gordonchafetz1990,maratsos1985}) are limited by researchers' inability to exert full control over a child's linguistic input: %
It would be difficult, for instance, to ensure that a child never hears a specific verb in the passive voice, as we did in Experiment~3. These issues are addressed, to a limited extent, by human artificial language learning experiments, which target specific hypotheses through controlled experimentation on constructed languages. But the artificial languages used in these experiments are by necessity much simpler than natural languages---often with vocabularies of fewer than 50 items---and it is unclear if they engage the same cognitive mechanisms as first language acquisition.

Using neural networks as model learners addresses these existing methodological lacunae in acquisition studies. Neural networks, unlike human learners, are trained on data over which we have full control. Unlike the symbolic models sometimes used in language acquisition research, which are often simplified proof-of-concept systems, neural networks are broad-coverage models that can be trained on a corpus that is as close to the input to human children as possible \citep{warstadtetal2023, vongetal2024}. The greater degree of researcher control afforded by computational experiments is not limited to intervention experiments: while we have not explored this direction in the present work, the neural network paradigm makes it possible to probe a model's internal processes to understand which mechanisms are vital to the model's learning process and form hypotheses about how humans may learn \citep{baroni2022,lakretzetal2021}.

Neural network models can also be used in future work to shed light on cross-lingual typological patterns of exceptions. Models can be trained on mixes of corpora in different languages which can be strictly controlled to explore the relationship between the languages (e.g. \citealt{constantinescuetal2024}). Mechanistic analysis of these multilingual models may reflect shared representations and mechanisms, which would support the argument that passive constructions across different languages reflect a common semantic universal \citep{papadimitriouetal2021,ambridgeetal2023}. Finally, withholding instances of the passive in one language and not another may allow researchers to explore whether models extend the passive construction across languages \citep{papadimitriouetal2023}.

While neural network language models have substantial potential as model learners for grammatical phenomena, there are a number of methodological limitations to this modeling approach. First, the value of modeling is limited by the interpretability and cognitive plausibility of the models we use  \citep{baroni2022}. Without a clear understanding of the inductive biases of the particular neural network chosen for comparison, we cannot make a fair comparison between these models and our theories of human cognition. Although we highlighted some similarities between the \mbox{GPT-2} architecture and human language learning and processing, this architecture is clearly not a perfect model for human language learning (for example, transformers' working memory constraints are fundamentally different from those of humans; \citealt{armenietal2022,timkeylinzen2023}), and care should be taken to make fair comparisons between the two.

Secondly, the methodology of intervening on a corpus by removing particular classes of examples, or swapping some words out for others in particular syntactic contexts, relies heavily on the accuracy of the linguistic analysis tools we have at our disposal. Without the ability to precisely and accurately parse and alter a corpus containing heterogeneous data, it is difficult to ensure the feasibility and reliability of any intervention. For instance, the filters we used to make our intervention in Experiment~2A may have introduced new confounds in the training data by removing clear examples of passive sentences while failing to identify and filter out more complex examples. While our goal in Experiment~2B was to transplant verbs into agent-patient environments, our intervention was limited to replacing active transitive sentences with other active transitive sentences, and so could not account for verb-level differences in syntactic distribution. Refining the filtering process would allow us to test hypotheses that are more narrowly defined, and thus to make conclusions at a more granular level.

The paradigm we used in Experiment~3, where we construct new sentences with a novel verb and insert them into the corpus, may serve as a cleaner and easier to control alternative to the syntactic filtering paradigm. However, this method is contingent on using another language model to create human-like language input. While generative models can create much larger-scale datasets than can be hand-crafted, the distribution of the data formed by these models may not align completely with that of human-created text. Any analyses based on these synthetic datasets may thus reflect artifacts of the language model's generative procedure, rather than features of human language use.

Finally, the computational cost of training models on each modified corpus can be high. We trained a total of 125 models for Experiment~2, each requiring two days to train and using approximately 3e15 floating point operations (FLOPs). These training regimes are highly computationally expensive and their potential environmental impacts should be considered. These limitations notwithstanding, we hope that the use of targeted interventions on naturalistic data can be used to compare the plausibility of hypotheses in situations where such interventions are not possible in human research.

\subsection{Implications for human learners}
What are the consequences of our results to the study of human learners? We have shown in causal experiments using sentence deletion (Experiment~2A) and novel verbs (Experiment~3) that models can leverage the relative frequency of the active and passive constructions in training data to learn exceptions. This finding is consistent with usage-based approaches to human language acquisition \citep{tomasello2000,goldberg2006}, and can be taken as an existence proof demonstrating that a learner could display a human-like pattern as a consequence of tracking the statistics of verb-construction co-occurrences.

At the same time, humans and language models clearly differ in their learning mechanisms, goals, and resources. It is possible, then, that while our models' acceptability judgments are largely similar to those of humans, they achieve such behavior via a very different developmental pathway. Indeed, since our neural networks learn the task of next word prediction by tracking word co-occurrences across a corpus, the fact that they are highly sensitive to frequency statistics is unsurprising. Although human learners can also track statistical information in their linguistic input \citep{saffranetal1996,thompsonnewport2007}, they do not rely solely on distributional statistics: language acquisition is also shaped by interactive and communicative social pressures. When a child produces an utterance that an adult finds unacceptable, the adult may repeat that utterance but produce a change in the erroneous portion which the child may then take up in their next utterance if the correction matches their intended meaning \citep{chouinardclark2003}. Children can learn from these reformulations in order to further their communicative goals---goals which our models lack. 

Humans can learn the concept of affectedness---a key factor in Experiment~2B and Experiment~3---from their direct experiences with the world. By contrast, our models were trained without access to sensorimotor input, and thus may lack these conceptual primitives (see, e.g., \citealt{lakeetal2017}); instead, they may have learned concepts that approximate, but are not identical to, human concepts of affectedness. Consequently, while in Experiment~2B and Experiment~3 our models were sensitive to the semantic context in which a verb occurred, we cannot conclude that people will be sensitive to this factor in the same way or to the same extent. For instance, while frequency played a larger causal role than affectedness in increasing unpassivizability in our models, semantic cues could be more predictive of passivizability for people, for whom the meaning of a sentence is key. The relative influence of affectedness and frequency on passivizability in humans thus remains an open question. That being said, we take the behavior exhibited by our models in Experiment~3 to constitute a hypothesis for how humans might behave in a learning task, which should be tested in future work.

In sum, while we have illustrated a potential pathway by which a neural network learner might conclude that a verb is unpassivizable, the same learning mechanisms might not be at play in human learners. Repeating these experiments using models that differ in architecture, and in particular models that have access to interaction and causal primitives, may help to disentangle which capacities are required to learn about passivizability.

\subsection{Other sources of indirect evidence}
While we have found support for the role of both entrenchment and affectedness in language models' judgments of passivizability, neither factor alone \textit{fully} accounted for the models' judgments: our interventions in both Experiments 2A and 2B did not consistently make a mutating verb as passivizable as its corresponding target verb. We showed in Experiment~3 that the effects of entrenchment and affectedness are additive, but did not compare these effect sizes to the effect sizes we found in actual unpassivizable verbs. Future work could use the generative method we propose in Experiment~3 to reconstruct the passivizability of a verb from these two sources to explore whether entrenchment and affectedness together can explain the full degree of a verb's passivizability.

If the effects of entrenchment and affectedness alone are insufficient to arrive at the full amount of unpassivizability we found in Experiment~1B, what other sources of evidence could language models---and, potentially, humans---rely on? One potential such source of evidence is the existence of similar constructions to the passive with a \textit{by}-phrase. These include (\nextx b), which differs from the passive (\nextx a) by just one word, or the active construction (\nextx c), which is used in functionally similar contexts:
\pex
\a Two hours were required by the meeting.
\a Two hours were required for the meeting.
\a The meeting required two hours.
\xe

\noindent The existence of alternations like those illustrated in (\lastx) could affect the acquisition of passive exceptions in two ways. 
First, if learners often hear (\lastx b) or (\lastx c) in contexts where they might otherwise expect to hear (\lastx a), they may conclude that (\lastx a) is unacceptable \citep{ambridgeetal2015,boydgoldberg2011,clark1987,goldberg1995}, as we hypothesized was the case for the frequency asymmetry between actives and passives in Experiment~2A. Second, the existence of alternatives like (\lastx b) for some but not all verbs may also help to explain the gradience in acceptability that we see across verbs within a verb class in Experiment~1A, if viewed through the lens of the noisy channel theory \citep{levy2008,gibsonetal2013}. If English speakers find (\lastx b) to be unacceptable but have access to an alternative like (\lastx a) that is acceptable, they may process (\lastx b) as a corruption of the acceptable (\lastx a), and thus judge it as more acceptable than a similar passive sentence for which there is no corresponding alternative. These hypotheses and their interactions can be explored through similar training data interventions to the ones we implemented in this paper.

\section{Conclusion}
How is knowledge of grammar related to the learner's linguistic input? In this paper, we studied how exceptions to passivization in English, which must be learned through indirect evidence, can be learned by a neural network language model, a broad-coverage model that can learn from amounts of data comparable to those that humans learn from. We manipulated the training corpus of the neural network language models to test the causal links between input and the models' behavior. We first showed that passivizability judgments extracted from a language model match human acceptability judgments to a substantial extent (Experiment~1B). We then made targeted changes to the models' training data to measure the effects of entrenchment and affectedness, two factors that have been argued to be implicated in the learning of passives in humans, on the models' learning of these patterns. Through our interventions, we found that changing the verb's relative frequency of occurrence in the active and passive voice affected the models' judgments of its passivizability, as predicted by the entrenchment hypothesis (Experiment~2A). We also found that manipulating a verb's semantics by changing the arguments it appears with in active transitive sentences significantly affected the passivizability of the verb, although the magnitude of this effect interacted with verb identity (Experiment 2B). In Experiment~3, a more tightly-controlled novel verb experiment, we found further evidence that frequency and affectedness both significant affect passivizability judgments, and additionally found that these two factors do not interact. These findings illustrate a method for testing hypotheses about how large-scale linguistic input affects learning, and raise new questions that can be tested with human participants.

\section*{CRediT authorship contribution statement}
\textbf{Cara Su-Yi Leong}: Conceptualization, Data curation, Formal analysis, Methodology, Investigation, Visualization, Writing – original draft, Writing - review \& editing. 
\textbf{Tal Linzen}: Conceptualization, Formal analysis, Funding acquisition, Methodology, Writing – review \& editing, Supervision.
\section*{Declaration of competing interest}
The authors declare that there is no any commercial or associative interest that represents a conflict of interest in connection with the work submitted.
\section*{Acknowledgments}
This material is based upon work supported by the National Science Foundation (NSF) under Grant No. BCS-2114505. We thank the audience of Society for Computation in Linguistics 2023 and members of the NYU Computation and Psycholinguistics Lab for
comments. This work was supported in part through the NYU IT High Performance Computing resources, services, and staff expertise.

\section*{Data availability}
The materials, acceptability judgment data, and analysis scripts are available at the following website:
\href{https://github.com/craaaa/exceptions}{https://github.com/craaaa/exceptions}

\bibliographystyle{elsarticle-harv}
\bibliography{passives,additional}

@article{huang2024large,
  title={Large-scale benchmark yields no evidence that language model surprisal explains syntactic disambiguation difficulty},
  author={Huang, Kuan-Jung and Arehalli, Suhas and Kugemoto, Mari and Muxica, Christian and Prasad, Grusha and Dillon, Brian and Linzen, Tal},
  journal={Journal of Memory and Language},
  volume={137},
  pages={104510},
  year={2024},
  publisher={Elsevier}
}

@article{touvron2023llama,
 author = {Touvron, Hugo and Martin, Louis and Stone, Kevin and Albert, Peter and Almahairi, Amjad and Babaei, Yasmine and Bashlykov, Nikolay and Batra, Soumya and Bhargava, Prajjwal and Bhosale, Shruti and others},
 journal = {ArXiv preprint},
 title = {Llama 2: Open foundation and fine-tuned chat models},
 url = {https://arxiv.org/abs/2307.09288},
 volume = {abs/2307.09288},
 year = {2023}
}

@article{achiam2023gpt,
 author = {Achiam, Josh and Adler, Steven and Agarwal, Sandhini and Ahmad, Lama and Akkaya, Ilge and Aleman, Florencia Leoni and Almeida, Diogo and Altenschmidt, Janko and Altman, Sam and Anadkat, Shyamal and others},
 journal = {ArXiv preprint},
 title = {Gpt-4 technical report},
 url = {https://arxiv.org/abs/2303.08774},
 volume = {abs/2303.08774},
 year = {2023}
}

@article{frank2023bridging,
  title={Bridging the data gap between children and large language models},
  author={Frank, Michael C},
  journal={Trends in Cognitive Sciences},
  volume={27},
  number={11},
  pages={990--992},
  year={2023},
  publisher={Elsevier}
}

@article{wilcox2025,
  title={Bigger is not always better: The importance of human-scale language modeling for psycholinguistics},
  author={Wilcox, Ethan Gotlieb and Hu, Michael Y and Mueller, Aaron and Warstadt, Alex and Choshen, Leshem and Zhuang, Chengxu and Williams, Adina and Cotterell, Ryan and Linzen, Tal},
  journal={Journal of Memory and Language},
  volume={144},
  pages={104650},
  year={2025},
  publisher={Elsevier}
}

@article{linzen2021syntactic,
	Author = {Linzen, Tal and Baroni, Marco},
	Journal = {Annual Review of Linguistics},
	Pages = {195--212},
	Title = {Syntactic structure from deep learning},
	Volume = {7},
	Year = {2021}}

@inproceedings{mccoy-etal-2020-berts,
    title = "{BERT}s of a feather do not generalize together: Large variability in generalization across models with similar test set performance",
    author = "McCoy, R. Thomas  and
      Min, Junghyun  and
      Linzen, Tal",
    editor = "Alishahi, Afra  and
      Belinkov, Yonatan  and
      Chrupa{\l}a, Grzegorz  and
      Hupkes, Dieuwke  and
      Pinter, Yuval  and
      Sajjad, Hassan",
    booktitle = "Proceedings of the Third BlackboxNLP Workshop on Analyzing and Interpreting Neural Networks for NLP",
    month = nov,
    year = "2020",
    address = "Online",
    publisher = "Association for Computational Linguistics",
    url = "https://aclanthology.org/2020.blackboxnlp-1.21",
    doi = "10.18653/v1/2020.blackboxnlp-1.21",
    pages = "217--227",
}

@article{ambridgeetal2015,
  title = {Preemption versus {{Entrenchment}}: {{Towards}} a {{Construction-General Solution}} to the {{Problem}} of the {{Retreat}} from {{Verb Argument Structure Overgeneralization}}},
  shorttitle = {Preemption versus {{Entrenchment}}},
  author = {Ambridge, Ben and Bidgood, Amy and Twomey, Katherine E. and Pine, Julian M. and Rowland, Caroline F. and Freudenthal, Daniel},
  year = 2015,
  month = apr,
  journal = {PLoS ONE},
  volume = {10},
  number = {4},
  pages = {e0123723},
  issn = {1932-6203},
  doi = {10.1371/journal.pone.0123723},
  urldate = {2024-05-20},
  pmcid = {PMC4412412},
  pmid = {25919003}
}

@article{ambridgeetal2016,
  title = {Is {{Passive Syntax Semantically Constrained}}? {{Evidence From Adult Grammaticality Judgment}} and {{Comprehension Studies}}},
  shorttitle = {Is {{Passive Syntax Semantically Constrained}}?},
  author = {Ambridge, Ben and Bidgood, Amy and Pine, Julian M. and Rowland, Caroline F. and Freudenthal, Daniel},
  year = 2016,
  month = aug,
  journal = {Cognitive Science},
  volume = {40},
  number = {6},
  pages = {1435--1459},
  issn = {0364-0213},
  doi = {10.1111/cogs.12277},
  urldate = {2022-09-22},
  pmcid = {PMC4996337},
  pmid = {26607289}
}

@article{ambridgeetal2023,
  title = {He Was Run-over by a Bus: {{Passive}} -- but Not Pseudo-Passive -- Sentences Are Rated as More Acceptable When the Subject Is Highly Affected. {{New}} Data from {{Hebrew}}, and a Meta-Analytic Synthesis across {{English}}, {{Balinese}}, {{Hebrew}}, {{Indonesian}} and {{Mandarin}}},
  shorttitle = {He Was Run-over by a Bus},
  author = {Ambridge, Ben and Arnon, Inbal and Bekman, Dani},
  year = 2023,
  month = jul,
  journal = {Glossa Psycholinguistics},
  volume = {2},
  number = {1},
  issn = {2767-0279},
  doi = {10.5070/G6011177},
  urldate = {2024-11-22},
  copyright = {Copyright: \copyright{} 2023 The Author(s). This is an open-access article distributed under the terms of the Creative Commons Attribution 4.0 International License (CC-BY 4.0), which permits unrestricted use, distribution, and reproduction in any medium, provided the original author and source are credited. See http://creativecommons.org/licenses/by/4.0/.},
  langid = {english}
}

@inproceedings{armenietal2022,
  title = {Characterizing Verbatim Short-Term Memory in Neural Language Models},
  booktitle = {Proceedings of the 26th Conference on Computational Natural Language Learning ({{CoNLL}})},
  author = {Armeni, Kristijan and Honey, Christopher and Linzen, Tal},
  editor = {Fokkens, Antske and Srikumar, Vivek},
  year = 2022,
  month = dec,
  pages = {405--424},
  publisher = {Association for Computational Linguistics},
  address = {Abu Dhabi, United Arab Emirates (Hybrid)},
  doi = {10.18653/v1/2022.conll-1.28}
}

@article{aryawibawaambridge2018,
  title = {Is {{Syntax Semantically Constrained}}? {{Evidence From}} a {{Grammaticality Judgment Study}} of {{Indonesian}}},
  shorttitle = {Is {{Syntax Semantically Constrained}}?},
  author = {Aryawibawa, I Nyoman and Ambridge, Ben},
  year = 2018,
  journal = {Cognitive Science},
  volume = {42},
  number = {8},
  pages = {3135--3148},
  issn = {1551-6709},
  doi = {10.1111/cogs.12697},
  urldate = {2024-11-22},
  copyright = {\copyright{} 2018 Cognitive Science Society, Inc.},
  langid = {english}
}

@article{bach1980,
  title = {In {{Defense}} of {{Passive}}},
  author = {Bach, Emmon W.},
  year = 1980,
  journal = {Linguistics and Philosophy},
  volume = {3},
  number = {3},
  eprint = {25001027},
  eprinttype = {jstor},
  pages = {297--341},
  publisher = {Springer},
  issn = {0165-0157},
  urldate = {2023-04-29}
}

@article{bakemanmcarthur1996,
  title = {Picturing Repeated Measures: {{Comments}} on {{Loftus}}, {{Morrison}}, and Others},
  shorttitle = {Picturing Repeated Measures},
  author = {Bakeman, Roger and Mcarthur, Duncan},
  year = 1996,
  month = dec,
  journal = {Behavior Research Methods, Instruments \& Computers},
  volume = {28},
  number = {4},
  pages = {584--589},
  issn = {1532-5970},
  doi = {10.3758/BF03200546},
  urldate = {2024-12-10},
  langid = {english}
}

@article{baker1979,
  title = {Syntactic {{Theory}} and the {{Projection Problem}}},
  author = {Baker, C. L.},
  year = 1979,
  journal = {Linguistic Inquiry},
  volume = {10},
  number = {4},
  eprint = {4178133},
  eprinttype = {jstor},
  pages = {533--581},
  publisher = {The MIT Press},
  issn = {0024-3892},
  urldate = {2023-02-10}
}

@misc{baroni2022,
  title = {On the Proper Role of Linguistically-Oriented Deep Net Analysis in Linguistic Theorizing},
  author = {Baroni, Marco},
  year = 2022,
  month = mar,
  number = {arXiv:2106.08694},
  eprint = {2106.08694},
  primaryclass = {cs},
  publisher = {arXiv},
  urldate = {2024-05-20},
  archiveprefix = {arXiv},
  langid = {english}
}

@article{beavers2011,
  title = {On Affectedness},
  author = {Beavers, John},
  year = 2011,
  journal = {Natural Language \& Linguistic Theory},
  volume = {29},
  number = {2},
  eprint = {41475291},
  eprinttype = {jstor},
  pages = {335--370},
  publisher = {Springer},
  issn = {0167-806X},
  urldate = {2022-06-29}
}

@article{bidgoodetal2020,
  title = {Syntactic {{Representations Are Both Abstract}} and {{Semantically Constrained}}: {{Evidence From Children}}'s and {{Adults}}' {{Comprehension}} and {{Production}}/{{Priming}} of the {{English Passive}}},
  shorttitle = {Syntactic {{Representations Are Both Abstract}} and {{Semantically Constrained}}},
  author = {Bidgood, Amy and Pine, Julian M. and Rowland, Caroline F. and Ambridge, Ben},
  year = 2020,
  journal = {Cognitive Science},
  volume = {44},
  number = {9},
  pages = {e12892},
  issn = {1551-6709},
  doi = {10.1111/cogs.12892},
  urldate = {2024-11-22},
  copyright = {\copyright{} 2020 The Authors. Cognitive Science published by Wiley Periodicals LLC on behalf of Cognitive Science Society (CSS)},
  langid = {english}
}

@article{boydgoldberg2011,
  title = {Learning {{What}} {{{\textbf{NOT}}}} to {{Say}}: {{The Role}} of {{Statistical Preemption}} and {{Categorization}} in {{{\emph{A}}}}-{{Adjective Production}}},
  shorttitle = {Learning {{What}} {{{\textbf{NOT}}}} to {{Say}}},
  author = {Boyd, Jeremy K. and Goldberg, Adele E.},
  year = 2011,
  journal = {Language},
  volume = {87},
  number = {1},
  pages = {55--83},
  issn = {1535-0665},
  doi = {10.1353/lan.2011.0012},
  urldate = {2023-04-09},
  langid = {english}
}

@incollection{brainebrooks1995,
  title = {Verb {{Argument Structure}} and the {{Problem}} of {{Avoiding}} an {{Overgeneral Grammar}}},
  booktitle = {Beyond Names for Things: {{Young}} Children's Acquisition of Verbs},
  author = {Braine, Martin D S and Brooks, Patricia J},
  editor = {Tomasello, Michael and Merriman, William Edward},
  year = 1995,
  pages = {353--376},
  publisher = {L. Erlbaum},
  address = {Hillsdale, N.J},
  isbn = {978-0-8058-1250-3},
  lccn = {P118 .B49 1995}
}

@article{brookstomasello1999,
  title = {Young Children Learn to Produce Passives with Nonce Verbs.},
  author = {Brooks, Patricia J. and Tomasello, Michael},
  year = 1999,
  journal = {Developmental Psychology},
  volume = {35},
  number = {1},
  pages = {29},
  publisher = {US: American Psychological Association},
  issn = {1939-0599},
  doi = {10.1037/0012-1649.35.1.29},
  urldate = {2023-02-14}
}

@book{brown1973,
  title = {A First Language: {{The}} Early Stages.},
  author = {Brown, Roger},
  year = 1973,
  publisher = {Harvard University Press.},
  address = {Cambridge, MA}
}

@article{chouinardclark2003,
  title = {Adult Reformulations of Child Errors as Negative Evidence},
  author = {Chouinard, Michelle M. and Clark, Eve V.},
  year = 2003,
  journal = {Journal of Child Language},
  volume = {30},
  number = {3},
  pages = {637--669},
  publisher = {Cambridge University Press},
  address = {United Kingdom},
  issn = {1469-7602},
  doi = {10.1017/S0305000903005701}
}

@incollection{clark1987,
  title = {The Principle of Contrast: {{A}} Constraint on Language Acquisition},
  shorttitle = {The Principle of Contrast},
  booktitle = {Mechanisms of Language Aquisition},
  author = {Clark, Eve V.},
  year = 1987,
  pages = {1--33},
  publisher = {Lawrence Erlbaum Associates, Inc},
  address = {Hillsdale, NJ, US},
  isbn = {978-0-89859-596-3 978-0-89859-973-2}
}

@incollection{comrie1988,
  title = {Passive and Voice},
  booktitle = {Passive and {{Voice}}},
  author = {Comrie, Bernard},
  editor = {Shibatani, Masayoshi},
  year = 1988,
  month = jan,
  series = {Typological {{Studies}} in {{Language}}},
  pages = {9--24},
  publisher = {John Benjamins Publishing Company},
  doi = {10.1075/tsl.16.04com},
  urldate = {2023-05-15},
  isbn = {978-90-272-2889-5 978-1-55619-018-6 978-90-272-2890-1 978-1-55619-019-3 978-90-272-8613-0},
  langid = {english}
}

@incollection{comrieetal1977,
  title = {In {{Defense}} of {{Spontaneous Demotion}}: {{The Impersonal Passive}}},
  shorttitle = {In {{Defense}} of {{Spontaneous Demotion}}},
  booktitle = {Grammatical {{Relations}}},
  author = {Comrie, Bernard and Cole, Peter and Sadock, Jerrold M.},
  year = 1977,
  month = dec,
  pages = {47--58},
  publisher = {Brill},
  doi = {10.1163/9789004368866_004},
  urldate = {2024-06-29},
  chapter = {Grammatical Relations},
  langid = {english}
}

@misc{constantinescuetal2024,
  title = {Investigating {{Critical Period Effects}} in {{Language Acquisition}} through {{Neural Language Models}}},
  author = {Constantinescu, Ionut and Pimentel, Tiago and Cotterell, Ryan and Warstadt, Alex},
  year = 2024,
  month = oct,
  number = {arXiv:2407.19325},
  eprint = {2407.19325},
  primaryclass = {cs},
  publisher = {arXiv},
  doi = {10.48550/arXiv.2407.19325},
  urldate = {2025-08-10},
  archiveprefix = {arXiv},
  langid = {english}
}

@inproceedings{dankersetal2022,
  title = {Can {{Transformer}} Be {{Too Compositional}}? {{Analysing Idiom Processing}} in {{Neural Machine Translation}}},
  shorttitle = {Can {{Transformer}} Be {{Too Compositional}}?},
  booktitle = {Proceedings of the 60th {{Annual Meeting}} of the {{Association}} for {{Computational Linguistics}} ({{Volume}} 1: {{Long Papers}})},
  author = {Dankers, Verna and Lucas, Christopher and Titov, Ivan},
  year = 2022,
  pages = {3608--3626},
  publisher = {Association for Computational Linguistics},
  address = {Dublin, Ireland},
  doi = {10.18653/v1/2022.acl-long.252},
  urldate = {2023-04-30},
  langid = {english}
}

@article{darmasetiyawanetal2022,
  title = {Is {{Passive Priming Really Impervious}} to {{Verb Semantics}}? {{A High-Powered Replication}} of {{Messenger Et}} al. (2012)},
  shorttitle = {Is {{Passive Priming Really Impervious}} to {{Verb Semantics}}?},
  author = {Darmasetiyawan, I Made Sena and Messenger, Kate and Ambridge, Ben},
  year = 2022,
  month = jan,
  journal = {Collabra: Psychology},
  volume = {8},
  number = {1},
  pages = {31055},
  issn = {2474-7394},
  doi = {10.1525/collabra.31055},
  urldate = {2022-10-28},
  langid = {english}
}

@incollection{demuth2011,
  title = {The Role of Frequency in Language Acquisition},
  booktitle = {The Role of Frequency in Language Acquisition},
  author = {Demuth, Katherine},
  year = 2011,
  month = may,
  pages = {383--388},
  publisher = {De Gruyter Mouton},
  doi = {10.1515/9783110977905.383},
  urldate = {2023-04-26},
  isbn = {978-3-11-097790-5},
  langid = {english}
}

@article{foxgrodzinsky1998,
  title = {Children's {{Passive}}: {{A View}} from the {{By-Phrase}}},
  shorttitle = {Children's {{Passive}}},
  author = {Fox, Danny and Grodzinsky, Yosef},
  year = 1998,
  journal = {Linguistic Inquiry},
  volume = {29},
  number = {2},
  eprint = {4179020},
  eprinttype = {jstor},
  pages = {311--332},
  publisher = {The MIT Press},
  issn = {0024-3892},
  urldate = {2023-03-15}
}

@article{gibsonetal2013,
  title = {A {{Noisy-Channel Account}} of {{Crosslinguistic Word-Order Variation}}},
  author = {Gibson, Edward and Piantadosi, Steven T. and Brink, Kimberly and Bergen, Leon and Lim, Eunice and Saxe, Rebecca},
  year = 2013,
  month = jul,
  journal = {Psychological Science},
  volume = {24},
  number = {7},
  pages = {1079--1088},
  issn = {0956-7976, 1467-9280},
  doi = {10.1177/0956797612463705},
  urldate = {2024-05-31},
  langid = {english}
}

@article{gilkersonetal2017,
  title = {Mapping the {{Early Language Environment Using All-Day Recordings}} and {{Automated Analysis}}},
  author = {Gilkerson, Jill and Richards, Jeffrey A. and Warren, Steven F. and Montgomery, Judith K. and Greenwood, Charles R. and Kimbrough, Oller D. and Hansen, John H. L. and Paul, Terrance D.},
  year = 2017,
  month = may,
  journal = {American Journal of Speech-Language Pathology},
  volume = {26},
  number = {2},
  pages = {248--265},
  publisher = {American Speech-Language-Hearing Association},
  doi = {10.1044/2016_AJSLP-15-0169},
  urldate = {2023-04-30}
}

@misc{gokaslancohen2019,
  title = {{{OpenWebText}} Corpus},
  author = {Gokaslan, Aaron and Cohen, Vanya},
  year = 2019
}

@book{goldberg1995,
  title = {Constructions: A Construction Grammar Approach to Argument Structure},
  shorttitle = {Constructions},
  author = {Goldberg, Adele E.},
  year = 1995,
  series = {Cognitive Theory of Language and Culture},
  publisher = {University of Chicago Press},
  address = {Chicago},
  isbn = {978-0-226-30085-6 978-0-226-30086-3},
  lccn = {P291 .G65 1995}
}

@book{goldberg2006,
  title = {Constructions at Work: The Nature of Generalization in Language},
  shorttitle = {Constructions at Work},
  author = {Goldberg, Adele E.},
  year = 2006,
  series = {Oxford Linguistics},
  edition = {1. publ},
  publisher = {Oxford University Press},
  address = {Oxford},
  isbn = {978-0-19-926851-1 978-0-19-926852-8},
  langid = {english}
}

@article{gordonchafetz1990,
  title = {Verb-Based versus Class-Based Accounts of Actionality Effects in Children's Comprehension of Passives},
  author = {Gordon, Peter and Chafetz, Jill},
  year = 1990,
  month = sep,
  journal = {Cognition},
  volume = {36},
  number = {3},
  pages = {227--254},
  issn = {0010-0277},
  doi = {10.1016/0010-0277(90)90058-R},
  urldate = {2023-02-16},
  langid = {english}
}

@inproceedings{gulordavaetal2018,
  title = {Colorless {{Green Recurrent Networks Dream Hierarchically}}},
  booktitle = {Proceedings of the 2018 {{Conference}} of the {{North American Chapter}} of the {{Association}} for {{Computational Linguistics}}: {{Human Language Technologies}}, {{Volume}} 1 ({{Long Papers}})},
  author = {Gulordava, Kristina and Bojanowski, Piotr and Grave, Edouard and Linzen, Tal and Baroni, Marco},
  editor = {Walker, Marilyn and Ji, Heng and Stent, Amanda},
  year = 2018,
  month = jun,
  pages = {1195--1205},
  publisher = {Association for Computational Linguistics},
  address = {New Orleans, Louisiana},
  doi = {10.18653/v1/N18-1108},
  urldate = {2024-06-29}
}

@inproceedings{hawkinsetal2020,
  title = {Investigating Representations of Verb Bias in Neural Language Models},
  booktitle = {Proceedings of the 2020 {{Conference}} on {{Empirical Methods}} in {{Natural Language Processing}} ({{EMNLP}})},
  author = {Hawkins, Robert and Yamakoshi, Takateru and Griffiths, Thomas and Goldberg, Adele},
  year = 2020,
  pages = {4653--4663},
  publisher = {Association for Computational Linguistics},
  address = {Online},
  doi = {10.18653/v1/2020.emnlp-main.376},
  urldate = {2023-02-16},
  langid = {english}
}

@inproceedings{heafield2011,
  title = {{{KenLM}}: {{Faster}} and Smaller Language Model Queries},
  booktitle = {Proceedings of the Sixth Workshop on Statistical Machine Translation},
  author = {Heafield, Kenneth},
  year = 2011,
  month = jul,
  pages = {187--197},
  publisher = {Association for Computational Linguistics},
  address = {Edinburgh, Scotland},
  month_numeric = {7}
}

@misc{honnibaletal2020,
  title = {{{spaCy}}: {{Industrial-strength}} Natural Language Processing in Python},
  author = {Honnibal, Matthew and Montani, Ines and Van Landeghem, Sofie and Boyd, Adriane},
  year = 2020
}

@misc{jumeletetal2021,
  title = {Language {{Models Use Monotonicity}} to {{Assess NPI Licensing}}},
  author = {Jumelet, Jaap and Deni{\'c}, Milica and Szymanik, Jakub and Hupkes, Dieuwke and {Steinert-Threlkeld}, Shane},
  year = 2021,
  month = may,
  number = {arXiv:2105.13818},
  eprint = {2105.13818},
  primaryclass = {cs},
  publisher = {arXiv},
  doi = {10.48550/arXiv.2105.13818},
  urldate = {2024-12-08},
  archiveprefix = {arXiv}
}

@incollection{keenandryer2007,
  title = {Passive in the World's Languages},
  booktitle = {Language {{Typology}} and {{Syntactic Description}}},
  author = {Keenan, Edward L. and Dryer, Matthew S.},
  editor = {Shopen, Timothy},
  year = 2007,
  month = oct,
  edition = {2},
  pages = {325--361},
  publisher = {Cambridge University Press},
  doi = {10.1017/CBO9780511619427.006},
  urldate = {2023-04-29},
  isbn = {978-0-521-58156-1 978-0-521-58857-7 978-0-511-61942-7},
  langid = {english}
}

@inproceedings{kingmaba2015,
  title = {Adam: {{A Method}} for {{Stochastic Optimization}}},
  shorttitle = {Adam},
  booktitle = {{{arXiv}}:1412.6980 [Cs]},
  author = {Kingma, Diederik P. and Ba, Jimmy},
  year = 2015,
  eprint = {1412.6980},
  primaryclass = {cs},
  urldate = {2019-12-01},
  archiveprefix = {arXiv}
}

@article{lakeetal2017,
  title = {Building Machines That Learn and Think like People},
  author = {Lake, Brenden M. and Ullman, Tomer D. and Tenenbaum, Joshua B. and Gershman, Samuel J.},
  year = 2017,
  journal = {Behavioral and Brain Sciences},
  volume = {40},
  pages = {e253},
  issn = {0140-525X, 1469-1825},
  doi = {10.1017/S0140525X16001837},
  urldate = {2024-05-31},
  copyright = {https://www.cambridge.org/core/terms},
  langid = {english}
}

@article{lakretzetal2021,
  title = {Mechanisms for Handling Nested Dependencies in Neural-Network Language Models and Humans},
  author = {Lakretz, Yair and Hupkes, Dieuwke and Vergallito, Alessandra and Marelli, Marco and Baroni, Marco and Dehaene, Stanislas},
  year = 2021,
  month = aug,
  journal = {Cognition},
  volume = {213},
  pages = {104699},
  issn = {1873-7838},
  doi = {10.1016/j.cognition.2021.104699},
  langid = {english},
  pmid = {33941375}
}

@article{lauetal2017,
  title = {Grammaticality, {{Acceptability}}, and {{Probability}}: {{A Probabilistic View}} of {{Linguistic Knowledge}}},
  shorttitle = {Grammaticality, {{Acceptability}}, and {{Probability}}},
  author = {Lau, Jey Han and Clark, Alexander and Lappin, Shalom},
  year = 2017,
  month = jul,
  journal = {Cognitive Science},
  volume = {41},
  number = {5},
  pages = {1202--1241},
  issn = {03640213},
  doi = {10.1111/cogs.12414},
  urldate = {2022-10-13},
  langid = {english}
}

@inproceedings{leonglinzen2023,
  title = {Language {{Models Can Learn Exceptions}} to {{Syntactic Rules}}},
  booktitle = {Proceedings of the {{Society}} for {{Computation}} in {{Linguistics}}},
  author = {Leong, Cara Su-Yi and Linzen, Tal},
  year = 2023,
  volume = {6},
  publisher = {University of Massachusetts Amherst},
  address = {Amherst},
  doi = {10.7275/H25Z-0Y75},
  urldate = {2023-06-16}
}

@book{levin1993,
  title = {English Verb Classes and Alternations: A Preliminary Investigation},
  shorttitle = {English Verb Classes and Alternations},
  author = {Levin, Beth},
  year = 1993,
  publisher = {University of Chicago Press},
  address = {Chicago},
  isbn = {978-0-226-47532-5 978-0-226-47533-2},
  lccn = {PE1271 .L48 1993}
}

@inproceedings{levy2008,
  title = {A Noisy-Channel Model of Rational Human Sentence Comprehension under Uncertain Input},
  booktitle = {Proceedings of the {{Conference}} on {{Empirical Methods}} in {{Natural Language Processing}} - {{EMNLP}} '08},
  author = {Levy, Roger},
  year = 2008,
  pages = {234},
  publisher = {Association for Computational Linguistics},
  address = {Honolulu, Hawaii},
  doi = {10.3115/1613715.1613749},
  urldate = {2024-11-25},
  langid = {english}
}

@inproceedings{linzen2020,
  title = {How {{Can We Accelerate Progress Towards Human-like Linguistic Generalization}}?},
  booktitle = {Proceedings of the 58th {{Annual Meeting}} of the {{Association}} for {{Computational Linguistics}}},
  author = {Linzen, Tal},
  year = 2020,
  month = jul,
  pages = {5210--5217},
  publisher = {Association for Computational Linguistics},
  address = {Online},
  doi = {10.18653/v1/2020.acl-main.465},
  urldate = {2023-05-15}
}

@article{linzenetal2016,
  title = {Assessing the {{Ability}} of {{LSTMs}} to {{Learn Syntax-Sensitive Dependencies}}},
  author = {Linzen, Tal and Dupoux, Emmanuel and Goldberg, Yoav},
  year = 2016,
  journal = {Transactions of the Association for Computational Linguistics},
  volume = {4},
  pages = {521--535},
  publisher = {MIT Press},
  address = {Cambridge, MA},
  doi = {10.1162/tacl_a_00115},
  urldate = {2023-04-28}
}

@article{liuambridge2021,
  title = {Balancing Information-Structure and Semantic Constraints on Construction Choice: Building a Computational Model of Passive and Passive-like Constructions in {{Mandarin Chinese}}},
  shorttitle = {Balancing Information-Structure and Semantic Constraints on Construction Choice},
  author = {Liu, Li and Ambridge, Ben},
  year = 2021,
  month = sep,
  journal = {Cognitive Linguistics},
  volume = {32},
  number = {3},
  pages = {349--388},
  issn = {0936-5907, 1613-3641},
  doi = {10.1515/cog-2019-0100},
  urldate = {2024-08-14},
  copyright = {http://creativecommons.org/licenses/by/4.0},
  langid = {english}
}

@book{macwhinney2000,
  title = {The {{CHILDES Project}}: {{Tools}} for Analyzing Talk. {{Third Edition}}.},
  author = {MacWhinney, Brian},
  year = 2000,
  publisher = {Lawrence Erlbaum Associates},
  address = {Mahwah, NJ}
}

@article{maratsos1985,
  title = {Semantic Restrictions on Children's Passives},
  author = {Maratsos, M},
  year = 1985,
  journal = {Cognition},
  volume = {19},
  number = {2},
  pages = {167--191},
  issn = {00100277},
  doi = {10.1016/0010-0277(85)90017-4},
  urldate = {2022-10-30},
  langid = {english}
}

@inproceedings{marvinlinzen2018,
  title = {Targeted {{Syntactic Evaluation}} of {{Language Models}}},
  booktitle = {Proceedings of the 2018 {{Conference}} on {{Empirical Methods}} in {{Natural Language Processing}}},
  author = {Marvin, Rebecca and Linzen, Tal},
  editor = {Riloff, Ellen and Chiang, David and Hockenmaier, Julia and Tsujii, Jun'ichi},
  year = 2018,
  month = oct,
  pages = {1192--1202},
  publisher = {Association for Computational Linguistics},
  address = {Brussels, Belgium},
  doi = {10.18653/v1/D18-1151},
  urldate = {2024-06-29}
}

@article{messengeretal2012,
  title = {Is Young Children's Passive Syntax Semantically Constrained? {{Evidence}} from Syntactic Priming},
  shorttitle = {Is Young Children's Passive Syntax Semantically Constrained?},
  author = {Messenger, Katherine and Branigan, Holly P. and McLean, Janet F. and Sorace, Antonella},
  year = 2012,
  month = may,
  journal = {Journal of Memory and Language},
  volume = {66},
  number = {4},
  pages = {568--587},
  issn = {0749596X},
  doi = {10.1016/j.jml.2012.03.008},
  urldate = {2022-10-30},
  langid = {english}
}

@misc{misramahowald2024,
  title = {Language {{Models Learn Rare Phenomena}} from {{Less Rare Phenomena}}: {{The Case}} of the {{Missing AANNs}}},
  shorttitle = {Language {{Models Learn Rare Phenomena}} from {{Less Rare Phenomena}}},
  author = {Misra, Kanishka and Mahowald, Kyle},
  year = 2024,
  month = mar,
  number = {arXiv:2403.19827},
  eprint = {2403.19827},
  primaryclass = {cs},
  publisher = {arXiv},
  urldate = {2024-04-16},
  archiveprefix = {arXiv}
}

@misc{openaietal2024,
  title = {{{GPT-4o System Card}}},
  author = {OpenAI and Hurst, Aaron and Lerer, Adam and Goucher, Adam P. and Perelman, Adam and Ramesh, Aditya and Clark, Aidan and Ostrow, A. J. and Welihinda, Akila and Hayes, Alan and Radford, Alec and M{\k a}dry, Aleksander and {Baker-Whitcomb}, Alex and Beutel, Alex and Borzunov, Alex and Carney, Alex and Chow, Alex and Kirillov, Alex and Nichol, Alex and Paino, Alex and Renzin, Alex and Passos, Alex Tachard and Kirillov, Alexander and Christakis, Alexi and Conneau, Alexis and Kamali, Ali and Jabri, Allan and Moyer, Allison and Tam, Allison and Crookes, Amadou and Tootoochian, Amin and Tootoonchian, Amin and Kumar, Ananya and Vallone, Andrea and Karpathy, Andrej and Braunstein, Andrew and Cann, Andrew and Codispoti, Andrew and Galu, Andrew and Kondrich, Andrew and Tulloch, Andrew and Mishchenko, Andrey and Baek, Angela and Jiang, Angela and Pelisse, Antoine and Woodford, Antonia and Gosalia, Anuj and Dhar, Arka and Pantuliano, Ashley and Nayak, Avi and Oliver, Avital and Zoph, Barret and Ghorbani, Behrooz and Leimberger, Ben and Rossen, Ben and Sokolowsky, Ben and Wang, Ben and Zweig, Benjamin and Hoover, Beth and Samic, Blake and McGrew, Bob and Spero, Bobby and Giertler, Bogo and Cheng, Bowen and Lightcap, Brad and Walkin, Brandon and Quinn, Brendan and Guarraci, Brian and Hsu, Brian and Kellogg, Bright and Eastman, Brydon and Lugaresi, Camillo and Wainwright, Carroll and Bassin, Cary and Hudson, Cary and Chu, Casey and Nelson, Chad and Li, Chak and Shern, Chan Jun and Conger, Channing and Barette, Charlotte and Voss, Chelsea and Ding, Chen and Lu, Cheng and Zhang, Chong and Beaumont, Chris and Hallacy, Chris and Koch, Chris and Gibson, Christian and Kim, Christina and Choi, Christine and McLeavey, Christine and Hesse, Christopher and Fischer, Claudia and Winter, Clemens and Czarnecki, Coley and Jarvis, Colin and Wei, Colin and Koumouzelis, Constantin and Sherburn, Dane and Kappler, Daniel and Levin, Daniel and Levy, Daniel and Carr, David and Farhi, David and Mely, David and Robinson, David and Sasaki, David and Jin, Denny and Valladares, Dev and Tsipras, Dimitris and Li, Doug and Nguyen, Duc Phong and Findlay, Duncan and Oiwoh, Edede and Wong, Edmund and Asdar, Ehsan and Proehl, Elizabeth and Yang, Elizabeth and Antonow, Eric and Kramer, Eric and Peterson, Eric and Sigler, Eric and Wallace, Eric and Brevdo, Eugene and Mays, Evan and Khorasani, Farzad and Such, Felipe Petroski and Raso, Filippo and Zhang, Francis and von Lohmann, Fred and Sulit, Freddie and Goh, Gabriel and Oden, Gene and Salmon, Geoff and Starace, Giulio and Brockman, Greg and Salman, Hadi and Bao, Haiming and Hu, Haitang and Wong, Hannah and Wang, Haoyu and Schmidt, Heather and Whitney, Heather and Jun, Heewoo and Kirchner, Hendrik and Pinto, Henrique Ponde de Oliveira and Ren, Hongyu and Chang, Huiwen and Chung, Hyung Won and Kivlichan, Ian and O'Connell, Ian and O'Connell, Ian and Osband, Ian and Silber, Ian and Sohl, Ian and Okuyucu, Ibrahim and Lan, Ikai and Kostrikov, Ilya and Sutskever, Ilya and Kanitscheider, Ingmar and Gulrajani, Ishaan and Coxon, Jacob and Menick, Jacob and Pachocki, Jakub and Aung, James and Betker, James and Crooks, James and Lennon, James and Kiros, Jamie and Leike, Jan and Park, Jane and Kwon, Jason and Phang, Jason and Teplitz, Jason and Wei, Jason and Wolfe, Jason and Chen, Jay and Harris, Jeff and Varavva, Jenia and Lee, Jessica Gan and Shieh, Jessica and Lin, Ji and Yu, Jiahui and Weng, Jiayi and Tang, Jie and Yu, Jieqi and Jang, Joanne and Candela, Joaquin Quinonero and Beutler, Joe and Landers, Joe and Parish, Joel and Heidecke, Johannes and Schulman, John and Lachman, Jonathan and McKay, Jonathan and Uesato, Jonathan and Ward, Jonathan and Kim, Jong Wook and Huizinga, Joost and Sitkin, Jordan and Kraaijeveld, Jos and Gross, Josh and Kaplan, Josh and Snyder, Josh and Achiam, Joshua and Jiao, Joy and Lee, Joyce and Zhuang, Juntang and Harriman, Justyn and Fricke, Kai and Hayashi, Kai and Singhal, Karan and Shi, Katy and Karthik, Kavin and Wood, Kayla and Rimbach, Kendra and Hsu, Kenny and Nguyen, Kenny and {Gu-Lemberg}, Keren and Button, Kevin and Liu, Kevin and Howe, Kiel and Muthukumar, Krithika and Luther, Kyle and Ahmad, Lama and Kai, Larry and Itow, Lauren and Workman, Lauren and Pathak, Leher and Chen, Leo and Jing, Li and Guy, Lia and Fedus, Liam and Zhou, Liang and Mamitsuka, Lien and Weng, Lilian and McCallum, Lindsay and Held, Lindsey and Ouyang, Long and Feuvrier, Louis and Zhang, Lu and Kondraciuk, Lukas and Kaiser, Lukasz and Hewitt, Luke and Metz, Luke and Doshi, Lyric and Aflak, Mada and Simens, Maddie and Boyd, Madelaine and Thompson, Madeleine and Dukhan, Marat and Chen, Mark and Gray, Mark and Hudnall, Mark and Zhang, Marvin and Aljubeh, Marwan and Litwin, Mateusz and Zeng, Matthew and Johnson, Max and Shetty, Maya and Gupta, Mayank and Shah, Meghan and Yatbaz, Mehmet and Yang, Meng Jia and Zhong, Mengchao and Glaese, Mia and Chen, Mianna and Janner, Michael and Lampe, Michael and Petrov, Michael and Wu, Michael and Wang, Michele and Fradin, Michelle and Pokrass, Michelle and Castro, Miguel and de Castro, Miguel Oom Temudo and Pavlov, Mikhail and Brundage, Miles and Wang, Miles and Khan, Minal and Murati, Mira and Bavarian, Mo and Lin, Molly and Yesildal, Murat and Soto, Nacho and Gimelshein, Natalia and Cone, Natalie and Staudacher, Natalie and Summers, Natalie and LaFontaine, Natan and Chowdhury, Neil and Ryder, Nick and Stathas, Nick and Turley, Nick and Tezak, Nik and Felix, Niko and Kudige, Nithanth and Keskar, Nitish and Deutsch, Noah and Bundick, Noel and Puckett, Nora and Nachum, Ofir and Okelola, Ola and Boiko, Oleg and Murk, Oleg and Jaffe, Oliver and Watkins, Olivia and Godement, Olivier and {Campbell-Moore}, Owen and Chao, Patrick and McMillan, Paul and Belov, Pavel and Su, Peng and Bak, Peter and Bakkum, Peter and Deng, Peter and Dolan, Peter and Hoeschele, Peter and Welinder, Peter and Tillet, Phil and Pronin, Philip and Tillet, Philippe and Dhariwal, Prafulla and Yuan, Qiming and Dias, Rachel and Lim, Rachel and Arora, Rahul and Troll, Rajan and Lin, Randall and Lopes, Rapha Gontijo and Puri, Raul and Miyara, Reah and Leike, Reimar and Gaubert, Renaud and Zamani, Reza and Wang, Ricky and Donnelly, Rob and Honsby, Rob and Smith, Rocky and Sahai, Rohan and Ramchandani, Rohit and Huet, Romain and Carmichael, Rory and Zellers, Rowan and Chen, Roy and Chen, Ruby and Nigmatullin, Ruslan and Cheu, Ryan and Jain, Saachi and Altman, Sam and Schoenholz, Sam and Toizer, Sam and Miserendino, Samuel and Agarwal, Sandhini and Culver, Sara and Ethersmith, Scott and Gray, Scott and Grove, Sean and Metzger, Sean and Hermani, Shamez and Jain, Shantanu and Zhao, Shengjia and Wu, Sherwin and Jomoto, Shino and Wu, Shirong and Shuaiqi and Xia and Phene, Sonia and Papay, Spencer and Narayanan, Srinivas and Coffey, Steve and Lee, Steve and Hall, Stewart and Balaji, Suchir and Broda, Tal and Stramer, Tal and Xu, Tao and Gogineni, Tarun and Christianson, Taya and Sanders, Ted and Patwardhan, Tejal and Cunninghman, Thomas and Degry, Thomas and Dimson, Thomas and Raoux, Thomas and Shadwell, Thomas and Zheng, Tianhao and Underwood, Todd and Markov, Todor and Sherbakov, Toki and Rubin, Tom and Stasi, Tom and Kaftan, Tomer and Heywood, Tristan and Peterson, Troy and Walters, Tyce and Eloundou, Tyna and Qi, Valerie and Moeller, Veit and Monaco, Vinnie and Kuo, Vishal and Fomenko, Vlad and Chang, Wayne and Zheng, Weiyi and Zhou, Wenda and Manassra, Wesam and Sheu, Will and Zaremba, Wojciech and Patil, Yash and Qian, Yilei and Kim, Yongjik and Cheng, Youlong and Zhang, Yu and He, Yuchen and Zhang, Yuchen and Jin, Yujia and Dai, Yunxing and Malkov, Yury},
  year = 2024,
  month = oct,
  number = {arXiv:2410.21276},
  eprint = {2410.21276},
  primaryclass = {cs},
  publisher = {arXiv},
  doi = {10.48550/arXiv.2410.21276},
  urldate = {2024-12-03},
  archiveprefix = {arXiv}
}

@inproceedings{papadimitriouetal2021,
  title = {Deep {{Subjecthood}}: {{Higher-Order Grammatical Features}} in {{Multilingual BERT}}},
  shorttitle = {Deep {{Subjecthood}}},
  booktitle = {Proceedings of the 16th {{Conference}} of the {{European Chapter}} of the {{Association}} for {{Computational Linguistics}}: {{Main Volume}}},
  author = {Papadimitriou, Isabel and Chi, Ethan A. and Futrell, Richard and Mahowald, Kyle},
  editor = {Merlo, Paola and Tiedemann, Jorg and Tsarfaty, Reut},
  year = 2021,
  month = apr,
  pages = {2522--2532},
  publisher = {Association for Computational Linguistics},
  address = {Online},
  doi = {10.18653/v1/2021.eacl-main.215},
  urldate = {2025-08-10}
}

@inproceedings{papadimitriouetal2023,
  title = {Multilingual {{BERT}} Has an Accent: {{Evaluating English}} Influences on Fluency in Multilingual Models},
  shorttitle = {Multilingual {{BERT}} Has an Accent},
  booktitle = {Findings of the {{Association}} for {{Computational Linguistics}}: {{EACL}} 2023},
  author = {Papadimitriou, Isabel and Lopez, Kezia and Jurafsky, Dan},
  editor = {Vlachos, Andreas and Augenstein, Isabelle},
  year = 2023,
  month = may,
  pages = {1194--1200},
  publisher = {Association for Computational Linguistics},
  address = {Dubrovnik, Croatia},
  doi = {10.18653/v1/2023.findings-eacl.89},
  urldate = {2025-08-10}
}

@misc{patiletal2024,
  title = {Filtered {{Corpus Training}} ({{FiCT}}) {{Shows}} That {{Language Models}} Can {{Generalize}} from {{Indirect Evidence}}},
  author = {Patil, Abhinav and Jumelet, Jaap and Chiu, Yu Ying and Lapastora, Andy and Shen, Peter and Wang, Lexie and Willrich, Clevis and {Steinert-Threlkeld}, Shane},
  year = 2024,
  month = may,
  number = {arXiv:2405.15750},
  eprint = {2405.15750},
  primaryclass = {cs},
  publisher = {arXiv},
  urldate = {2024-06-11},
  archiveprefix = {arXiv},
  langid = {english}
}

@article{perforsetal2010,
  title = {Variability, Negative Evidence, and the Acquisition of Verb Argument Constructions},
  author = {Perfors, Amy and Tenenbaum, Joshua B. and Wonnacott, Elizabeth},
  year = 2010,
  month = jun,
  journal = {Journal of Child Language},
  volume = {37},
  number = {3},
  pages = {607--642},
  issn = {0305-0009, 1469-7602},
  doi = {10.1017/S0305000910000012},
  urldate = {2022-10-13},
  langid = {english}
}

@book{pinker1989,
  title = {Learnability and Cognition: The Acquisition of Argument Structure},
  shorttitle = {Learnability and Cognition},
  author = {Pinker, Steven},
  year = 1989,
  series = {Learning, Development, and Conceptual Change},
  edition = {1. paperback ed., 4. print},
  publisher = {MIT Press},
  address = {Cambridge, Mass.},
  isbn = {978-0-262-66073-0 978-0-262-16111-4},
  langid = {english}
}

@article{pinkeretal1987,
  title = {Productivity and Constraints in the Acquisition of the Passive},
  author = {Pinker, Steven and Lebeaux, David S. and Frost, Loren Ann},
  year = 1987,
  month = aug,
  journal = {Cognition},
  volume = {26},
  number = {3},
  pages = {195--267},
  issn = {0010-0277},
  doi = {10.1016/S0010-0277(87)80001-X},
  urldate = {2023-02-10},
  langid = {english}
}

@book{postal2004,
  title = {Skeptical Linguistic Essays},
  author = {Postal, Paul Martin},
  year = 2004,
  publisher = {Oxford University Press},
  address = {Oxford ; New York},
  isbn = {978-0-19-516672-9 978-0-19-516671-2},
  langid = {english},
  lccn = {P151 .P655 2004}
}

@techreport{radfordetal2019,
  title = {Language {{Models}} Are {{Unsupervised Multitask Learners}}},
  author = {Radford, Alec and Wu, Jeffrey and Child, Rewon and Luan, David and Amodei, Dario and Sutskever, Ilya},
  year = 2019,
  pages = {24},
  institution = {OpenAI},
  langid = {english}
}

@article{regiergahl2004,
  title = {Learning the Unlearnable: The Role of Missing Evidence},
  shorttitle = {Learning the Unlearnable},
  author = {Regier, Terry and Gahl, Susanne},
  year = 2004,
  month = sep,
  journal = {Cognition},
  volume = {93},
  number = {2},
  pages = {147--155},
  issn = {0010-0277},
  doi = {10.1016/j.cognition.2003.12.003},
  urldate = {2023-02-03},
  langid = {english}
}

@article{reisingeretal2015,
  title = {Semantic {{Proto-Roles}}},
  author = {Reisinger, Drew and Rudinger, Rachel and Ferraro, Francis and Harman, Craig and Rawlins, Kyle and Van Durme, Benjamin},
  year = 2015,
  month = dec,
  journal = {Transactions of the Association for Computational Linguistics},
  volume = {3},
  pages = {475--488},
  issn = {2307-387X},
  doi = {10.1162/tacl_a_00152},
  urldate = {2022-07-13},
  langid = {english}
}

@article{rolandetal2007,
  title = {Frequency of Basic {{English}} Grammatical Structures: {{A}} Corpus Analysis},
  shorttitle = {Frequency of Basic {{English}} Grammatical Structures},
  author = {Roland, Douglas and Dick, Frederic and Elman, Jeffrey L.},
  year = 2007,
  month = oct,
  journal = {Journal of Memory and Language},
  volume = {57},
  number = {3},
  pages = {348--379},
  issn = {0749596X},
  doi = {10.1016/j.jml.2007.03.002},
  urldate = {2022-07-20},
  langid = {english}
}

@article{saffranetal1996,
  title = {Statistical {{Learning}} by 8-{{Month-Old Infants}}},
  author = {Saffran, Jenny R. and Aslin, Richard N. and Newport, Elissa L.},
  year = 1996,
  month = dec,
  journal = {Science},
  volume = {274},
  number = {5294},
  pages = {1926--1928},
  publisher = {American Association for the Advancement of Science},
  doi = {10.1126/science.274.5294.1926},
  urldate = {2023-03-01}
}

@article{spearman1910,
  title = {Correlation {{Calculated}} from {{Faulty Data}}},
  author = {Spearman, C.},
  year = 1910,
  journal = {British Journal of Psychology, 1904-1920},
  volume = {3},
  number = {3},
  pages = {271--295},
  issn = {2044-8295},
  doi = {10.1111/j.2044-8295.1910.tb00206.x},
  urldate = {2024-06-29},
  copyright = {1910 The British Psychological Society},
  langid = {english}
}

@article{sprousealmeida2017,
  title = {Design Sensitivity and Statistical Power in Acceptability Judgment Experiments},
  author = {Sprouse, Jon and Almeida, Diogo},
  year = 2017,
  month = feb,
  journal = {Glossa: a journal of general linguistics},
  volume = {2},
  number = {1},
  issn = {2397-1835},
  doi = {10.5334/gjgl.236},
  urldate = {2024-06-21},
  copyright = {https://creativecommons.org/licenses/by/4.0},
  langid = {english}
}

@article{theakston2004,
  title = {The Role of Entrenchment in Children's and Adults' Performance on Grammaticality Judgment Tasks},
  author = {Theakston, Anna L},
  year = 2004,
  month = jan,
  journal = {Cognitive Development},
  volume = {19},
  number = {1},
  pages = {15--34},
  issn = {0885-2014},
  doi = {10.1016/j.cogdev.2003.08.001},
  urldate = {2023-04-26},
  langid = {english}
}

@article{thompsonnewport2007,
  title = {Statistical {{Learning}} of {{Syntax}}: {{The Role}} of {{Transitional Probability}}},
  shorttitle = {Statistical {{Learning}} of {{Syntax}}},
  author = {Thompson, Susan P. and Newport, Elissa L.},
  year = 2007,
  journal = {Language Learning and Development},
  volume = {3},
  number = {1},
  pages = {1--42},
  publisher = {Taylor \& Francis},
  address = {United Kingdom},
  issn = {1547-3341},
  doi = {10.1207/s15473341lld0301_1}
}

@inproceedings{timkeylinzen2023,
  title = {A Language Model with Limited Memory Capacity Captures Interference in Human Sentence Processing},
  booktitle = {Findings of the Association for Computational Linguistics: {{EMNLP}} 2023},
  author = {Timkey, William and Linzen, Tal},
  editor = {Bouamor, Houda and Pino, Juan and Bali, Kalika},
  year = 2023,
  month = dec,
  pages = {8705--8720},
  publisher = {Association for Computational Linguistics},
  address = {Singapore},
  doi = {10.18653/v1/2023.findings-emnlp.582}
}

@inproceedings{tjuatjaetal2025,
  title = {What {{Goes Into}} a {{LM Acceptability Judgment}}? {{Rethinking}} the {{Impact}} of {{Frequency}} and {{Length}}},
  shorttitle = {What {{Goes Into}} a {{LM Acceptability Judgment}}?},
  booktitle = {Proceedings of the 2025 {{Conference}} of the {{Nations}} of the {{Americas Chapter}} of the {{Association}} for {{Computational Linguistics}}: {{Human Language Technologies}} ({{Volume}} 1: {{Long Papers}})},
  author = {Tjuatja, Lindia and Neubig, Graham and Linzen, Tal and Hao, Sophie},
  editor = {Chiruzzo, Luis and Ritter, Alan and Wang, Lu},
  year = 2025,
  month = apr,
  pages = {2173--2186},
  publisher = {Association for Computational Linguistics},
  address = {Albuquerque, New Mexico},
  doi = {10.18653/v1/2025.naacl-long.109},
  urldate = {2025-11-20},
  isbn = {979-8-89176-189-6}
}

@article{tomasello2000,
  title = {Do Young Children Have Adult Syntactic Competence?},
  author = {Tomasello, Michael},
  year = 2000,
  month = mar,
  journal = {Cognition},
  volume = {74},
  number = {3},
  pages = {209--253},
  issn = {0010-0277},
  doi = {10.1016/S0010-0277(99)00069-4},
  urldate = {2024-06-13}
}

@inproceedings{vaswanietal2017,
  title = {Attention {{Is All You Need}}},
  booktitle = {{{arXiv}}:1706.03762 [Cs]},
  author = {Vaswani, Ashish and Shazeer, Noam and Parmar, Niki and Uszkoreit, Jakob and Jones, Llion and Gomez, Aidan N. and Kaiser, Lukasz and Polosukhin, Illia},
  year = 2017,
  month = jun,
  eprint = {1706.03762},
  primaryclass = {cs},
  urldate = {2019-05-06},
  archiveprefix = {arXiv}
}

@article{vongetal2024,
  title = {Grounded Language Acquisition through the Eyes and Ears of a Single Child},
  author = {Vong, Wai Keen and Wang, Wentao and Orhan, A. Emin and Lake, Brenden M.},
  year = 2024,
  month = feb,
  journal = {Science},
  publisher = {American Association for the Advancement of Science},
  doi = {10.1126/science.adi1374},
  urldate = {2024-05-31},
  copyright = {Copyright \copyright{} 2024 The Authors, some rights reserved; exclusive licensee American Association for the Advancement of Science. No claim to original U.S. Government Works},
  langid = {english}
}

@incollection{warstadtbowman2022,
  title = {What Artificial Neural Networks Can Tell Us about Human Language Acquisition},
  booktitle = {Algebraic Structures in Natural Language},
  author = {Warstadt, Alex and Bowman, Samuel R.},
  year = 2022,
  pages = {17--60},
  publisher = {CRC Press}
}

@article{warstadtetal2020,
  title = {{{BLiMP}}: {{The Benchmark}} of {{Linguistic Minimal Pairs}} for {{English}}},
  shorttitle = {{{BLiMP}}},
  author = {Warstadt, Alex and Parrish, Alicia and Liu, Haokun and Mohananey, Anhad and Peng, Wei and Wang, Sheng-Fu and Bowman, Samuel R.},
  year = 2020,
  month = dec,
  journal = {Transactions of the Association for Computational Linguistics},
  volume = {8},
  pages = {377--392},
  issn = {2307-387X},
  doi = {10.1162/tacl_a_00321},
  urldate = {2023-04-28},
  langid = {english}
}

@inproceedings{warstadtetal2023,
  title = {Findings of the {{BabyLM Challenge}}: {{Sample-Efficient Pretraining}} on {{Developmentally Plausible Corpora}}},
  shorttitle = {Findings of the {{BabyLM Challenge}}},
  booktitle = {Proceedings of the {{BabyLM Challenge}} at the 27th {{Conference}} on {{Computational Natural Language Learning}}},
  author = {Warstadt, Alex and Mueller, Aaron and Choshen, Leshem and Wilcox, Ethan and Zhuang, Chengxu and Ciro, Juan and Mosquera, Rafael and Paranjabe, Bhargavi and Williams, Adina and Linzen, Tal and Cotterell, Ryan},
  editor = {Warstadt, Alex and Mueller, Aaron and Choshen, Leshem and Wilcox, Ethan and Zhuang, Chengxu and Ciro, Juan and Mosquera, Rafael and Paranjabe, Bhargavi and Williams, Adina and Linzen, Tal and Cotterell, Ryan},
  year = 2023,
  month = dec,
  pages = {1--34},
  publisher = {Association for Computational Linguistics},
  address = {Singapore},
  doi = {10.18653/v1/2023.conll-babylm.1},
  urldate = {2024-11-22}
}

@inproceedings{weietal2021,
  title = {Frequency {{Effects}} on {{Syntactic Rule Learning}} in {{Transformers}}},
  booktitle = {Proceedings of the 2021 {{Conference}} on {{Empirical Methods}} in {{Natural Language Processing}}},
  author = {Wei, Jason and Garrette, Dan and Linzen, Tal and Pavlick, Ellie},
  year = 2021,
  pages = {932--948},
  publisher = {Association for Computational Linguistics},
  address = {Online and Punta Cana, Dominican Republic},
  doi = {10.18653/v1/2021.emnlp-main.72},
  urldate = {2023-03-01},
  langid = {english}
}

@article{zwicky1987,
  title = {Slashes in the Passive},
  author = {Zwicky, Arnold M.},
  year = 1987,
  journal = {Linguistics},
  volume = {25},
  number = {4},
  issn = {0024-3949, 1613-396X},
  doi = {10.1515/ling.1987.25.4.639},
  urldate = {2022-06-06}
}
\newpage
\appendix
\section{Stimuli}
\label{sec:all_sentences}
This section lists the materials for acceptability judgments.
\subsection{Test sentences}
\label{sec:test-sentences}
\begin{longtable}{ll}
    \toprule
    Verb class & Sentence frame\\
    \midrule
    \multirow{5}{*}{Advantage}& My donation \gap{} many communities.\\
    & Your actions \gap{} your son.\\
    & Our friendship \gap{} our relationship. \\
    & The gift \gap{} my organization. \\
    & The treaty \gap{} both countries. \\
    \midrule
    \multirow{5}{*}{Price} & Your dish \gap{} ninety dollars.\\
    & The painting \gap{} 2000 dollars. \\
    & My initiative \gap{} some money.\\
    & Your book \gap{} thirty dollars. \\
    & His actions \gap{} the medal. \\
    \midrule
    \multirow{5}{*}{Ooze} & my friend \gap{} confidence.\\
    & The lightbulb \gap{} some light.\\
    & My machine \gap{} a sound. \\
    & The teacher \gap{} wisdom.\\
    & The trash \gap{} an odor.\\
    \midrule 
    \multirow{5}{*}{Estimation} & The caricature \gap{} an actor.\\
    & Your friend \gap{} my brother.\\
    & The sketch \gap{} my design.\\
    & Her son \gap{} her father.\\
    & The copy \gap{} the original.\\
    \midrule
    \multirow{5}{*}{Duration} & The journey \gap{} three days.\\
    & My meeting \gap{} two hours.\\
    & The surgery \gap{} some time. \\
    & Her speech \gap{} seventeen minutes. \\
    & His recovery \gap{} a month.\\
     \midrule
     \multirow{5}{*}{hit} & My brother hit your friend.\\ 
 & A boy hit my bag.\\ 
 & Your dog hit the toy.\\ 
 & The child hit a monkey.\\ 
 & The arrow hit the target.\\
\midrule
\multirow{5}{*}{kicked} & My brother kicked your friend.\\ 
 & A boy kicked my bag.\\ 
 & Your dog kicked the toy.\\ 
 & The child kicked a monkey.\\ 
 & My friend kicked the wall.\\ 
 \midrule
 \multirow{5}{*}{carried} & A boy carried my bag.\\ 
 & My brother carried your friend.\\ 
 & The dog carried the toy.\\ 
 & Your mother carried the child.\\ 
 & The donkey carried the load.\\ 
\midrule
\multirow{5}{*}{pushed} & A boy pushed the cup.\\ 
 & My brother pushed a child.\\ 
 & A child pushed the bag.\\ 
 & The mother pushed my toy.\\ 
 & Your sister pushed your friend.\\ 
\midrule
\multirow{5}{*}{washed} & A boy washed the cup.\\ 
 & My brother washed my plate.\\ 
 & A child washed the bag.\\ 
 & The mother washed my toy.\\ 
 & Your sister washed a towel.\\ 
\midrule
\multirow{5}{*}{dropped} & A boy dropped the cup.\\ 
 & My brother dropped my plate.\\ 
 & A child dropped the bag.\\ 
 & The mother dropped my toy.\\ 
 & Your sister dropped a book.\\
\bottomrule
\end{longtable}

\newpage
\subsection{Filler sentences}
\label{sec:filler-sentences}
\begin{longtable}{ll}
    \toprule
    Type & Sentence \\
    \midrule
    \multirow{24}{*}{Acceptable} & She was worried about the problem. \\
    & Your knife needs to be sharpened. \\
    & Her sister failed her test. \\
    & The bank is located across the road. \\
    & His mother thought that your friendship was strong. \\
    & Attention check: select 'Completely acceptable'. \\
    & The dog bit its owner. \\
    & The girl was unexcited about the trip. \\
    & Her father said that your recovery was quick. \\
    & The meeting ended quickly. \\
    & Your sister claimed that the machine broke. \\
    & A woman sang beautifully. \\
    & The ship was sunk by the enemy. \\
    & The opportunity presented itself. \\
    & His sister slept at my house. \\
    & Your child played the game. \\
    & My brother sold your friend a plate. \\
    & It rained yesterday at noon. \\
    & Attention check: select 'Completely acceptable'. \\
    & Your job requires concentration. \\
    & My mother read the child a book. \\
    & The goldfish died alone. \\
    & The monkey wanted to eat a banana. \\
    & Glass bottles are very fragile. \\
    \midrule
    \multirow{14}{*}{Unacceptable} & A bottle breaking last night. \\
    & The company lent the employee. \\
    & A cat met either mouse. \\
    & My sister said a word all night. \\
    & Your friend is walks home. \\
    & On a book the floor sat. \\
    & The driver handed the keys. \\
    & An infant asleep. \\
    & My friend liked your car at all. \\
    & A doctor was give the dog a toy. \\
    & A ball hit with great force. \\
    & Puppy my bit hand your. \\
    & The teacher bought for the students. \\
    & Candlesticks a picnic. \\
    \multirow{32}{*}{Unacceptable} & A student playing piano well. \\
    & My bottle holds. \\
    & Sat on the floor your sister. \\
    & Her daughter will watches a movie. \\
    & Attention check: select 'Completely unacceptable'. \\
    & My key a cabinet. \\
    & The boy saw anyone. \\
    & The chicken killed. \\
    & Snack this delicious taste. \\
    & That wall are green. \\
    & The car the light. \\
    & Your friend lifted a finger to help. \\
    & His friend is painted his grandmother a portrait. \\
    & The child brought to school. \\
    & The boy looked the picture. \\
    & The cow are grazing in the field. \\
    & The classroom silent. \\
    & Any girls passed the test. \\
    & My feelings were hurting by my brother. \\
    & The class went to on Tuesday. \\
    & The singer are practicing a song. \\
    & The opportunity some wallpaper. \\
    & There is every fly in my soup. \\
    & Attention check: select 'Completely unacceptable'. \\
    & The car driven. \\
    & The doctor disliked last week. \\
    & Box a opened the boy. \\
    & This plates has been chipped. \\
    & The bank will lend me. \\
    & Your backpack heavy. \\
    & Your mother bought any cups. \\
    & The essay was wrote by a genius.\\
    \bottomrule
 \end{longtable}
 
\section{Human acceptability judgment task instructions}
\label{sec:instructions}
\noindent\texttt{
In this experiment, you will rate English sentences based on how acceptable they sound to you. Try to answer based on your gut reaction, without analyzing the sentences. There are no `right' or `wrong' answers. The first two questions will be practice questions to familiarize you with the task. }

\texttt{< Participant clicks *Next* button >}

\noindent
\texttt{S1: How acceptable is this sentence?
The mirrors reflected light.}

\noindent\texttt{\textit{Hint}: For many people, this sentence is completely acceptable. Move the slider to the right corner of the scale to rate the sentence if you agree. Then, click the Next button or press the spacebar to continue.}

\texttt{< Participant rates S1 and clicks *Next* button to continue >}

\noindent
\texttt{S2: How acceptable is this sentence? The teacher was spoke.}

\noindent\texttt{Hint\textit{:} For many people, this sentence is completely unacceptable.
        Move the slider to the left corner of the scale to rate this sentence if you agree. 
        Then click the Next button or press the spacebar to continue}
        
\texttt{< Participant rates S2 and clicks *Next* button >}

\section{Sample parsing errors}
\label{sec:errors}
These four sentences were parsed by the spaCy model \texttt{en\textunderscore core\textunderscore webtrf} as passive uses of the verb \textit{last}:
\begin{enumerate}
    \item  It's lasted for 52 years so far, whether on television or in spin-off media, and that's in no small part because of the original idea to recast the title character in 1966, thus creating the concept of regeneration.
    \item But everyone saying this should have to add “however, he’s certainly lasted much longer than we originally predicted.”
    \item It’s lasted 75 years.
    \item Fans of the living dead have one man to thank for the birth of the modern cinematic zombie genre: George A. Romero, the filmmaker who made Night of the Living Dead on the cheap in 1968 and kicked off a zombie obsession that’s lasted for decades.
\end{enumerate}

\section{Raw verb counts in original and frequency-altered corpora}
\label{sec:raw_counts}
\begin{table}[h]
\setlength{\tabcolsep}{4pt}
\centering
\begin{tabular}{lcc|*{8}{c}}
\toprule
& \multicolumn{2}{c}{Original} & \multicolumn{2}{c}{carry} & \multicolumn{2}{c}{drop} & \multicolumn{2}{c}{hit} & \multicolumn{2}{c}{push} \\
\cmidrule(lr){2-3} \cmidrule(lr){4-5} \cmidrule(lr){6-7} \cmidrule(lr){8-9} \cmidrule(lr){10-11}
& Active & Passive & Active & Passive & Active & Passive & Active & Passive & Active & Passive\\
Original && & 7660 & 2181 & 3288 & 1143 & 9372 & 2452 & 4461 & 1269 \\
\midrule
last & 723 & 4 & 7591 & 42 & 3253 & 18 & 9218 & 51 & 4338 & 24 \\
cost & 4131 & 21 & 7475 & 38 & 3147 & 16 & 9245 & 47 & 4327 & 22\\
resemble & 1373 & 1 & 6865 & 5 & 2746 & 2 & 8238 & 6 & 4119 & 3\\
\bottomrule
\end{tabular}
\caption{Verbs' frequency of occurrence in original corpus and after interventions to match the relative active-passive ratio of the target verb.}
\label{tab:data_triples}
\end{table}

\section{Experiment~3 stimuli details}
\subsection{Verbs used in Experiment~3 stimuli}
\label{sec:exp3_verbs}
\paragraph{Verbs used in high-affectedness training sentences}

executed (129), drowned (128), crushed (126), stabbed (118), burned (113), murdered (105), suffocated (77), terrorized (76), broke (69), strangled (69), exterminated (64), frightened (61), knifed (59), demolished (55), robbed (44), pushed (42), shot (36), hit (28), wrecked (28), killed (26), boiled (24), kicked (24), smashed (20), destroyed (19), shattered (16), bit (15), hammered (14), punched (14), shook (13), crafted (10), designed (10), annihilated (9), assassinated (9), eradicated (9), painted (9), constructed (8), built (7), discovered (7), invented (7), obliterated (7), programmed (7), thumped (7), composed (6), devastated (6), developed (6), sculpted (6), baked (5), dismantled (5), engineered (5), exploded (5), investigated (5), knocked (5), repaired (5), rescued (5), saved (5), carved (4), cooked (4), extinguished (4), solved (4), wrote (4), collapsed (3), completed (3), decimated (3), eliminated (3), filmed (3), planted (3), stole (3), terrified (3), vaporized (3), arrested (2), brewed (2), burnt (2), captured (2), caught (2), collected (2), created (2), devoured (2), directed (2), engulfed (2), fixed (2), formulated (2), hacked (2), harvested (2), ignited (2), incinerated (2), launched (2), massacred (2), plowed (2), pruned (2), published (2), shelved (2), struck (2), taught (2), toppled (2), tried (2), trimmed (2), uncovered (2), abandoned (1), abducted (1), altered (1), ambushed (1), analyzed (1), apprehended (1), aspired (1), assailed (1), attacked (1), attempted (1), authorized (1), axed (1), bashed (1), battered (1), beat (1), blew (1), bombed (1), breathed (1), bulldozed (1), capsized (1), chiseled (1), choked (1), chopped (1), coded (1), comforted (1), conducted (1), cracked (1), crashed (1), crossed (1), cultivated (1), curated (1), cut (1), defeated (1), delivered (1), detonated (1), discarded (1), displaced (1), drew (1), drove (1), edited (1), emptied (1), enacted (1), erased (1), evaporated (1), explained (1), fired (1), flattened (1), forged (1), found (1), grilled (1), hung (1), imagined (1), influenced (1), injured (1), inked (1), innovated (1), inspected (1), kneaded (1), lit (1), looted (1), manufactured (1), mapped (1), mugged (1), neutralized (1), ordered (1), organized (1), photographed (1), picked (1), pinned (1), plucked (1), poisoned (1), polished (1), popped (1), prepared (1), pulverized (1), pummeled (1), purchased (1), questioned (1), recorded (1), removed (1), replaced (1), ruptured (1), sank (1), scared (1), sentenced (1), set (1), sewed (1), shaped (1), shocked (1), shoved (1), slaughtered (1), slayed (1), sliced (1), squashed (1), steered (1), studied (1), suspected (1), swallowed (1), synthesized (1), tailored (1), thwarted (1), torched (1), tore (1), trampled (1), translated (1), unearthed (1), unraveled (1), violated (1), won (1), wracked (1)

\paragraph{Verbs used in high-affectedness test sentences}
executed (11), exterminated (7), crushed (6), burned (5), drowned (5), knifed (5), terrorized (5), broke (4), pushed (4), stabbed (4), demolished (3), frightened (3), suffocated (3), assassinated (2), built (2), murdered (2), wrecked (2), believed (1), bit (1), conducted (1), crashed (1), crippled (1), destroyed (1), developed (1), discovered (1), fried (1), hammered (1), hit (1), incinerated (1), kicked (1), obliterated (1), ordered (1), photographed (1), punched (1), robbed (1), saved (1), shot (1), slaughtered (1), smothered (1), stole (1), strangled (1), transformed (1), weakened (1), wrote (1)

\paragraph{Verbs used in low-affectedness training sentences}

heard (113), believed (108), resembled (108), feared (103), noticed (101), recognized (95), lacked (94), trusted (89), missed (85), cost (82), overheard (72), respected (68), saw (67), needed (66), bordered (65), liked (65), dreaded (64), remembered (63), had (57), understood (56), knew (52), abutted (44), fit (41), received (39), spotted (28), totaled (28), lasted (23), looked (19), sensed (19), underwent (13), slept (12), admired (3), appeared (3), contained (3), exceeded (2), included (2), overlooked (2), possessed (2), provided (2), represented (2), accommodated (1), appreciated (1), approached (1), arrived (1), belonged (1), carried (1), cautioned (1), conveyed (1), created (1), delivered (1), described (1), disliked (1), displayed (1), enclosed (1), ended (1), envied (1), evoked (1), felt (1), fitted (1), hated (1), held (1), led (1), matched (1), misunderstood (1), observed (1), owned (1), packed (1), produced (1), published (1), ran (1), recalled (1), reflected (1), regretted (1), seated (1), seemed (1), sounded (1), spanned (1), was (1), weighed (1), wore (1)

\paragraph{Verbs used in low-affectedness test sentences} liked (7), missed (6), recognized (6), trusted (6), abutted (5), believed (5), feared (5), heard (5), knew (5), resembled (5), cost (4), fit (4), had (4), remembered (4), dreaded (3), lacked (3), noticed (3), overheard (3), respected (3), understood (3), needed (2), spotted (2), contained (1), lasted (1), received (1), saw (1), sensed (1), slept (1), underwent (1)
\subsection{GPT-4o prompt for creation of affected sentences}
The prompt used to generate high-affectedness sentences is given below. The scores reported in the example below are the mean scores given to the sentence by humans in the Semantic Proto-Roles dataset \citep{reisingeretal2015}. The low-affectedness prompt uses the same prompt format with different target values (\texttt{1s} instead of \texttt{5s}) and a different example adapted from the dataset that received low affectedness scores.
\label{sec:exp3_prompts}
\begin{spverbatim}
    You will be given a number. Generate that number of examples by following these steps.
1. Find a verb to use. Try to pick a verb that will score all 5s on the tasks below.
2. Use the verb in a transitive sentence in the past tense. This means that the sentence contains a subject and an object, for instance 'After going to the supermarket, the man ate two bagels after dinner'. Let's call this the target sentence. Here, the subject is 'the man who loved watching television', and the object is 'two bagels'. An example of a sentence that is not transitive is 'The worry ate at the man' or 'The man ate'. Try to make a sentence that will score all 5s on the tasks below.
3. Find the subject and object of the target sentence.
4. Create 2 sentences of context before the target sentence and 2 sentences after the target sentence.
5. Rate the sentence on a scale of 1-5 for how likely it is that the subject of the target sentence caused the action to happen.
6. Rate the sentence on a scale of 1-5 for how likely it is that the subject of the target sentence chose to be involved in the action.
7. Rate the sentence on a scale of 1-5 for how likely it is that the subject of the target sentence was aware of the action.
8. Rate the sentence on a scale of 1-5 for how likely it is that the subject of the target sentence was sentient.
9. Rate the sentence on a scale of 1-5 for how likely it is that the object of the target sentence was altered or somehow changed during or by the end of the action in the target sentence.
10. Rate the sentence on a scale of 1-5 for how likely it is that the action in the target sentence caused a change in the object of the target sentence.
11. Rate the sentence on a scale of 1-5 for how likely it is that the object of the target sentence changed possession during the action in the target sentence.
12. Rate the sentence on a scale of 1-5 for how likely it is that the object of the target sentence changed location during the action in the target sentence.
13. Repeat steps 1-13 using verbs and sentences that will score all 5s on the ratings. Do this until there are as many examples as was requested by the user. 
14.  Report the scores for all of the sentences in json format. In the json, include the following fields: 'subject', 'object', 'verb', 'sentence', 'paragraph', and 'scores', as follows: 
[{'subject': '...', ...},
...,
{'subject': '...', ...},
{'subject': '...', ...}
]

------
EXAMPLE
Let's think step by step. First, we find a verb that we want to use. 
Verb: killed.
Next, we make a sentence using that verb.
Sentence: Saddam killed every month more people than all those who died from suicide murders since the Coalition occupation of Iraq.
Now we know the subject and object of the sentence: the subject is 'Saddam', the object is 'more people than all those who died from suicide murders since the Coalition occupation of Iraq', and the action is 'killed'.
Then, we create two sentences of context before and after the sentence.
Sentence in context: 
Sept. 11 was quantitatively much less lethal than many earthquakes. More people die from AIDS in one day in Africa than all the Russians who died at the hands of Chechnya-based Moslem suicide murderers since that conflict started. Saddam killed every month more people than all those who died from suicide murders since the Coalition occupation of Iraq. So what is all the fuss about suicide killings? It creates headlines.
Finally, we rate the sentence on a scale of 1-5 for each of the questions.
First, rating for how likely it is that the subject of the target sentence, 'Saddam', caused the action, 'killing', to happen.
Rating: 5
Rating for how likely it is that the subject of the sentence,'Saddam', chose to be involved in the action 'killing'.
Rating: 5
Rating for how likely it is that the subject of the target sentence, 'Saddam', was aware of the action 'killing'.
Rating: 5
Rating for how likely it is that the subject of the target sentence, 'the criminal', was sentient.
Rating: 5
Rating for how likely it is that the object of the target sentence, 'more people... than since the conflict started', was altered or somehow changed during or by the end of the action 'kidnapping'.
Rating: 4.5
Rating for how likely it is that the action in the target sentence caused a change in the object of the target sentence.
Rating: 4.5
Rating for how likely it is that the object of the target sentence changed possession during the action in the target sentence.
Rating: 2.5
Rating for how likely it is that the object of the target sentence changed location during the action in the target sentence.
Rating: 4

Now we report our answer as a json.
```
[
{'subject': 'Saddam',
'object': 'more people than all those who died from suicide murders since the Coalition occupation of Iraq',
'verb': 'killed',
'sentence': 'Saddam killed every month more people than all those who died from suicide murders since the Coalition occupation of Iraq.',
'paragraph': 'Sept. 11 was quantitatively much less lethal than many earthquakes. More people die from AIDS in one day in Africa than all the Russians who died at the hands of Chechnya-based Moslem suicide murderers since that conflict started. Saddam killed every month more people than all those who died from suicide murders since the Coalition occupation of Iraq. So what is all the fuss about suicide killings? It creates headlines.',
'ratings':  [5,5,5,5,4.5,4.5,2.5,4]
}
...
]
```
\end{spverbatim}
\end{document}